\documentclass{article}

\usepackage[final]{neurips_2021}
\PassOptionsToPackage{square,sort,comma,numbers}{natbib}

\usepackage[utf8]{inputenc} 
\usepackage[T1]{fontenc}    
\usepackage{hyperref}       
\usepackage{url}            
\usepackage{booktabs}       
\usepackage{amsfonts}       
\usepackage{nicefrac}       
\usepackage{microtype}      
\usepackage{xcolor}         

\usepackage{subcaption}
\usepackage{enumitem}
\usepackage{amsmath}
\usepackage{amssymb}
\usepackage{amsthm}
\usepackage{graphicx}
\usepackage{makecell}
\usepackage{xargs}

\title{Similarity and Matching\\ of Neural Network Representations}

\author{
 \parbox{\linewidth}{\centering
Adrián Csiszárik$^{\dag, \circ}$, Péter Kőrösi-Szabó$^{\dag}$, Ákos K. Matszangosz$^{\dag}$\\ Gergely Papp$^{\dag}$, Dániel Varga$^{\dag}$
}
\\
\\
$^{\dag}$Alfréd Rényi Insititute of Mathematics, Budapest, Hungary\\
$^{\circ}$Eötvös Loránd University, Budapest, Hungary\\
\small{\texttt{\{csadrian, koszpe, matszang, gergopool, daniel\}@renyi.hu}}
}

\begin{document}

\maketitle

\begin{abstract}
We employ a toolset --- dubbed Dr. Frankenstein --- to analyse the similarity of representations in deep neural networks.
With this toolset, we aim to match the activations on given layers of two trained neural networks by joining them with a stitching layer. 
We demonstrate that the inner representations emerging in deep convolutional neural networks with the same architecture but different initializations can be matched with a surprisingly high degree of accuracy even with a single, affine stitching layer. We choose the stitching layer from several possible classes of linear transformations and investigate their performance and properties. The task of matching representations is closely related to notions of similarity. Using this toolset, we also provide a novel viewpoint on the current line of research regarding similarity indices of neural network representations: the perspective of the performance on a task.
\end{abstract}

\newcommand{\R}{\mathbb{R}}
\newcommand{\N}{\mathbb{N}}
\newcommand{\la}{\lambda}
\newcommand{\norm}[1]{\left\lVert#1\right\rVert}
\newcommand{\bra}{\langle}
\newcommand{\ket}{\rangle}
\newcommand{\stb}{,\ldots,}
\newcommand{\CKA}{\operatorname{CKA}}
\newcommand{\CCA}{\operatorname{CCA}}
\newcommand{\SVCCA}{\operatorname{SVCCA}}
\newcommand{\RLR}{\operatorname{R^2_{LR}}}
\newcommand{\RV}{\operatorname{RV}}
\newcommand{\KA}{\operatorname{KA}}
\newcommand{\Hom}{\operatorname{Hom}}
\newcommand{\C}{\mathcal{C}}

\newtheorem{theorem}{Theorem}[section]
\newtheorem{definition}{Definition}[section]
\newtheorem{statement}{Statement}

\def\changemargin#1#2{\list{}{\rightmargin#2\leftmargin#1}\item[]}
\let\endchangemargin=\endlist 
\section{Introduction}
\label{sec:intro}

A central topic in the analysis of deep neural networks is the investigation of the learned representations. A good understanding of the features learned on inner layers provides means to analyse and advance deep learning systems. A fundamental question in this line of inquiry is \emph{``when are two representations similar?’'} We contribute to this line of research by introducing a novel approach to study the above question. The key idea of our work is to ask the following, somewhat different question:  \emph{``in what way are two representations similar, once we know that they are similar?’'} In this paper, we present a conceptual framework and a methodology that allows to meaningfully pose and study this question. For this purpose, we will distinguish two approaches to define similarity: \emph{representational similarity} and \emph{functional similarity}.

\paragraph{Representational similarity} Representational similarity notions define similarity via computing statistics on two different data embeddings that capture appropriate geometrical or statistical properties. There exist several statistical measures, \citep{raghu2017svcca,DBLP:conf/nips/MorcosRB18, kornblith2019similarity} each of which serves as a different notion of similarity. These measures have been used successfully to obtain valuable insights about the inner workings of deep neural networks \citep{nguyen2021do, mirzadeh2021linear, DBLP:journals/corr/abs-2008-11687, wu-etal-2020-similarity}.

\paragraph{Functional similarity} While representational similarity concerns the data embeddings, in contrast, functional similarity concerns the functions that produce and process these embeddings, i.e., parts of the function compositions that the neural networks realize. With these elements, we can ask new types of questions that we were not able to ask using solely the data embeddings. For example, we can ask the following:  \emph{``can network B achieve its task using the representations of network A?’’} \\ Our paper focuses on this specific theme. In particular, we investigate this by taking the activations of network A at a given layer, transform it with an affine map, then use the result as an input of the same layer in network B. In other words, we stitch together the two networks with an affine stitching layer. This technique first appeared in \citet{lenc_vedaldi_19} to study the invariance and equivariance properties of convolutional networks. Evaluating the performance of this combined network provides an alternative viewpoint on similarity of representations, the viewpoint of functional similarity.

We investigate this novel perspective of reflecting on representational similarity through the lens of functional similarity. A brief outline of this work\footnote{Code is available at the project website: \url{https://bit.ly/similarity-and-matching}.}
is the following:

\begin{itemize}[leftmargin=*]

    \item After a detailed introduction to model stitching (Section~\ref{section:introducing_frankenstein}) and presenting two types of stitching methods (Section~\ref{section:stitching_methods}), we empirically
    demonstrate that trained convolutional networks with the same architecture but different initializations can be stitched together, even with a single, affine stitching layer in many cases without significant performance loss (Section~\ref{section:frankensten_matching}). We will refer to this compatibility property as the `matchability’ of representations.

    \item Observing the matchability of representations as a quite robust property of common vision models, we have a wide range of examples in our hands when we know that representations are functionally similar. This permits us to study the relation between representational similarity measures and functional similarity. In our experiments, we show that the values of similarity indices are not necessarily indicative of task performance on a stitched network, and perhaps more surprisingly, high-performance stitchings can have differences in the values of similarity indices such as Centered Kernel Alignment (CKA) \citep{cortes2012algorithms, kornblith2019similarity}. 
    This also reflects on phenomena experienced by the `end-users’ of these indices. For instance, in the context of continual learning \cite{mirzadeh2021linear} observe that CKA remains constant, while the accuracy on previous tasks drops drastically (Section~\ref{section:similarity_indices}).

    \item Constraining the space of transformations in the stitching layer provides means to analyse how the information is organised in the representations. As an example of this probing methodology, we study bottleneck stitching layers, and show that when doing SVD in the representational space, the magnitudes of the principal components are not in direct correspondence with the information content of the latent directions (Section~\ref{section:lowrank}). As SVD is commonly used with embeddings, and there are representational similarity measures, for example SVCCA \citep{raghu2017svcca}, that use SVD as an ingredient, these results might be of interest to both theorists and practitioners.
    
    \item The transformation matrices of well-performing stitching layers capture how representations are functionally similar. This provides the opportunity to empirically investigate \emph{``in what way are two representations similar, once we know that they are similar?’’}. As a first step on this avenue, we present results regarding the uniqueness and sparsity properties of the stitching matrices (Section~\ref{section:properties}). 
    
\end{itemize}

\section{Related work}

\paragraph{Matching representations} The idea of 1x1 convolution \textit{stitching layers} connecting the lower half of a network with the upper half of another network first appears in \cite{lenc_vedaldi_19}. They demonstrate that such hybrid ``Franken-CNNs'' can approach the accuracy of the original models in many cases. For shallower layers, this phenomenon can appear even when the first network is trained on a different task, suggesting that lower level representations are sufficiently generic.

\textit{Mapping layers} as introduced by \cite{li_2016_ICLR} are single layer linear 1x1 convolutional networks mapping between two representations. They do not incorporate information from the second half of the network. The authors train mapping layers with L1 regularized MSE, given their main goal of identifying sparse mappings between the filters of two different networks. As \cite{li_2016_ICLR} notes, learning mapping layers can get stuck at poor local minima, even without L1 regularization. This forces the authors to restrict their experiments to the shallowest layers of the networks they inspect. We manage to overcome the issue of poor local optima by initializing from the optimal least squares matching (Section~\ref{subsec:initexperiment}), and as our experiments demonstrate, this alleviates the issue, resulting in stitched networks that often achieve the performance of their constituents.

Concurrently and independently of our work, \citet{bansal2021revisiting} also apply model stitching to investigate similarities between deep learning models. They observe that for a given task, even very different architectures can be successfully stitched, e.g., a transformer with a residual network.

\paragraph{Similarity indices} Canonical Correlation Analysis (CCA) provides a statistical measure of similarity. CCA itself has a tendency to give too much emphasis to the behavior of low-magnitude high-noise latent directions. Similarity indices were designed to make CCA more robust in this sense, namely Singular Value CCA \citep{raghu2017svcca} and Projection Weighted CCA \citep{DBLP:conf/nips/MorcosRB18}.

The current state of the art in measuring similarity is Centered Kernel Alignment (CKA) \citep{cristianini2001kernel, cortes2012algorithms, kornblith2019similarity}. \cite{kornblith2019similarity} conducted a comparative study of the measures CCA, SVCCA, PWCCA as similarity indices, and they observed that the other similarity indices do not succeed in identifying similar layers of identical networks trained from different initializations, whereas CKA manages to do so, therefore in this paper we chose to focus on the latter. \cite{kornblith2019similarity} also consider nonlinear CKA kernels, but as they demonstrate this does not lead to meaningful improvements, so we focus only on the linear variant (also commonly known as RV coefficient \cite{robert1976unifying}), in line with other literature employing CKA similarity \citep{nguyen2021do, mirzadeh2021linear, DBLP:journals/corr/abs-2008-11687, wu-etal-2020-similarity}.

\paragraph{Functional similarity} The general framework of knowledge distillation \citep{hinton2015distilling, romero2014fitnets} can be seen as fostering end-to-end functional similarity of a student network to a teacher network. This relation and the versatile utilisation of this technique underpins the importance of the notion of functional similarity. \citet{yosinski2014transferable} studies transfer learning from a related perspective.

\section{Introducing Dr. Frankenstein}
\label{section:introducing_frankenstein}

\subsection{Preliminaries and notation}

Let $f_{\theta}:\mathcal{X} \to \mathcal{Y}$ denote the input-output function of a feedforward artificial neural network with $m \in \N$ layers, thus, $f_{\theta} = f_{m} \circ \cdots \circ f_{1}$, where $\mathcal{X}$ is the input space, $\mathcal{Y}$ is the output space, $f_i:\mathcal{A}_{i-1}\to \mathcal{A}_{i}$ are maps between activation spaces $\mathcal{A}_{i-1}$ and $\mathcal{A}_{i}$ for $i \in [m]$ with $\mathcal{A}_{0} = \mathcal{X}$, and $\theta$ are the parameters of the network. We consider models that are trained on a dataset $D = \{(x_i, y_i)\}_{i=1}^{n}$ in a supervised manner with inputs $x_i \in \mathcal{X}$ and labels $y_i \in \mathcal{Y}$, where $n \in \N$ is the dataset size.

The central tool of our investigation involves splitting up the network at a given layer to a \emph{representation map} and a \emph{task map}. The \emph{representation map at layer $L$} is a function $R_{L}:\mathcal{X} \to \mathcal{A}_{L}$ that maps each data point $x_i$ to its \emph{activation vector} $a_{i,L}$ at layer $L$: $$ R_{L}(x_i) = f_{L} \circ \cdots \circ f_{1}(x_i) = a_{i,L},$$ where $\mathcal{A}_L$ is the activation space at layer $L$. The activation vectors for the dataset $D$ at layer $L$ are simply denoted by $A_L=(a_{i,L})_{i=1}^n$. The \emph{task map at layer $L$} is a function $T_L:\mathcal{A}_L \to \mathcal{Y}$ that maps an activation vector at layer $L$ to the final output of the network: $$T_L(a_{i,L}) = f_{m} \circ \cdots \circ f_{L+1}(a_{i,L}).$$ Thus, the input-output map of the network $f_{\theta}$ is simply the composition $f_{\theta} = T_L \circ R_L: \mathcal{X} \to \mathcal{Y}$ for all $L=1\stb m$. We will omit the index of the layer $L$ when this does not hurt clarity.

\subsection{The Frankenstein learning task}

\begin{definition}[Frankenstein network]

Fix two (pretrained) neural networks $f_{\theta}$ and $f_{\phi}$.
We will call a transformation $S:\mathcal{A}_{\theta,L}\to \mathcal{A}_{\phi,M}$ between the two activation spaces at layers $L$ and $M$ a \emph{stitching layer}. Given a stitching layer $S$, \emph{the Frankenstein} or \emph{stitched network} corresponding to $S$ is the composition $F=F_{S}:\mathcal{X} \to \mathcal{Y}$, $$F(x) = T_{\phi,M} \circ S \circ R_{\theta,L}(x),$$ 
where $T_{\phi, M}$ is the task map at layer $M$ for the network $f_{\phi}$ and $R_{\theta,L}$ is the representation map at layer $L$ for network $f_{\theta}$.
\end{definition}

The stitching layer thus realizes a correspondence between the representations of two neural networks: it transforms the activations of one network at a particular layer to be a suitable input to the corresponding layer in the other network. For better readability, we will often call $f_{\theta}$ as Model 1 and $f_{\phi}$ as Model 2, and index the corresponding data accordingly ($f_1=f_\theta$, $f_2=f_\phi$, etc.). In the experiments of this paper, Model 1 and Model 2 have the same architecture and $L=M$.

\begin{definition}[Frankenstein learning task]
Given two pretrained neural networks $f_\theta$ and $f_\phi$, a task given by a labeled dataset $D = \{(x_i, y_i)\in \mathcal{X}\times \mathcal{Y}:i=1\stb n\}$ and a loss function $\mathcal{L}: \mathcal{Y} \times  \mathcal{Y} \to \mathbb{R}$, we will call the \emph{Frankenstein learning task at layer $L$} the task of finding the stitching layer $S:\mathcal{A}_{\theta,L}\to \mathcal{A}_{\phi, L}$ which minimizes the total value of the loss $\mathcal{L}$ on the dataset $D$.
\end{definition}

\paragraph{Frankenstein is an evaluation framework} The Frankenstein learning task primarily provides an \emph{evaluation framework} for matching representations, and it does not tie our hands on how the transformation function $S$ is produced. As an evaluation metric, we can either use the mean loss value of the Frankenstein learning task on an evaluation set, or any other appropriate metric (e.g., accuracy in the case of classification tasks).

\paragraph{Stitching convolutional representations} Our toolset is generic, but here we will only consider convolutional architectures. Thus, the activation spaces have the structure of rank-3 tensors $\R^{w\times h\times c}$, where $w$ and $h$ are the spatial width and height and $c$ is the number of feature maps. We will only consider linear stitching layers of the form $M:\R^c\to \R^c$ ($1\times1$ convolutional layers). When formulating least squares problems on $n$ many activation tensors, we always mean the least squares problem in $\R^c$, which in practice means reshaping the $n\times w\times h\times c$ tensor to a $nwh\times c$ matrix.

\section{Methods to solve the Frankenstein learning task}
\label{section:stitching_methods}

\begin{figure}
\centering
\begin{subfigure}{.45\textwidth}
  \centering
    \includegraphics[width=1.0\linewidth]{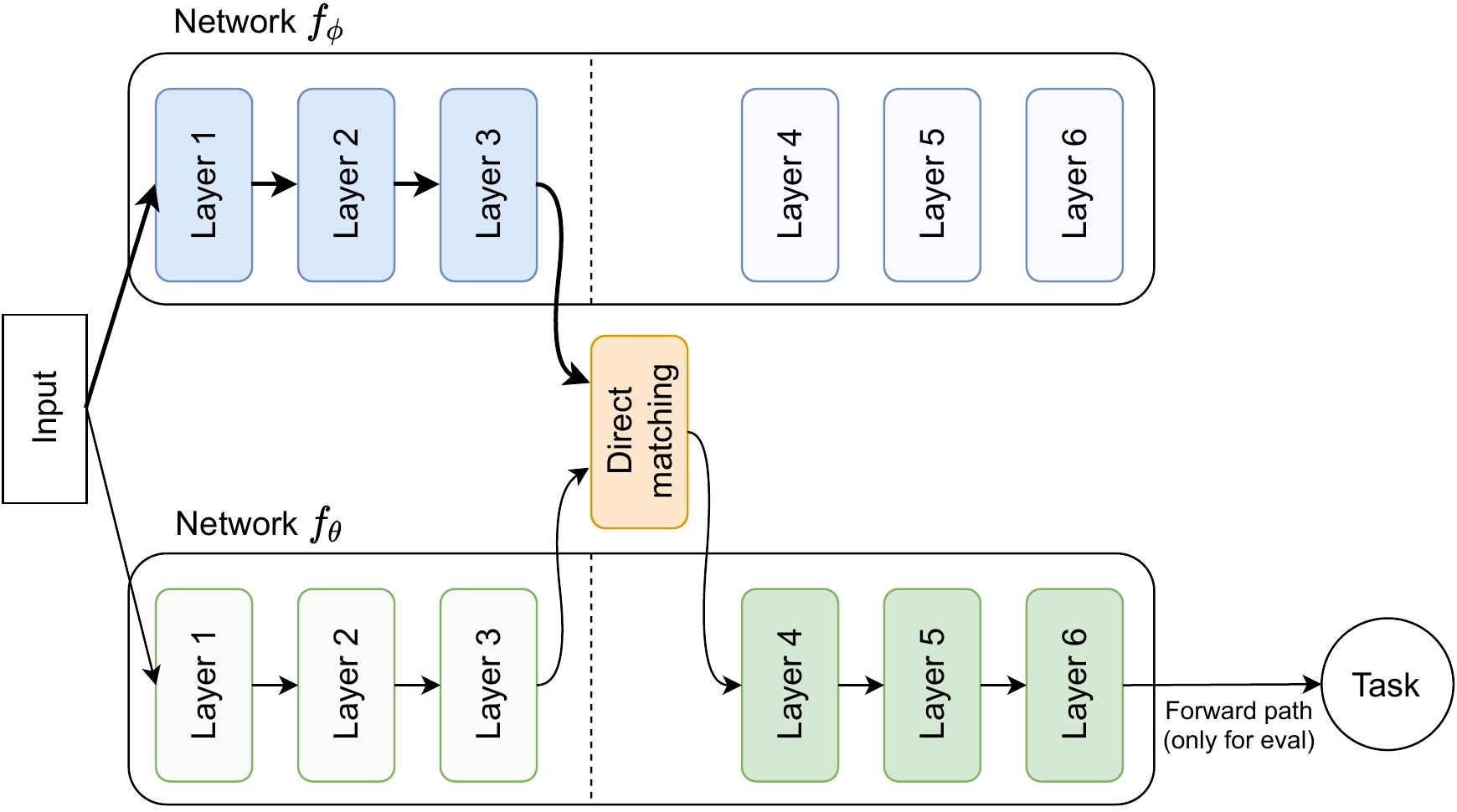}
  \caption{Direct matching of representations.}
  \label{fig:two-types-of-matching-direct}
\end{subfigure}
\begin{subfigure}{.45\textwidth}
  \centering
  \includegraphics[width=1.0\linewidth]{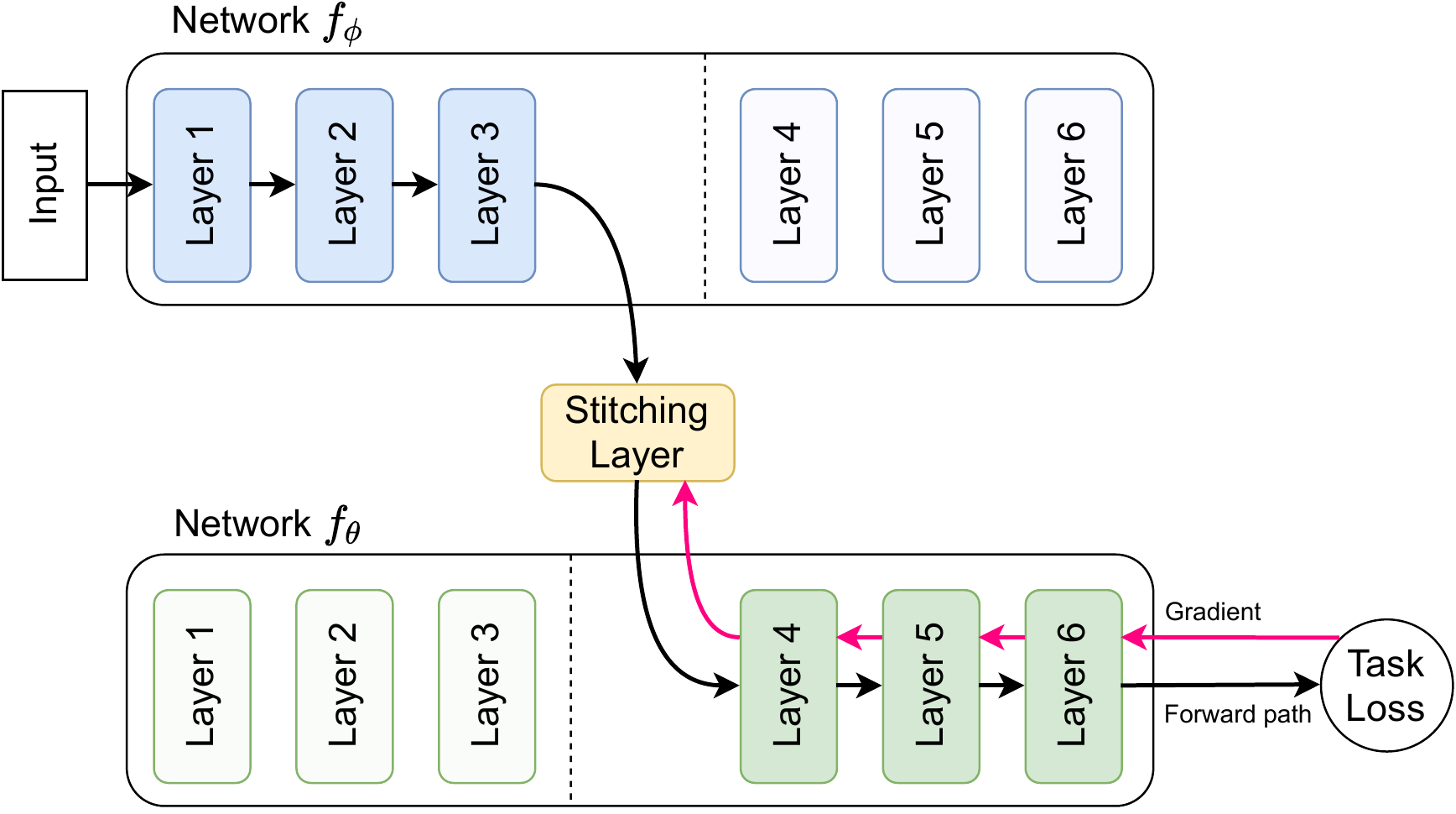}
  \caption{Task loss matching of representations.}
  \label{fig:two-types-of-matching-task-loss}\end{subfigure}%
\caption{Illustrating two types of matching approaches. \textbf{(a) Direct matching}: first align the representations using only the representations themselves, and use the forward path on the task map of Model 2 --- with the transformed activations --- only to evaluate the performance of the stitched model. \textbf{(b) Task loss matching}: the forward path from the input starts on Model 1, then through the stitching layer it continues on Model 2, then task loss gradients are backpropagated to update the weights of the stitching layer.}
\label{fig:two-types-of-matching}
\vspace{-1ex}
\end{figure}

Here we line up several possibilities to solve the Frankenstein learning task presented in the previous section. We distinguish two approaches: \textbf{(1) Direct matching}: utilize only the outputs of representation maps; \textbf{(2) Task loss matching}: utilize the loss of the Frankenstein learning task. Figure~\ref{fig:two-types-of-matching} illustrates the difference between the two approaches schematically.

\subsection{Task loss matching}

From the exposition of the Frankenstein learning task, a training method naturally follows: utilize the loss function in the Frankenstein learning task itself, and train an appropriately parametrised stitching layer with backpropagation (leaving the other parts of the Frankenstein network fixed).

We use the outputs of Model 2 as a soft label \citep{hinton2015distilling} and define the task loss with cross-entropy. Another meaningful option would be to use the original training objective (the one that was used for the training of Model 1 and Model 2, e.g., the one-hot labels in a classification task). Both are reasonable choices as tasks, one corresponding to the stitched network imitating Model 2, the other to solving the task itself. We have experimented with both losses and found very high correlation, with no difference regarding our main observations. We often present performance in terms of relative accuracy compared to Model 2, because it is easier to interpret than cross-entropy.

\subsection{Direct matching methods} \label{subsec:directmatching}

Due to their close connection to similarity indices of representations, we also consider methods that arrive at the transformation $S$ by using only the representations themselves, namely, utilizing only the outputs  $A, B\in \R^{n\times p}$ of the representation maps of Model 1 and Model 2, respectively.

\paragraph{Least squares methods} Given $A, B\in \R^{n\times p}$, and a class of transformations $\mathcal{C}\subseteq \Hom(\R^p, \R^p)$, we will consider least squares problems of the following kind: find an optimal $M_{o}\in \mathcal{C}$, such that
\begin{equation}\label{eq:LS}
\| AM_o-B\|_F= \min_{M\in \mathcal{C}}\| AM-B\|_F.
\end{equation}
We consider three variants of this problem: 1. arbitrary linear maps: $\C=\Hom(\R^p, \R^p)$, 2. orthogonal transformations: $\C=O(p)$, 3. linear maps in $\Hom(\R^p,\R^p)$ of rank at most $k$: $\C=\Sigma_k$.

In each case, there is a closed form expression relying on singular value decomposition. For $\C=\Hom(\R^p, \R^p)$, the minimal value of \eqref{eq:LS} is obtained through an orthogonal projection in the space of matrices \citep{penrose},
\begin{equation}\label{eq:lsmatching}
    M_o=M_{LS}:=A^\dagger B,
\end{equation} where $A^\dagger$ denotes the Moore-Penrose pseudoinverse of $A$. We will refer to this $M_{LS}$ as the \emph{least squares matching}. For results of this type, see Section \ref{section:frankensten_matching}.

In the case $\C=O(p)$, problem \eqref{eq:LS} is also known as the \emph{orthogonal Procrustes problem}, see e.g.\ \cite{GolubVanLoan2013}. If  $A^TB=US V^T$ is a singular value decomposition, then the optimal orthogonal map is $M_o=UV^T$.

Finally, the case $\C=\Sigma_k$ is also known as \emph{reduced rank regression} \cite{RRR}. In terms of the SVD of $AM_{LS}=USV^T$, the optimal rank $k$ map $M_o$ is given by $M_o=M_{LS}V_kV_k^T$, where $V_k$ consists of the first $k$ columns of $V$. For reduced rank matching results, see Section \ref{section:lowrank}.

\paragraph{Sparse matchings}\label{subsec:sparse} It is known \citep{wang18} that representations are not fully local, that is, matching cannot be achieved by permutation matrices. High quality sparse matchings between neurons can be interpreted to imply ``slightly distributed'' representations \citep{li_2016_ICLR}.

To achieve sparsity, a natural approach is to add the regularization term $\|M\|_{1}=\sum |m_{ij}|$ to the loss function, where $M$ is the transformation matrix. Then one can consider the L1 regularized versions of 1) task loss matching, and 2) least squares matching (Lasso regression, as in \cite{li_2016_ICLR}):
\begin{equation}\label{eq:lasso}
    \mathcal{L}(A,B)=\|AM-B\|_F + \alpha\cdot \|M\|_1,
\end{equation}
where $A,B$ denote the activations of the two models. See Section \ref{subsec:sparsematching} for these experiments.

\section{Matching representations in deep convolutional networks}
\label{section:frankensten_matching}

We now continue our exposition by presenting a series of experiments regarding matching neural representations. Our results in this section can be summarized in the following statement:

\emph{Neural representations arising on a given layer of convolutional networks that share the same architecture but differ in initialization can be matched with a single affine stitching layer, achieving close to original performance on the stitched network.}

\subsection{Experiment 1: Matching with least squares and task loss matching}

In this experiment, we compare the performance of direct matching and task loss matching. We take two networks of identical architectures, but trained from different initializations. We perform a direct matching as well as a task loss matching from Model 1 to Model 2. We then compare the performance of the stitched network to the performance of Model 2.

We conduct experiments on three different convolutional architectures: a simple, 10-layer convnet called Tiny-10 (used in \cite{kornblith2019similarity}), on a ResNet-20 \citep{resnet}, and on the Inception V1 network \citep{inceptionv1}. These networks are rather standard and represent different types of architectures. Tiny-10 and ResNet-20 are trained and evaluated on CIFAR-10, Inception V1 on the 40-label CelebA task. Further training details follow common practices (see Appendix~\ref{appendix:exp1_details}).

Figure~\ref{fig:frankenstein_vs_psinv} summarizes the results: task loss matching achieves near perfect accuracy on ResNet-20, and also succeeds in aligning representations on Tiny-10 without collapsing accuracy. While the direct matching significantly underperforms the task loss matching and collapses performance on several layers, it is still remarkable that direct matching performs fairly well in many cases: e.g., on the layers between Layer1.0 and Layer2.1 on ResNet-20 or almost every case on Tiny-10.

\paragraph{Further results} 

We have experimented with networks with batch normalization and without batch normalization, and networks trained with different optimizers (SGD, Adam). Still, we did not encounter any cases where there were unmatchable instances. The worst performing stitchings were after the first convolutional layer on ResNet-20, still achieving 80\% relative accuracy. While this method is not universal on arbitrary architectures, this does not modify the observation that `matchability’ of representations is a quite robust property of common vision models.

In the case of task loss matching, we found the most stable results when we initialized the training of the stitching layer from the least squares matching \eqref{eq:lsmatching}. See Section~\ref{subsec:initexperiment} on further details on how the initialization influences performance. We also evaluated Procrustes matching described in Section \ref{subsec:directmatching}, and found that on later layers, it achieves the same level of accuracy as linear least squares matching.

As detailed in Section~\ref{appendix:lucid} of the Appendix, accuracy drop during stitching can be surprisingly low even when we stitch networks trained on different datasets; in our case, ImageNet and CelebA.

\begin{figure}[t!]
\centering
\begin{subfigure}[t]{.32\textwidth}
  \centering
    \vspace{0pt}
    \includegraphics[width=1.0\linewidth]{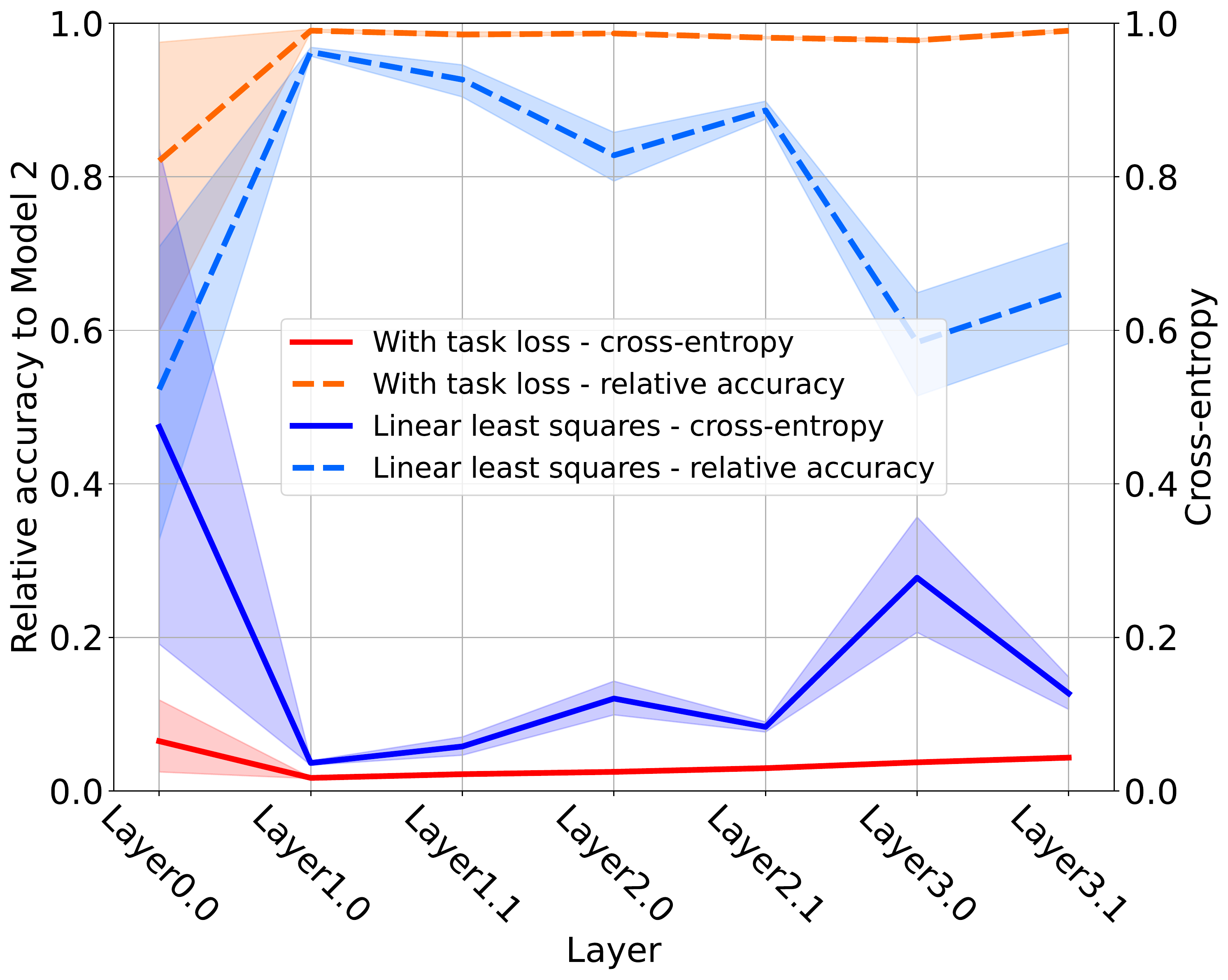}
    \caption{Results with ResNet-20.}
\end{subfigure}
\begin{subfigure}[t]{.32\textwidth}
  \centering
  \vspace{0pt}
  \includegraphics[width=1.0\linewidth]{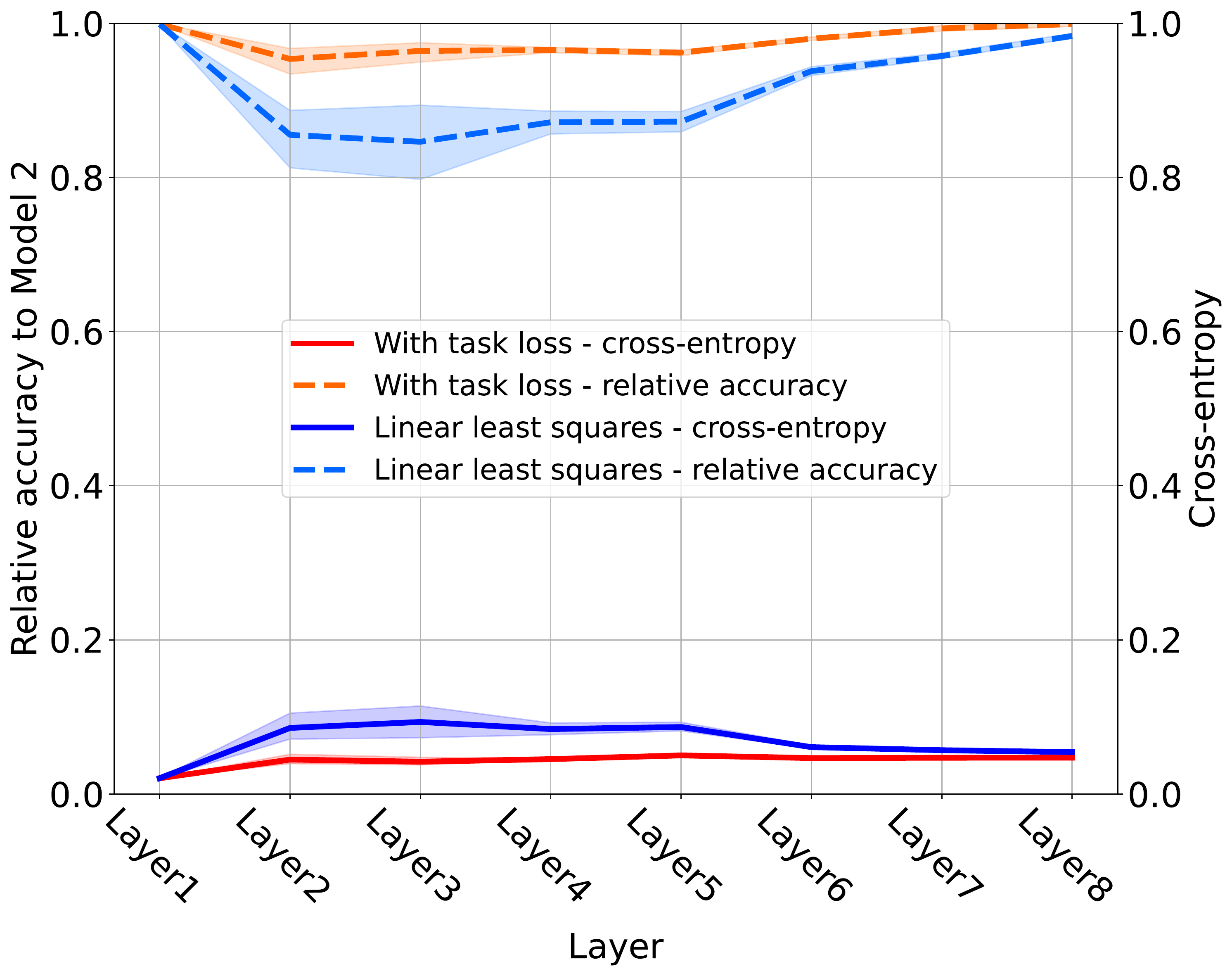}
    \caption{Results with Tiny10.}
\end{subfigure}
\begin{subfigure}[t]{.32\textwidth}
  \centering
  \vspace{0pt}
  \includegraphics[width=1.0\linewidth]{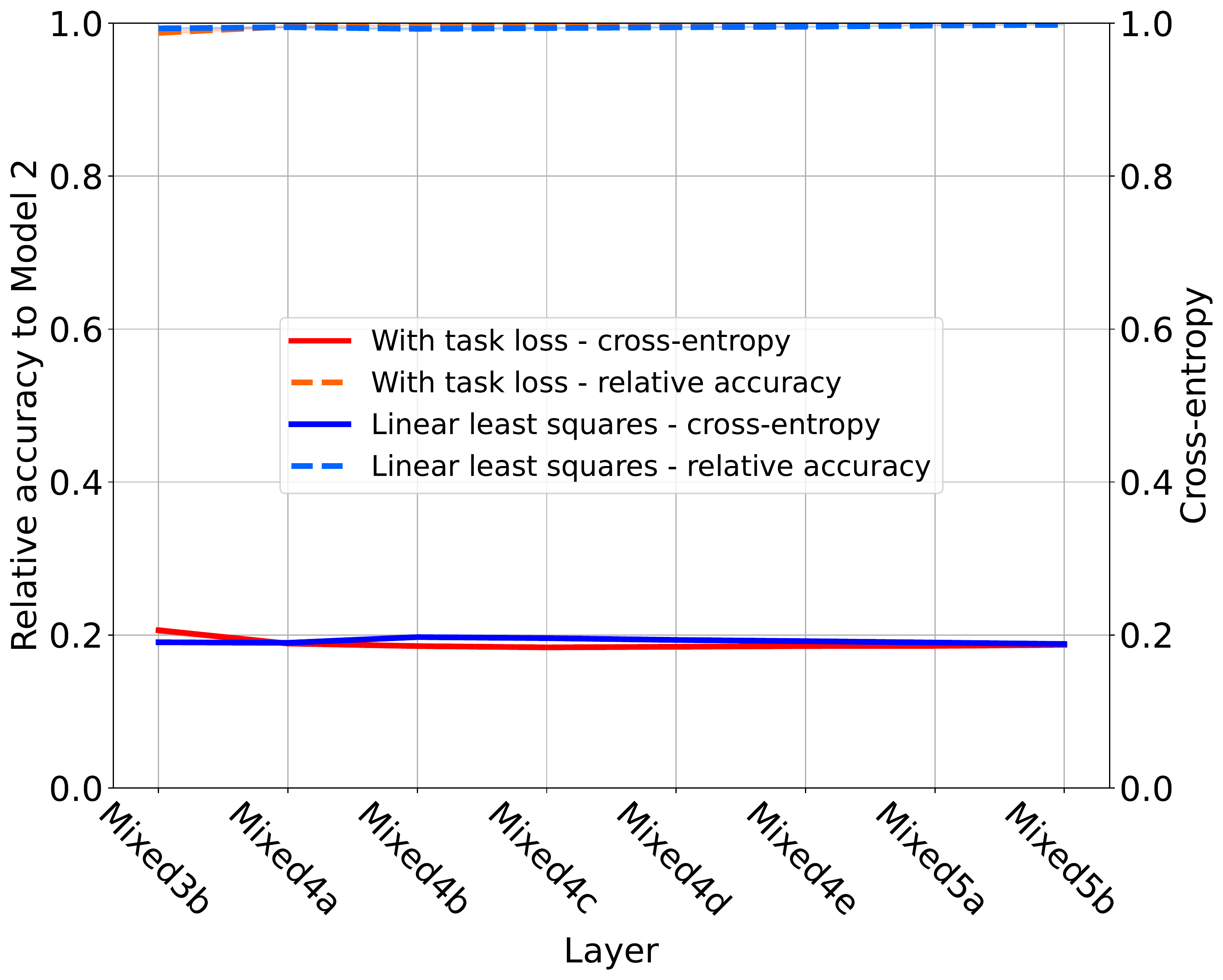}
    \caption{Results with Inception V1.}
\end{subfigure}
\caption{Performance of Frankenstein networks stitched together at specific layers (horizontal axis) measured in relative accuracy and cross-entropy (left and right vertical axes, respectively). This experiment compares two methods: task loss matching and linear least squares matching (a direct matching method). Results are averages of 10 runs, the bands represent standard deviations. As baseline values for cross-entropy, we provide average cross-entropies between Model 2 and a model stitched without transformation: Tiny-10: $6.99 \pm 1.96$, ResNet-20: $4.69 \pm 2.91$, InceptionV1: $0.36 \pm 0.08$ (averaged across all layers). 
}
\label{fig:frankenstein_vs_psinv}
\vspace{-2ex}
\end{figure}

\subsection{Experiment 2: Networks of different width}

\begin{figure}[t!]
\centering
\includegraphics[width=1.0\linewidth]{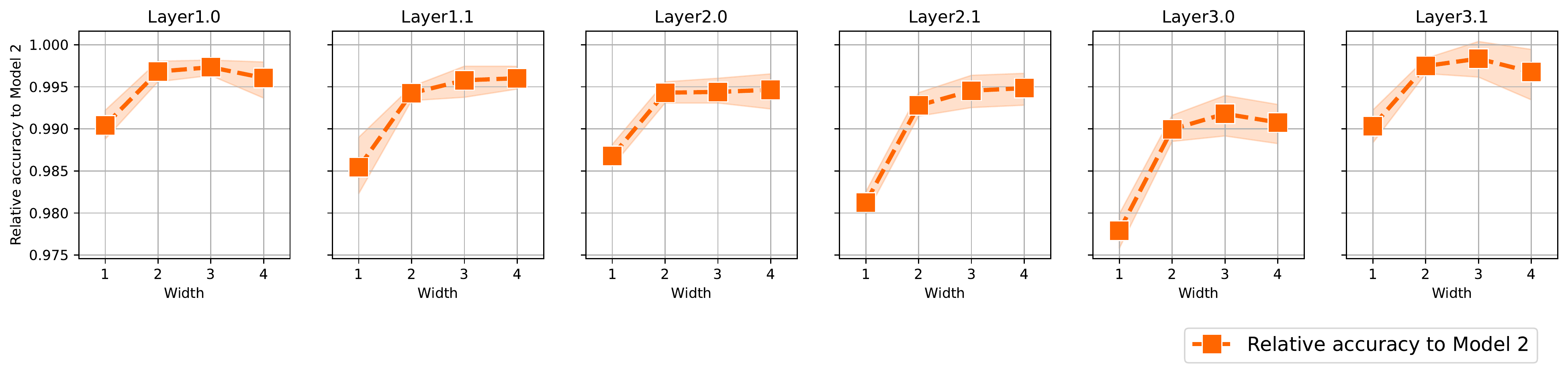}
\caption{Relative accuracy values with ResNet-20 nets of different width. The horizontal axis shows the width multiplier compared to the baseline network. Plots correspond to different layers with names denoted at the top. Averages of 10 runs, bands show standard deviations.}
\label{fig:width_plot}
\vspace{-3ex}
\end{figure}

This experiment focuses on the effect of network width on stitching performance. Matching happens with cross-entropy to Model 2 output as the task loss. The stitching is initialized to the optimal least squares matching. Performance is measured on the original classification task, relative to Model 2 performance. Figure~\ref{fig:width_plot} shows the results on ResNet-20 with various widths. The relative accuracy for 1x-width is already very high, but a clear trend of performance increasing with width is still visible.

\section{Similarity indices and matching representations}
\label{section:similarity_indices}

We now turn our attention to the relation of similarity and matchability of representations.  Finding an appropriate similarity index for neural representations is an active area of research \citep{HSIC, raghu2017svcca, DBLP:conf/nips/MorcosRB18, kornblith2019similarity}. In the pursuit of a suitable notion, one seeks to capture the right invariance and equivariance properties to provide a tool with theoretical guarantees and practical applicability. Our brief summarizing statement in this section acts as a warning for the end-users of statistical similarity indices:

\emph{By their nature, similarity indices depend on the task only indirectly, and their values are not necessarily indicative of task performance on a stitched network.}

\subsection{Similarity indices}
\label{subsec:similarityindices}
Given two sets of activations $A=(a_i)_{i=1}^n, B= (b_j)_{j=1}^n,$ in the activation spaces $a_i\in \mathcal{A}_1= \R^s, b_j\in \mathcal{A}_2= \R^r$, a \emph{similarity index} $s(A,B)\in [0,1]$, measures the similarity of the two data representations, i.e.\ it is a statistic on the observation space $s:\mathcal{A}_1\times \mathcal{A}_2\to [0,1]$.  We now briefly review the similarity indices under consideration. First, the least squares matching $M_{LS}$ defined earlier can be used to define a statistic that can be used as a similarity index \citep{kornblith2019similarity}:
\begin{definition}
 \emph{The coefficient of determination} of the best fit is given in terms of the optimal least squares matching $M_{LS}=A^\dagger B$ as:
$$ R^2_{LR}(A,B)=1-\frac{\lVert AM_{LS}-B \lVert _F^2}{\lVert B \lVert_F^2}.$$
\end{definition}
Let $k, l$ be positive definite kernels on a space $\mathcal{X}$ and let $K_{ij}=k(x_i,x_j)$ and $L_{ij}=l(x_i,x_j)$ be the Gram matrices of a finite sample $x_i\in \mathcal{X}$, $i=1\stb m$ drawn from a distribution on $\mathcal{X}$. The (empirical) \emph{alignment of the Gram matrices $K, L$} was defined by \cite{cristianini2001kernel} as
$$\KA(K,L)=\frac{\bra K, L\ket_F}{\| K\|_F\cdot \|L\|_F}.$$
The importance of centering the kernels in practical applications was highlighted in \cite{cortes2012algorithms}. Given a Gram matrix $K$, denote the \emph{centered Gram matrix} by $K^c:=CKC$, where $C=I_m-\frac{\mathbf{1}\cdot\mathbf{1}^T}{m}$ is the centering matrix of size $m$. The \emph{Centered Kernel Alignment of the Gram matrices $K,L$} is defined \citep{cortes2012algorithms} as $ \CKA(K,L)=\KA(K^c, L^c).$ If the kernels are linear on their respective feature spaces, and if $X,Y$ are samples of the distribution in these feature spaces, then $\CKA(K_X,L_Y)= \KA(K_{XC}, L_{YC})$, using that $K_X^c=K_{XC}$. In particular, in the case of linear kernels $\CKA$ is also referred to as the \emph{RV coefficient} \citep{robert1976unifying}. Furthermore, if $X, Y$  are the \emph{centered} matrices representing the samples of the distribution in the feature space, $\CKA$ can be given by the following definition:
\begin{definition}
The \emph{empirical (linear) Centered Kernel Alignment} or \emph{RV coefficient} of the samples $X,Y$ centered in feature space is
$$\RV(X,Y):=\CKA(K_X,K_Y)=\frac{\| X^TY\|_F^2 }{\| X^TX\|_F \cdot \| Y^TY\|_F}.$$
\end{definition}

In our context, the probability distribution describes the data distribution in $\mathcal{X}$, the feature spaces are the activation spaces and the feature maps are the representation maps $R:\mathcal{X}\to \mathcal{A}$.

As a measure of similarity of representations, $\CKA$ was introduced and used successfully in \cite{kornblith2019similarity}. They also considered other similarity indices such as CCA, SVCCA, PWCCA  \citep{raghu2017svcca,DBLP:conf/nips/MorcosRB18} in their experiments, but they found that the other similarity indices do not succeed in identifying similar layers of identical networks trained from different initializations, whereas CKA manages to do so. For this reason, we chose to focus our attention on CKA.

\vspace{-1ex}

\subsection{Experiment 3: Similarity indices and task performance}
\label{subsec:similarityandtask}

\begin{figure}[t!]
\centering
    \begin{minipage}[t]{0.63\textwidth}
        \includegraphics[width=0.485\linewidth]{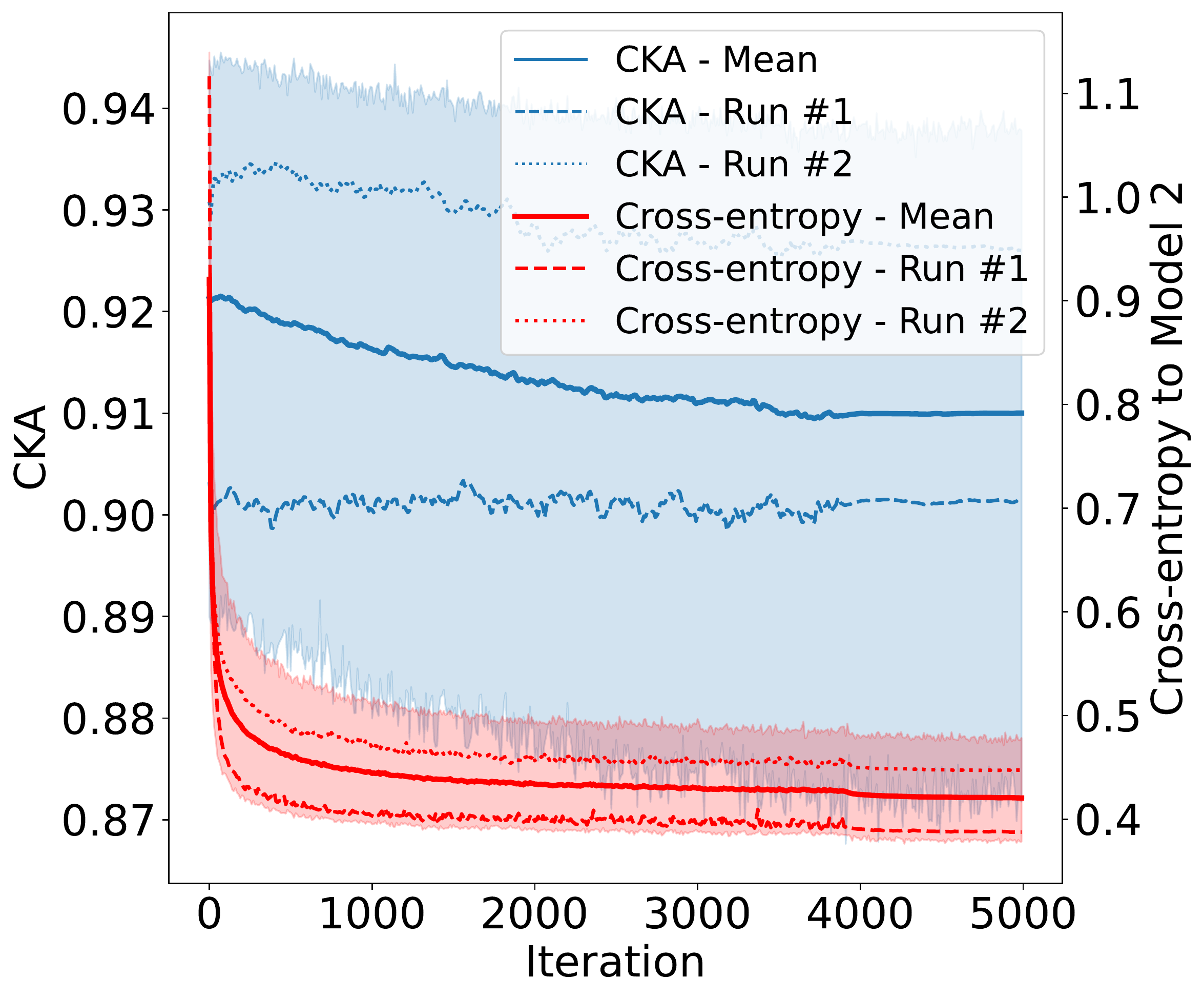}
        \includegraphics[width=0.50\linewidth]{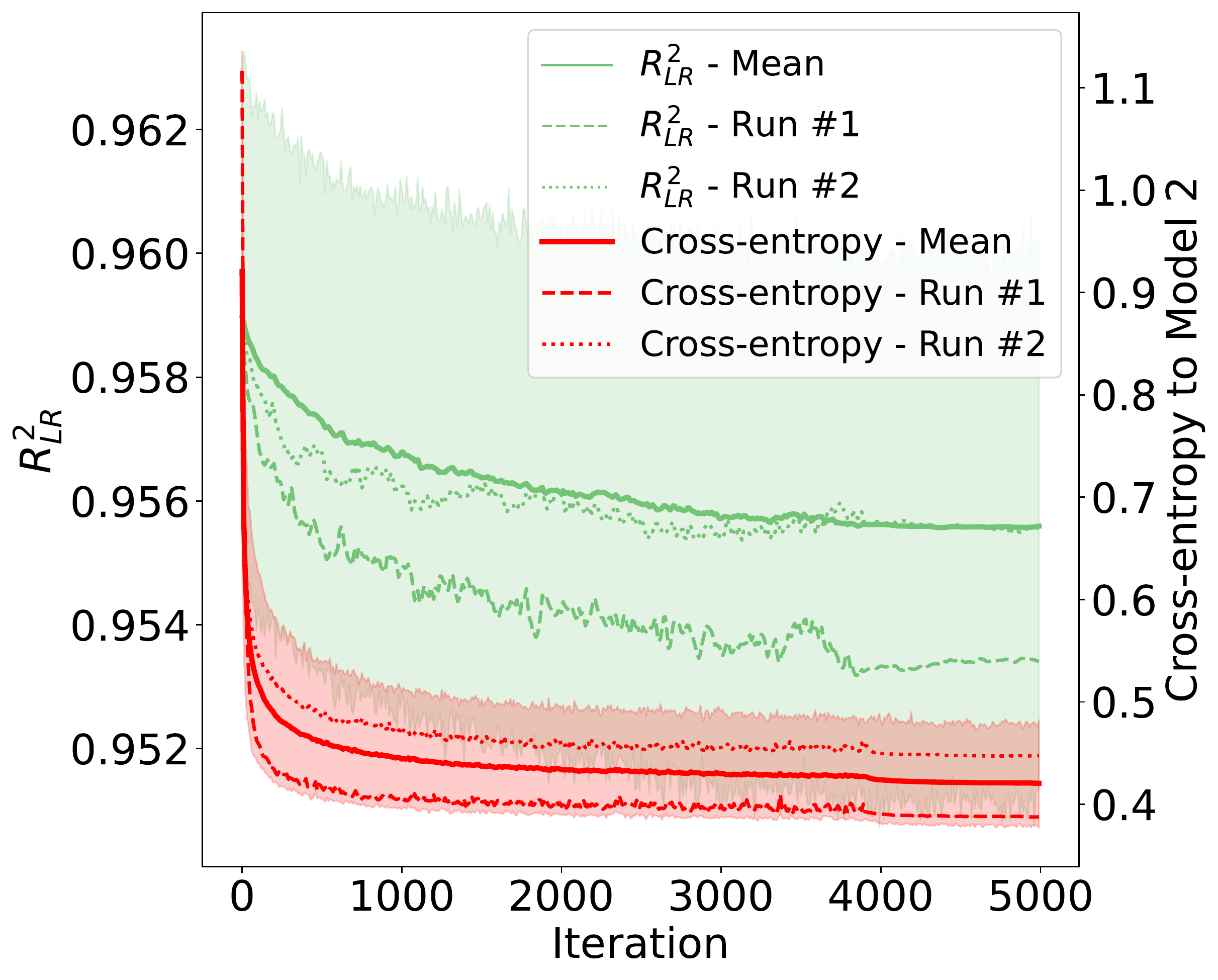}
        \caption{CKA, $\RLR$ and cross-entropy values over the training iterations of task loss matching started from the optimal least squares matching. Tiny-10 network, stitching at Layer 3. Bold lines are averages of 10 runs, with bands representing standard deviations; dotted and dashed lines correspond to two specific runs to show the individual characteristics of the training of the stitching layer. Further experiment details are in Appendix~\ref{appendix:exp_sim_ind}.}
        \label{fig:similarity_vs_frankenstein}
    \end{minipage}
    \quad
    \begin{minipage}[t]{0.33\textwidth}
        \centering
        \includegraphics[width=0.80\linewidth]{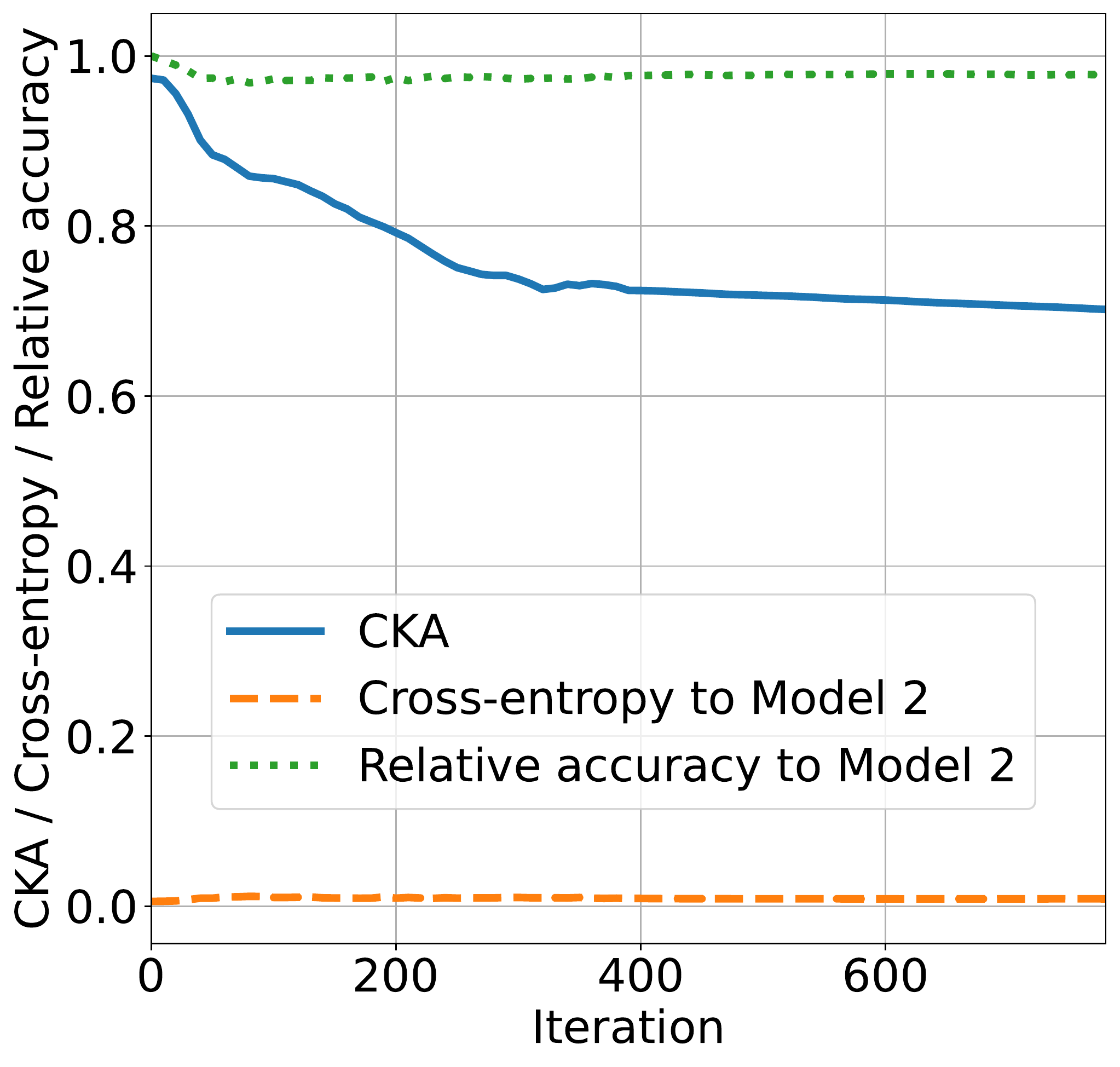}
        \caption{CKA and performance measures throughout the training for a Frankenstein task with a loss that explicitly penalizes CKA value while maintaining the model performance. Stitching is at Layer 2.0 of a 3-wide ResNet.}
        \label{fig:ruin_cka}
    \end{minipage}
\vspace{-2ex}
\end{figure}

In this set of experiments, we monitor the $\CKA$, $\RLR$, $\CCA$ and $\SVCCA$ indices throughout the training process of the stitching layer trained with task loss matching. We measure these values between Model 2 activations and the transformed activations of Model 1. Figure \ref{fig:similarity_vs_frankenstein} shows the results for CKA and $\RLR$. We observe that $\CKA$ has a large variance from the first step of the training process, and it is decreasing while the performance of the stitched model is increasing.

This experiment also highlights that $\RLR$ can correlate inversely with the performance: as the training proceeds with decreasing cross-entropy, the $\RLR$ values also decrease. Indeed, since we initialized the task loss training from the optimal least squares matching, the distances naturally must increase under any training process. However, this highlights the fact that the least squares matching is ignorant of the decision boundaries in a neural network.

$\CCA$ is invariant to invertible linear transformations, hence it ignores our linear stitching transformation during the training. $\SVCCA$ also performs $\CCA$ on truncated singular value decompositions of its inputs. Because its value only depends on the SVD step, it performs quite big jumps during the training, but on average it does not correlate with the task performance. See further results and experimental details in Appendix~\ref{appendix:exp_sim_ind}.

\vspace{-1ex}

\subsection{Experiment 4: Low CKA value with high accuracy}
\label{subsec:lowcka}

We now show that it is easy to align representations in a way that results in high task performance while the CKA value at the alignment is small. The setup is the following: we take one trained model and stitch it with \emph{itself}, utilizing a linear stitching layer initialised with the identity mapping. With this initialization we essentially reproduce the original model, thus, this trivially results in a Frankenstein network with the same performance as the original model. We train this stitching layer with an additional loss term that penalizes the CKA value at the stitching, meanwhile we maintain the task performance with the usual cross-entropy term. We use the CKA value itself as the penalty, as it is differentiable by definition and thus suitable for gradient-based optimization.

Figure~\ref{fig:ruin_cka} shows the outcome of such a training process. We find that the CKA value drops by a large amount (achieving lower than $0.7$), whereas the accuracy remains essentially unchanged.

As a baseline, we provide the CKA values between representations of the transformed activations of Model 1 and Model 2 averaged over 10 different network pairs and all layers. ResNet: $0.922\pm 0.098$, Tiny-10: $0.883\pm 0.050$. For further baseline CKA values we also refer to \citet{kornblith2019similarity}.

\vspace{-1ex}

\subsection{Discussion}

Notions of representational similarity and functional similarity have different motivations, and thus, aim to capture different invariances. Part of these invariances could be compatible or incompatible. For example, assuming that a similarity index $s$ is invariant under a given class $\mathcal{C}$ of transformations, it is not surprising that the high value of a similarity index $s(A,B)$ does not indicate the relative performance of the activations $A$ and $B$: indeed, given a set of activations $A$, $s(AF,A)=1$ for any $F\in \mathcal{C}$, while the task loss of the transformed activations $AF$ typically varies greatly.  On the other hand, it is somewhat surprising that high-performance matchings can have a significant drop in the values of meaningful similarity indices such as CKA.

\paragraph{The inadequateness of CKA as a direct matching method} CKA is not directly suitable as an objective for direct matching, since it is invariant to orthogonal transformations. A stitching with an optimal CKA value can be arbitrarily transformed via orthogonal transformations without affecting CKA, but most of these equivalent solutions are unsuitable for the matching task as trained models are not invariant for arbitrary rotations in the activation spaces.

\section{Low rank representations --- a gap between SVD and task loss matching}
\label{section:lowrank}

In this experiment, we compare the effectiveness of the low-rank analogues of the above presented least squares matching and task loss matching. More precisely, we apply SVD directly in the activation space to obtain the low-rank least squares matching, and use a \emph{bottleneck} stitching layer to realise a low-rank matching with the task loss.

Figure~\ref{fig:low_rank} depicts the results for two selected layers of Tiny-10 with prescribed ranks 4, 8, and 16. (We observed similar results for the other layers, and also for ResNet-20.) It is clear that task loss matching outperforms least squares matching in the activation space when the rank of the transformation is fixed. This suggests that the objective for the low-rank approximation of explaining most of the variance in the activation space (i.e., the objective of SVD) is not satisfactory to attain the highest possible performance with a prescribed rank. In other words, the magnitudes of the principal components are not in direct correspondence with the information content of the latent directions. Ultimately, this suggests that similarity indices that use SVD directly in the activation space --- either as a preprocessing step or as another key element in the computation --- might fail to capture some desired aspects of the representation. Details and figures for the other layers are in Appendix~\ref{appendix:experiment_details}.

\section{Structure of stitching layers: uniqueness, mode connectivity, sparsity}
\label{section:properties}

The previous experiments show that the stitching layer can be trained to achieve the same performance as the original Model 2. What is the structure of the obtained stitching matrix? In this chapter, we explore the question of sensitivity with respect to initialization, uniqueness properties of the stitching layer and the relationship of direct and task loss matching with sparsity.

\paragraph{Experiment 5: Dependence of accuracy on initializations}
\label{subsec:initexperiment}
Are stitching layers obtained from different initializations the same? A crude measure of this is the performance of the stitching layer on the stitched network. In this experiment, we fix two identical network architectures trained from different initializations. We train the stitching layer on task loss from 50 random and 50 optimal least squares initializations, and compare their performance.
We found that in most cases the stitched networks achieve the same order of accuracy, although not always; the training process might get stuck at poor local minima. However, the optimal least squares matching initialization performs consistently well. Figure~\ref{fig:init} shows a histogram of the results. For further details, see Appendix~\ref{appendix:experiment_details}.

\begin{figure}[t!]
\centering
    \begin{minipage}[t]{0.3\textwidth}
        \includegraphics[width=0.9\linewidth]{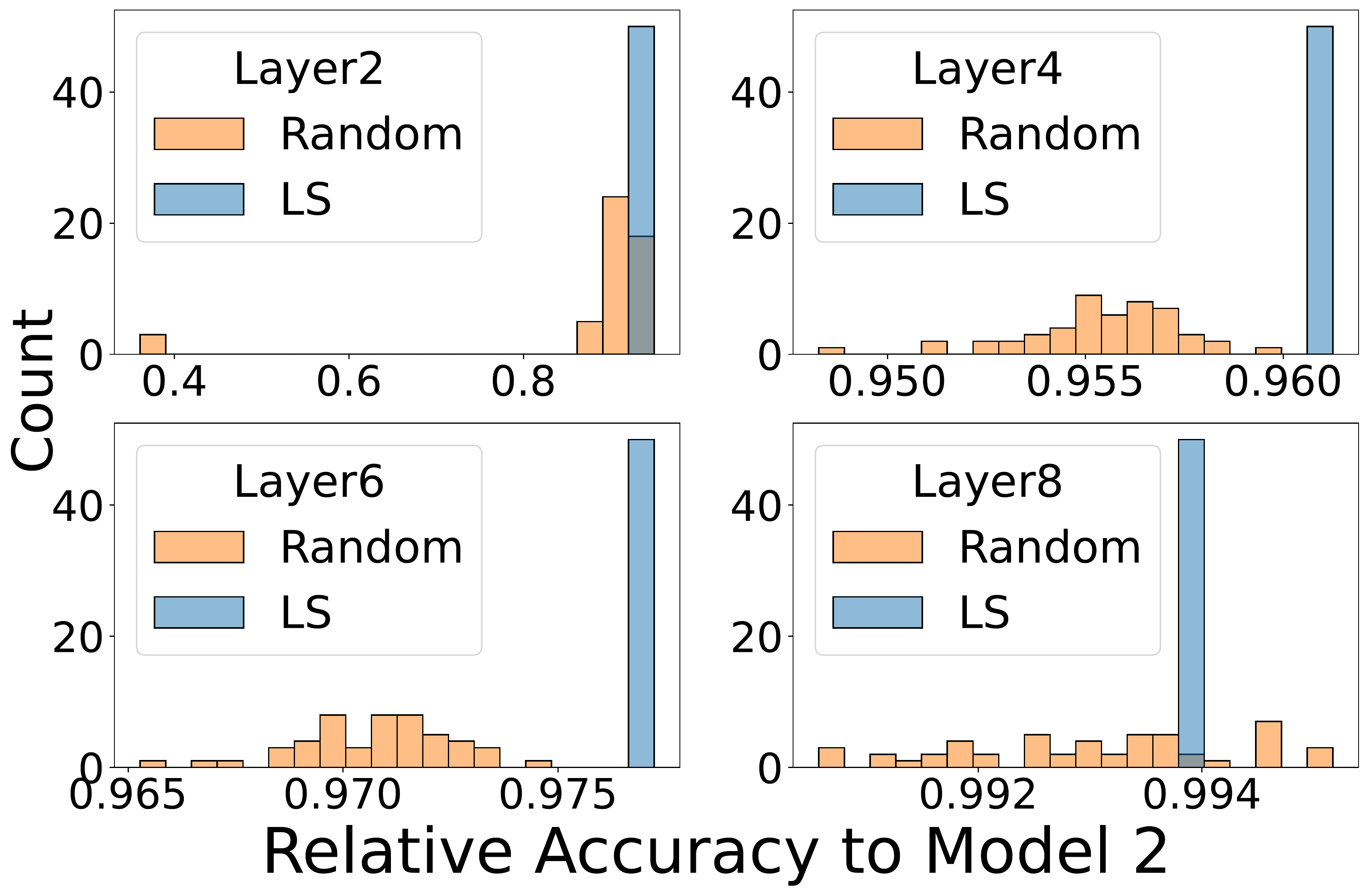}
        \caption{Relative accuracies by initialization type.}
        \label{fig:init}
    \end{minipage}
    \quad
    \begin{minipage}[t]{0.66\textwidth}
        \centering
        \includegraphics[width=0.45\linewidth]{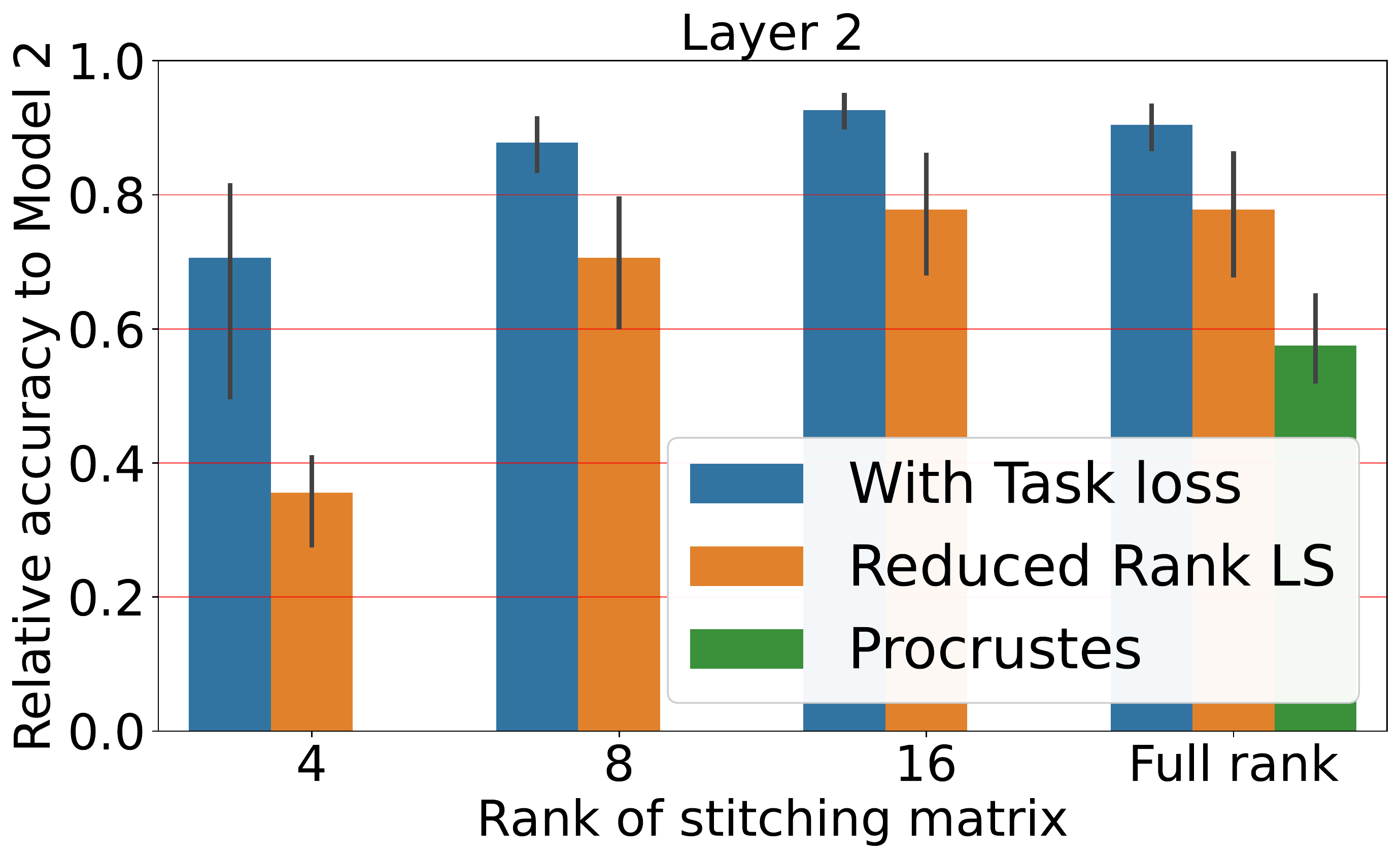}
        \includegraphics[width=0.45\linewidth]{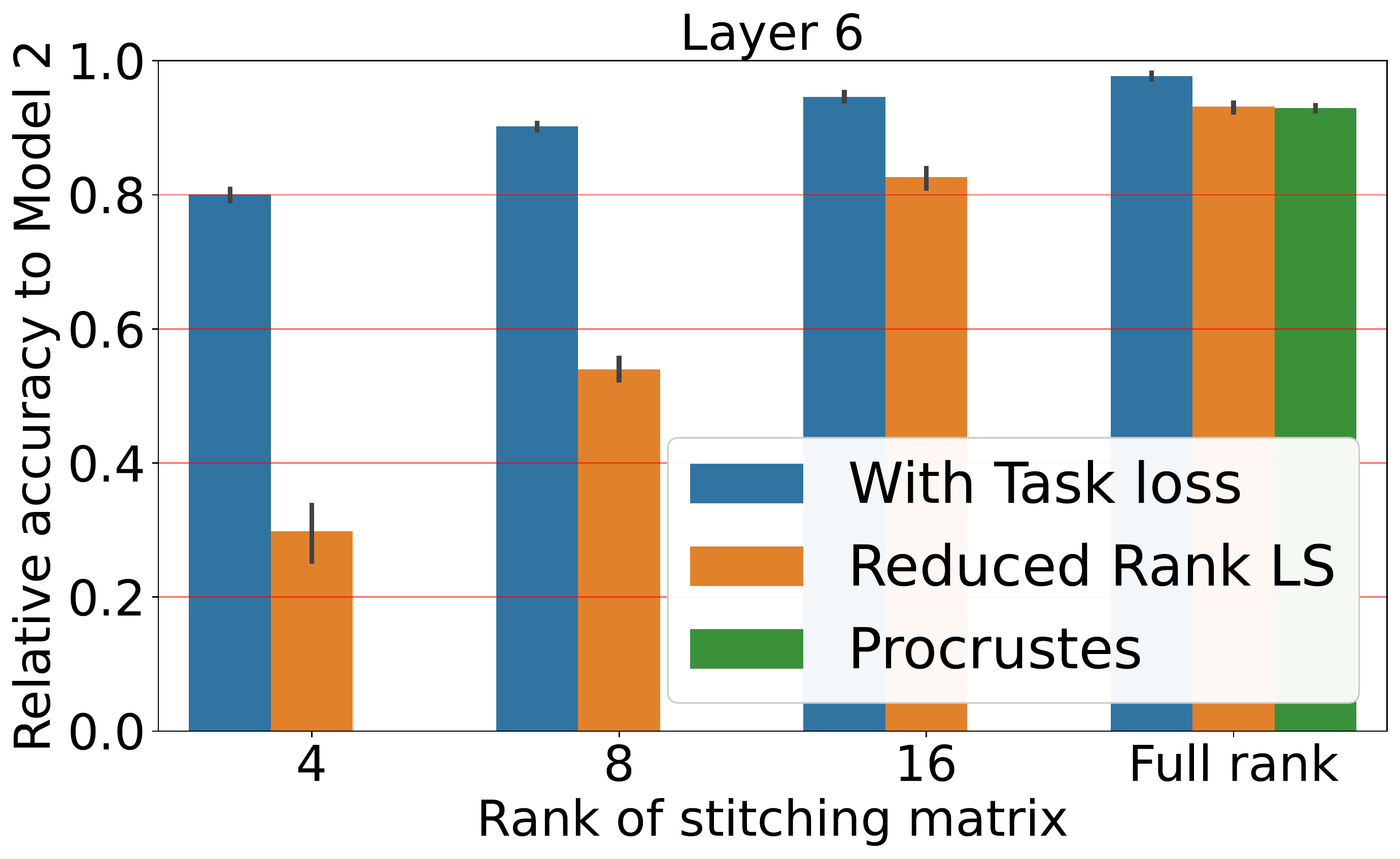}
        \caption{The performance of the low-rank analogues of least squares and task loss matchings in terms of relative accuracy.}
        \label{fig:low_rank}
    \end{minipage}
\vspace{-2ex}
\end{figure}

\paragraph{Experiment 6: Linear mode connectivity of transformation matrices}
\label{subsec:uniqueness}
We also investigated the uniqueness properties of the stitching layer by examining the linear mode connectivity of the found transformations. In particular, on a given layer of two trained copies of Tiny-10, we trained stitching transformations from different random initializations. We only kept the transformations achieving a relative accuracy of at least 80\% --- there is no reason to expect mode connectivity properties where there is a large difference in relative accuracy. We then evaluated the relative accuracy of the stitched network with the transformation matrix $M_{\lambda}=\lambda M_1 +(1-\lambda) M_{2}$ for pairs of transformation matrices $M_1$ and $M_2$, and different values of $\lambda\in [0,1]$. 
The results showed that mode connectivity holds on most layers, 
i.e.\ the convex combination of the stitching layers perform similarly. 
However, on the first three layers, we noticed an unforeseen lack of mode connectivity, even though the relative accuracies of the corresponding transformations were high. For further details, see Appendix \ref{appendix:exp6_details}.

\paragraph{Experiment 7: Sparsity --- direct vs.\ task loss matching}
\label{subsec:sparsematching}

\begin{figure}[t!]
\centering
\begin{subfigure}[t]{.60\textwidth}
    \centering
    \includegraphics[height=90pt]{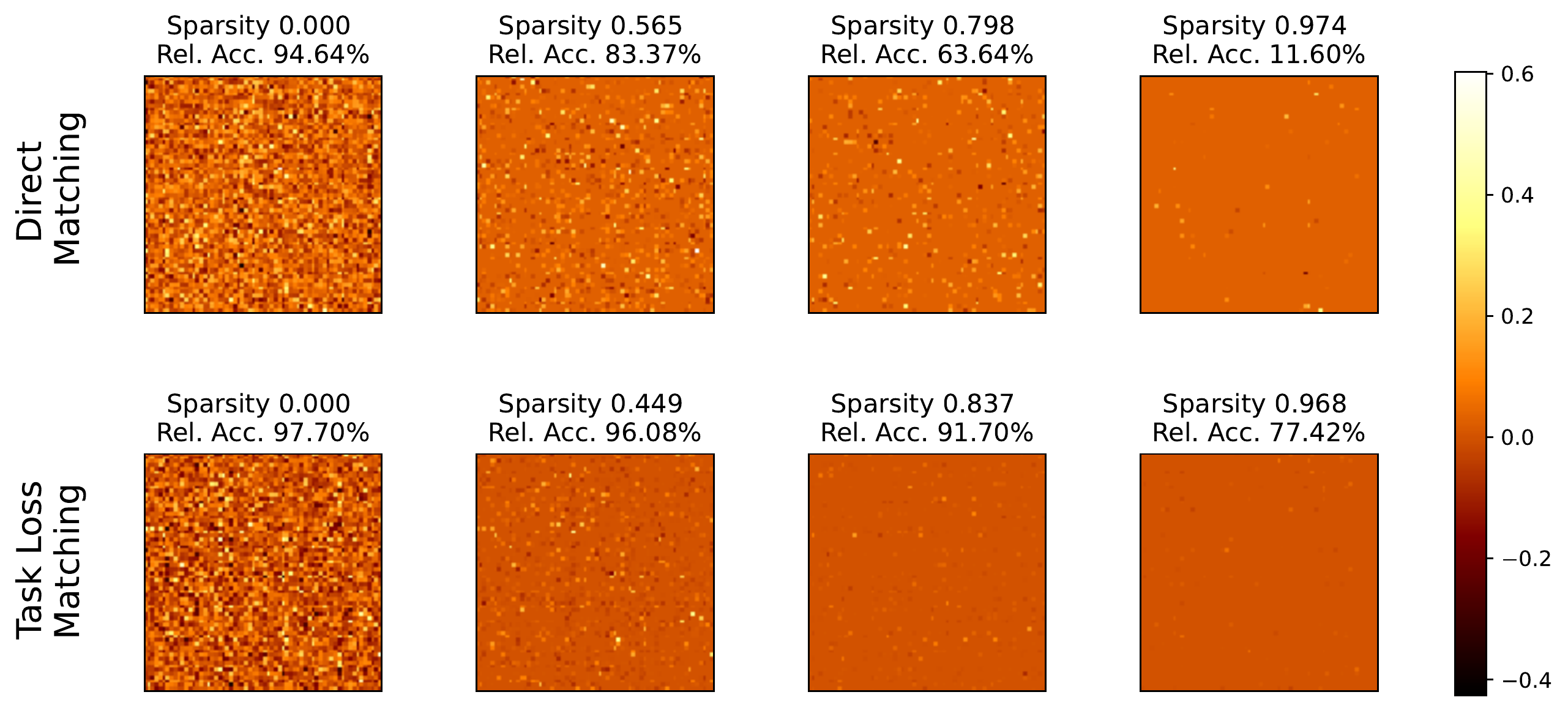}
\end{subfigure}
\quad
\begin{subfigure}[t]{.30\textwidth}
  \centering
  \includegraphics[height=90pt]{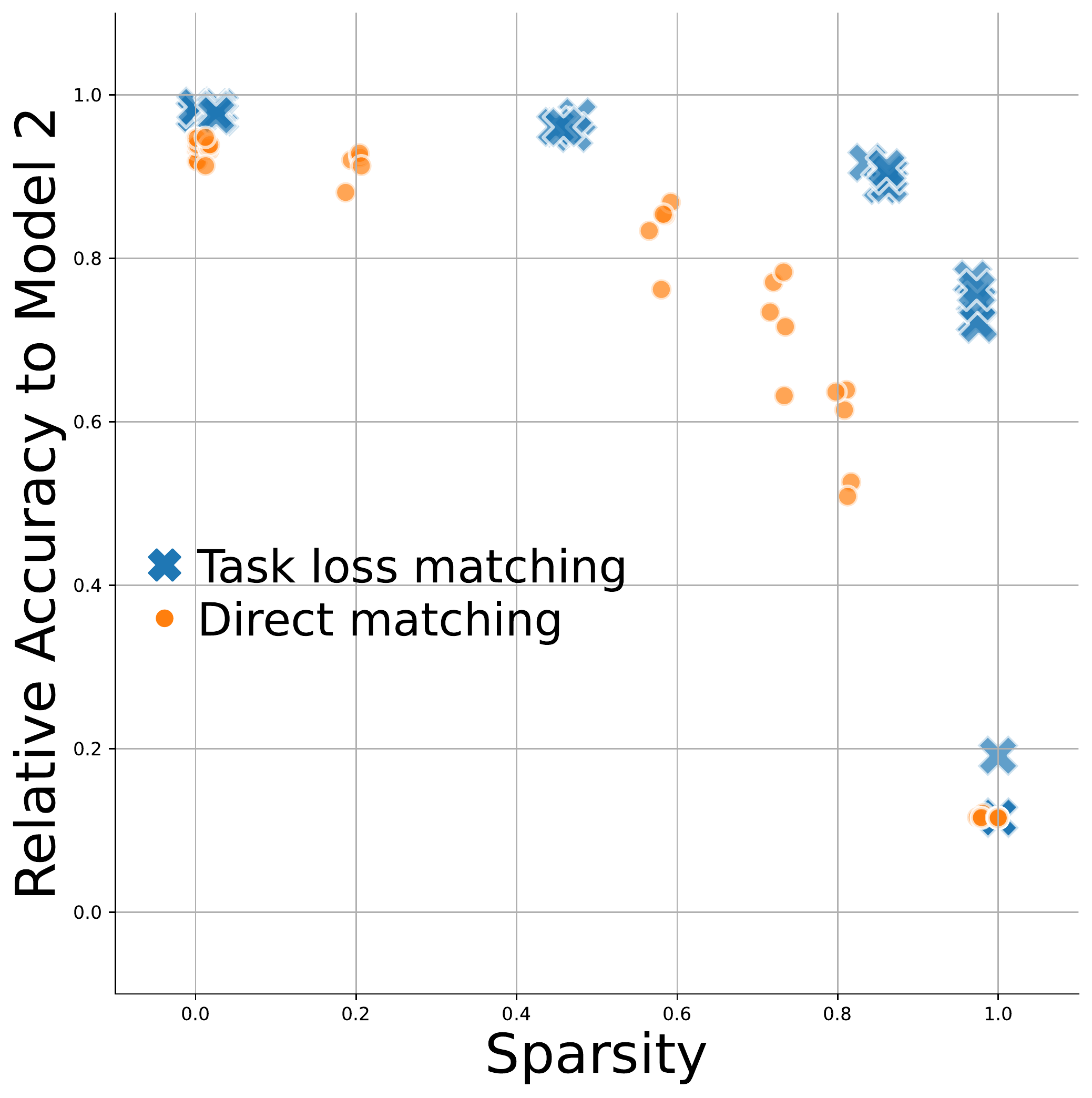}
\end{subfigure}
\caption{Comparison of stitching matrices and the relative accuracy of L1-regularized direct and task loss matching. To achieve sparsity, we set every entry of the stitching matrix to zero when the absolute value of the entry is below the threshold $10^{-4}$. Sparsity is the ratio between the zero elements and the total number of elements in the stitching matrix.
Relative accuracies with zero sparsity correspond to the unregularised baseline.
\textbf{Left:} Examples of stitching matrices with different sparsity. \textbf{Right:} The degradation of relative accuracy with respect to increasing sparsity is significantly less steep with task loss training. For more results, see Appendix~\ref{appendix:sparsity}.}
\label{fig:l1}
\vspace{-2ex}
\end{figure}

How does the sparsity of the stitching layer influence the performance of the stitched network? In this experiment, we achieve sparsity by adding an L1-regularizing term to the training as detailed in Section \ref{subsec:sparse}, similarly to the experiments in \cite{li_2016_ICLR}. We perform both an L1-regularized task loss matching as well as a direct loss matching with loss term \eqref{eq:lasso}. To obtain a truly sparse stitching matrix we set the elements to zero below a threshold. Sparsity is then computed as the ratio between the zero elements and the total number of elements in the matrix. Figure~\ref{fig:l1} shows the dependence of relative accuracy on sparsity compared for task loss and direct matching as well as a sample of the stitching matrices. As Figure~\ref{fig:l1} shows, at around 80-90 $\%$ sparsity, the relative accuracy of direct matching suffers a drastic drop, however task loss matching still manages to find a meaningful alignment of the neurons using additional information via the task loss. These experiments show that even though it is possible to achieve a certain degree of sparse alignment of the neurons, the relationship between two representations is typically not of a combinatorial nature, and much can be gained by considering more complex relationships between the two layers. Furthermore, the difference in the relative accuracy of direct and task loss matching is notable for higher values of sparsity. For further details, see Appendix \ref{appendix:sparsity}.

\vspace{-0.5em}
\section{Limitations}
\label{section:limitations}

The performance of the stitched solution is inherently just a lower bound due to possible local minima. The current study is restricted to well-established convolutional architectures. Our observations on network width is analysed only in isolation.

\vspace{-0.5em}
\section{Conclusions}

We studied similarity of neural network representations in conjunction with the notion of matchability. 
We demonstrated the matchability of representations in convolutional networks through several experiments and analysed the properties of the stitching transformations with respect to the task performance.
Regarding similarity notions, we provided a novel perspective which incorporates the performance on a task; a perspective which is yet unexplored in the field. We pointed out weaknesses of popular similarity notions that end-users might encounter.

\vspace{-0.5em}
\section*{Broader impact}

Our research is of foundational nature and advances machine learning at a fairly general level. While the authors do not foresee that the work herein will have immediate adverse ethical or societal consequences, such analysis of representation learning could help a malicious user avoid detection that is based on representational similarity; on the other hand, it could move forward the field towards more transparent and interpretable machine learning models.
\section*{Acknowledgements}

This work was supported by the European Union, co-financed by the European Social Fund (EFOP-3.6.3-VEKOP-16-2017-00002), the Hungarian National Excellence Grant 2018-1.2.1-NKP-00008 and by the Hungarian Ministry of Innovation and Technology NRDI Office within the framework of the Artificial Intelligence National Laboratory Program. We thank the anonymous reviewers for their many valuable suggestions.

\bibliography{references}
\bibliographystyle{plainnat}

\appendix
\section{Experiment details}
\label{appendix:experiment_details}

\subsection{Datasets, splits, preprocessing, data augmentation}

\paragraph{CIFAR-10} The CIFAR-10 dataset \cite{cifar10} consists of 60000 32x32 colour images in 10 classes, with 6000 images per class. There are 50000 training images and 10000 test images. We used the canonical train--test split. As a preprocessing, we normalized the images with the means (0.4914, 0.4822, 0.4465) and standard deviations (0.2023, 0.1994, 0.2010) for the three RGB channels, respectively. As augmentation, we used random horizontal flip, and 32x32 sized random crop from the zero padded 40x40 inputs.

\paragraph{CelebA} The CelebA dataset \cite{liu2015faceattributes} is a large-scale face attributes dataset, it consists of 202599 number of images depicting faces of celebrities, each with 40 attribute annotations. We used a random split with 80\% train and 20\% test set sizes. Originally the CelebA images are sized 178$\times$218 (width $\times$ height). As a preprocessing, we first reshaped these images to 256x256 pixels and then applied a center crop to 224x224 which was our final input shape for the Inception V1 networks. We subtracted 117 from each pixel. As augmentation, we applied a random rotation between degrees of -10 and 10.

\subsection{Network architectures and training details}

\paragraph{Tiny-10} Tiny-10 is a simple multi-layer convnet architecture. We use this model to provide a simple convnet for the experiments that trains fast on modern hardware. (A similar architecture was used also in \cite{kornblith2019similarity}.) Table~\ref{table:tiny-10} details the layers along with the names we used in the paper to refer to a particular part of the network. While this naming is ambiguous as it could refer to three different activation spaces (in a row of the table), we use the activations after the batch normalization and before the nonlinearity if not otherwise stated.

We trained the model on CIFAR-10 for 300 epochs, the optimizer was SGD with Nesterov momentum $0.9$. There was a schedule for the learning rate: started with the value of $0.1$ and it was divided by 10 at 1/3 of the training, and with another 10 at the 2/3 of the training. The batch size was $128$. We used weight decay with value $10^{-4}$. The average accuracy of the resulting models was 86.55\%.

\begin{table}[h]
\caption{The Tiny10 architecture.}
\label{table:tiny-10}
\centering
\begin{tabular}{l l}
\toprule
\multicolumn{1}{c}{Tiny10}\\
\midrule
Layers & Name \\
\midrule
3 × 3 conv. 16-BN-ReLu & Layer 1 \\
3 × 3 conv. 16-BN-ReLu  & Layer 2 \\
3 × 3 conv. 32 stride 2-BN-ReLu & Layer 3\\
3 × 3 conv. 32-BN-ReLu  & Layer 4 \\
3 × 3 conv. 32-BN-ReLu   & Layer 5 \\
3 × 3 conv. 64 stride 2-BN-ReLu  & Layer 6 \\
3 × 3 conv. 64 BN-ReLu  & Layer 7 \\
1 × 1 conv. 64-BN-ReLu  & Layer 8 \\
Global average pooling  &  \\
Dense & \\
Logits & \\
\bottomrule
\end{tabular}
\end{table}

\paragraph{ResNet-20} Our ResNet-20 \cite{resnet} variant follows common practices regarding CIFAR-10: we use a 3-level architecture with three residual blocks per level. A residual block contains the following layers: Conv-Batchnorm-ReLU-Conv-Batchnorm. After each residual block, a ReLU operation follows the addition operation. Convolution kernels are sized 3x3. In the paper, we use the following naming convention: \texttt{LayerX.Y} corresponds to activations after the addition operation following a residual block with index \texttt{Y} in level \texttt{X}, with the exception of \texttt{Layer0.0}, which corresponds to the activation space after the first Conv-Batchnorm layer in the network preceding the residual blocks.

We trained the model on CIFAR-10 for 300 epochs, the optimizer was SGD with Nesterov momentum $0.9$. There was a schedule for the learning rate: started with the value of $0.1$ and it was divided by 10 at 1/3 of the training, and with another 10 at the 2/3 of the training. The batch size was $128$. We used weight decay with value $10^{-4}$. The average accuracies were 91.95\%, 93.97\%, 94.59\%, 94.77\% for the 1-width, 2-width, 3-width, and 4-width ResNets, respectively.

\paragraph{Inception V1} For the Inception V1 \cite{inceptionv1}, we omit the detailed description of the architecture as a reiteration would be cumbersome, and there are no specifics for the task at hand. Regarding the layer names, we follow the standard naming conventions.

We trained the model on CelebA for 20 epochs, using the Adam optimizer   with parameters $\beta_1=0.9$ and $\beta_2=0.999$. Learning rate was $0.0001$, batch size was $128$.

\subsection{Experiment 1 - Details for matching with least squares and task loss matching}
\label{appendix:exp1_details}

In this experiment, we take network pairs (Model 1 and Model 2) which are of the same architecture but trained from different weight initializations and with different orderings of the training set.

\paragraph{Least squares matching} Let $A$ and $B$ denote the activation matrices for the training data of Model 1 and Model 2, respectively. We appended an all ones vector to the activation matrix $A$ of Model 1 to represent the bias. Then we calculated the pseudoinverse $A^\dagger$ of $A$ using SVD. The transformation matrix and the bias of the stitching layer was obtained by calculating $A^{\dagger}B$.

\paragraph{Task loss matching} We initialized the transformation matrix and the bias of the stitching layer to the least squares solution (which was calculated as described above). Then we trained the stitching layer on the train set for $30$ epochs. The utilized loss was cross-entropy to Model 2 activations. The optimizer was Adam with parameters $\beta_1=0.9$ and $\beta_2=0.999$, learning rate was set to $10^{-3}$, batch size was $128$.

We used the following hyperparameter selection protocol: after a grid search with Tiny-10 and ResNet consisting of the parameter settings \{ Optimizer: Adam, SGD \} $\times$ \{ Learning rate: 0.1, 0.01, 0.001, 0.0001, 0.00001 \} and training for 300 epochs, we observed that Adam is significantly better than SGD, and that learning rates below 0.001 do not affect performance, only prolong the training time. Moreover, we observed that with the selected hyperparameters the training of the stitching layer always reaches a plateau before the 30th epoch, thus, we set this hyperparameter accordingly.

On Tiny-10, we matched the activations after the batch normalization layer (which comes after the convolution and before the nonlinearity). With ResNets, we plotted the results that correspond to stitchings after the addition operations of the residual blocks. (Matchings on the inside layers of residual blocks are harder to interpret, while the results are very similar).

\subsection{Experiment 2 - Details for networks of different width}
\label{appendix:exp2_details}

We utilized the same methodology and settings to train the stitching layer as described in Appendix~\ref{appendix:exp1_details}. The baseline 3-level ResNet had 16, 32, and 64 filters in the convolution layers for each level, respectively. The networks of different width were obtained by multiplying these baseline filter numbers with the width multiplier.

\subsection{Experiment 3 - Details for similarity indices and task performance}
\label{appendix:exp_sim_ind}

We utilized the same methodology and settings to set or train the stitching layer as described in Appendix~\ref{appendix:exp1_details}.

$\CKA$, $\CCA$ and $\SVCCA$ is calculated on the whole validation set. We can fit this amount of data into memory without needing to resort to sampling methods like “minibatch CKA” \citet{nguyen2021do}. We note that the resulting CKA values are indistinguishable from values obtained when working with any number of data points between 2500 and 10000.

\begin{figure}[t!]
    \centering
    \includegraphics[width=0.49\linewidth]{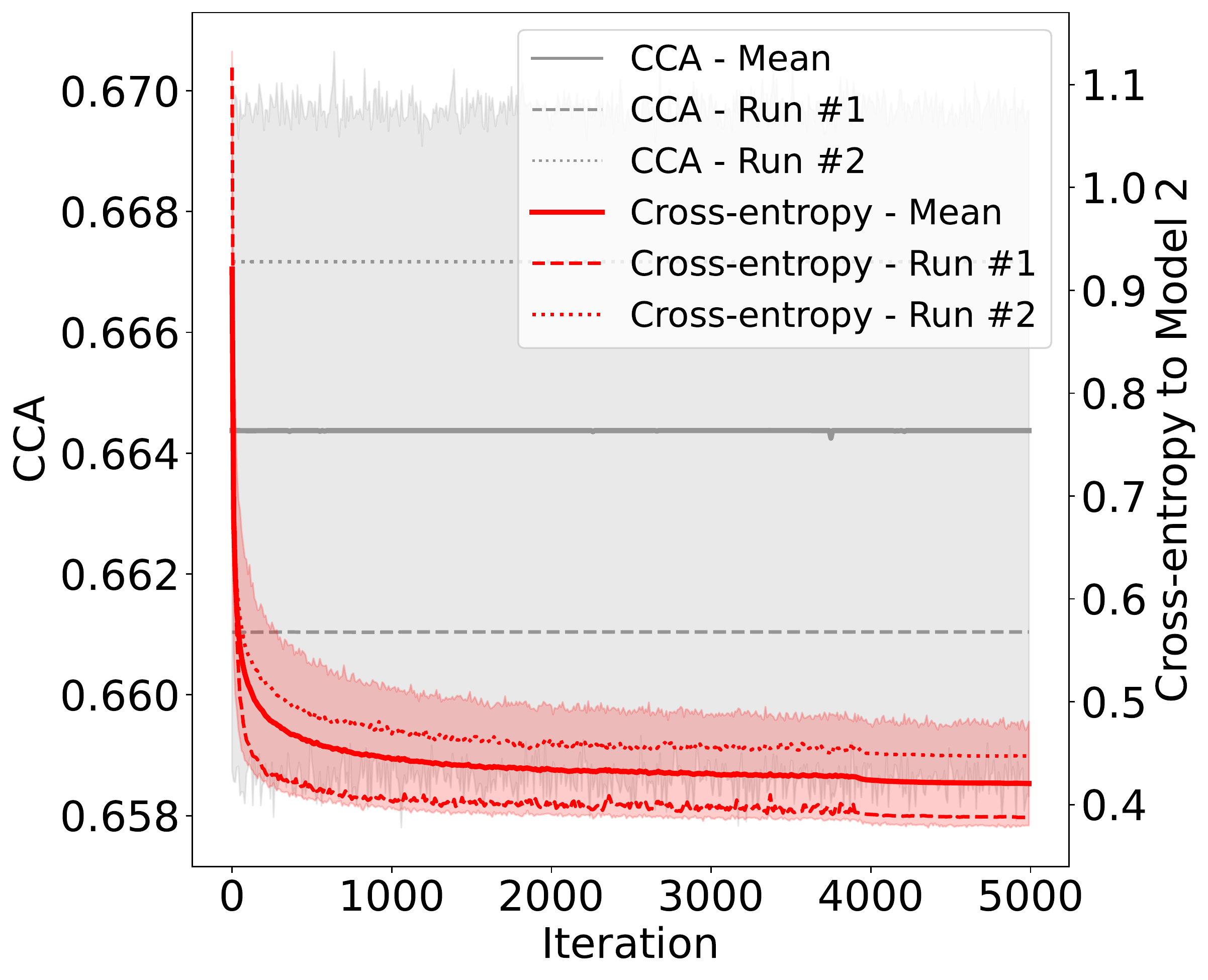}
    \quad
    \includegraphics[width=0.47\linewidth]{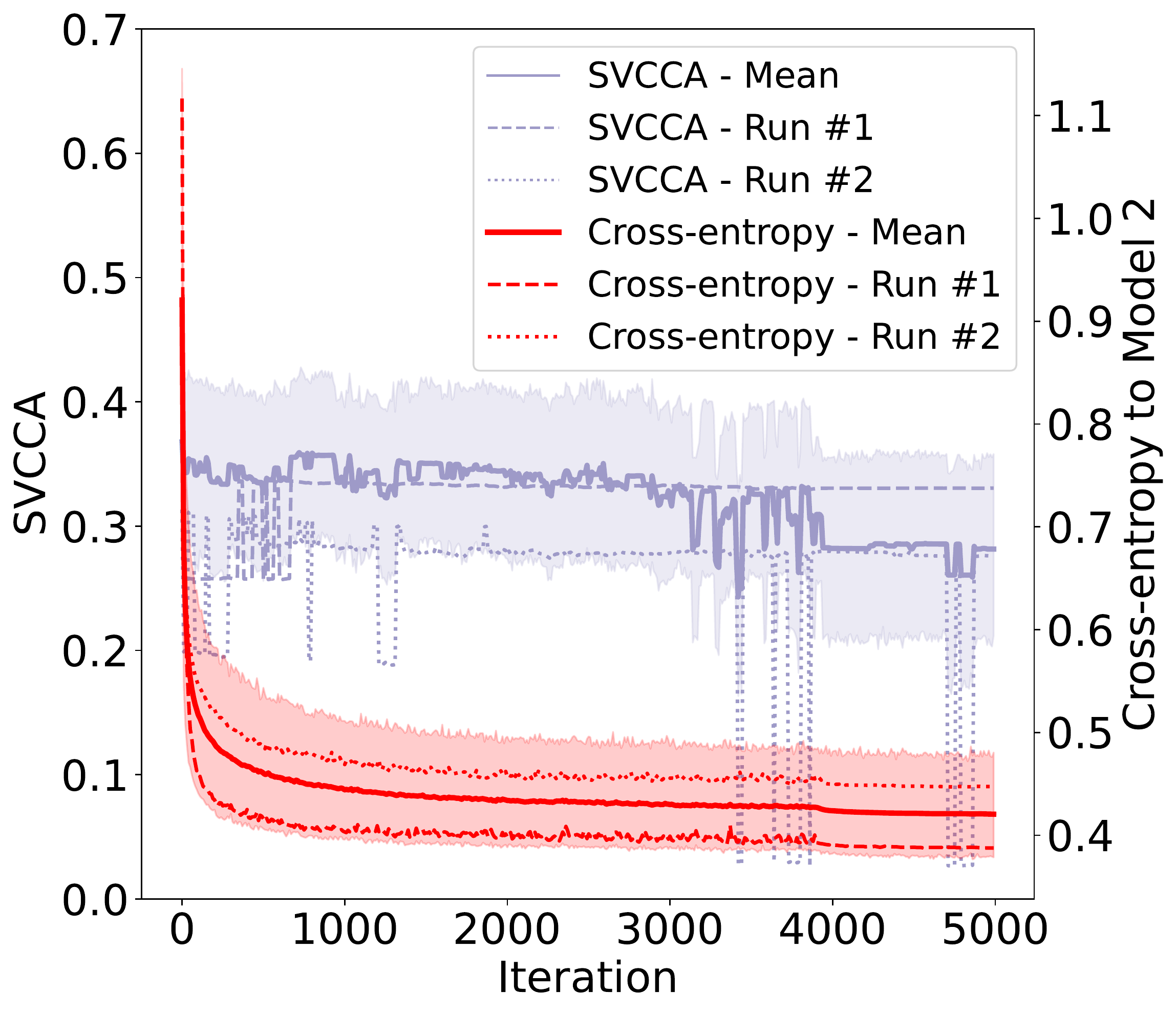}
    \caption{$\CCA$, $\SVCCA$ and cross-entropy values over the training iterations of task loss matching started from the optimal Least Squares matching. Tiny-10 network, stitching at Layer 3. Bold lines are averages of 10 runs, with bands representing standard deviations; dotted and dashed lines correspond to two specific runs to show the individual characteristics of the training of the stitching layer.}
    \label{fig:cca_vs_frankenstein}
\vspace{-1ex}
\end{figure}

\subsection{Experiment 4 - Details for low CKA value with high accuracy}
\label{appendix:exp_low_cka}

The loss penalizing CKA was calculated on the training minibatch, the reported values were calculated on the whole validation set. We trained the stitching layer for 2 epochs with batch size 128, CKA loss weight $0.1$, the optimizer was Adam with parameters $\beta_1=0.9$ and $\beta_2=0.999$, learning rate was $0.001$ at the first epoch and $0.0001$ in the second epoch. (Nevertheless, we observed that the general outcome of the experiment is quite robust for a large range of hyperparameters.)

\subsection{Experiment 5: Details for dependence of accuracy on initializations}

We train the stitching layer on task loss from 50 random and 50 optimal least squares initializations between two Tiny-10 architectures trained from different initializations. Other settings of the experiment are the same as presented in Appendix~\ref{appendix:exp1_details}.  Figure \ref{fig:init_all} shows the results for all layers, between three pairs of networks.

\begin{figure}[t]
    \centering
    \includegraphics[width=1.0\linewidth]{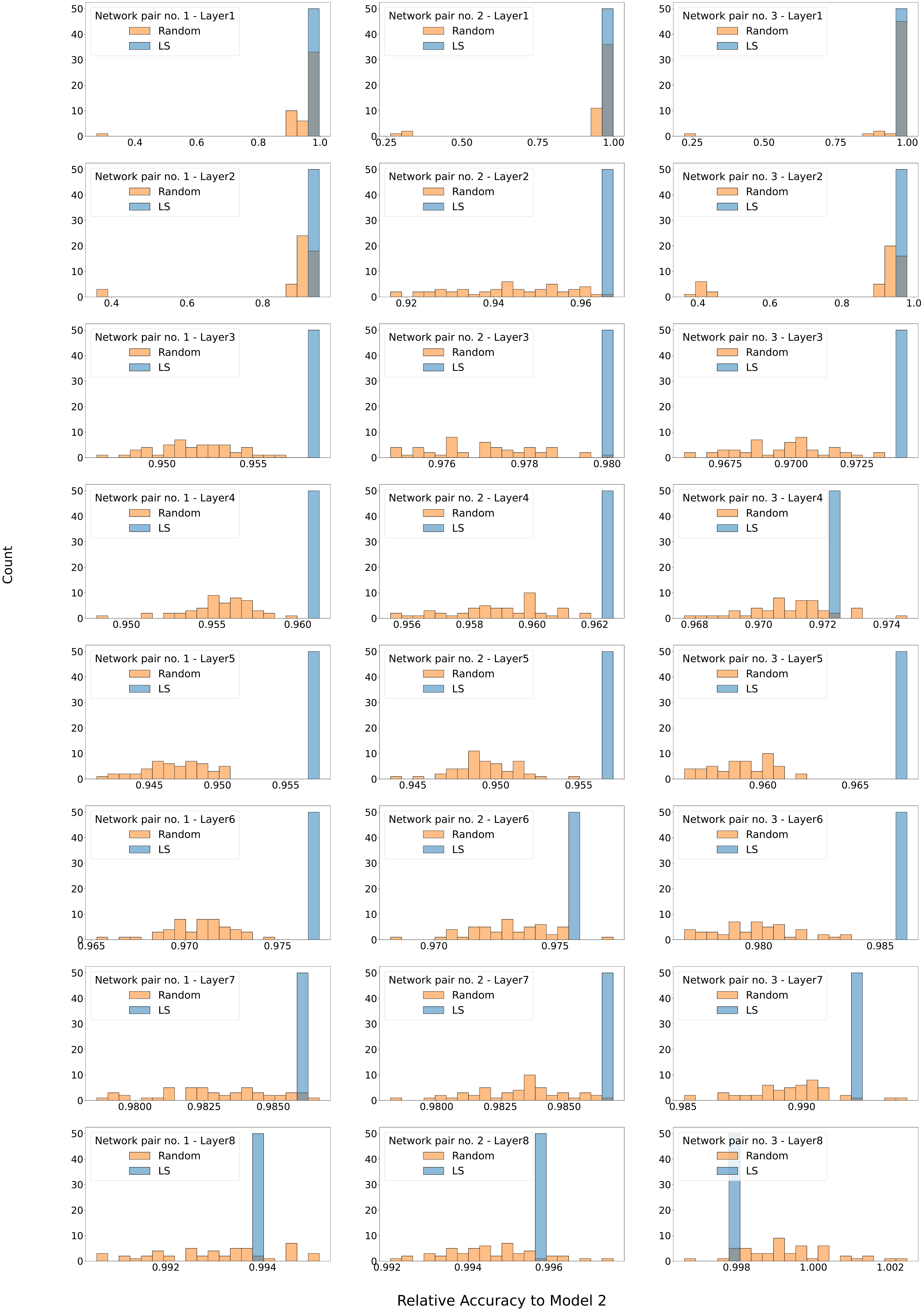}
    \caption{Comparing matching performance of randomly initialized transformation matrices versus transformation matrices initialized with the least squares solution. Different plots correspond to matchings on different layers, and different network pairs.}
    \label{fig:init_all}
\end{figure}

\label{appendix:exp5_details}
\subsection{Experiment 6: Details for linear mode connectivity of transformation matrices}
\label{appendix:exp6_details}
We utilized the same methodology and settings to train the stitching layers as described in Appendix~\ref{appendix:exp1_details}. See Figure~\ref{fig:mode_connect} for detailed results.

\begin{figure}[t]
    \centering
    \begin{subfigure}[t]{1.0\textwidth}
      \centering
        \includegraphics[width=1.0\linewidth]{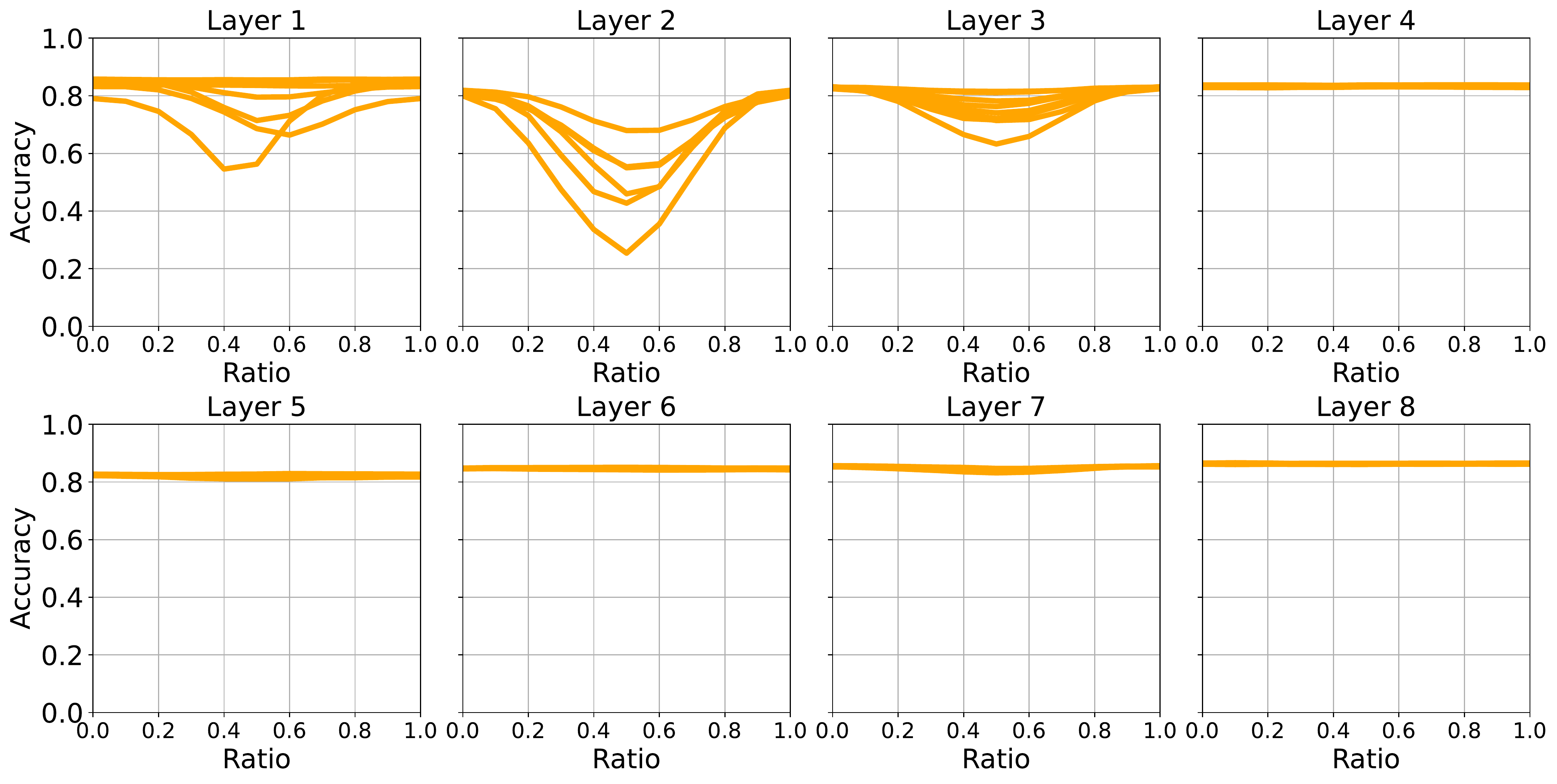}
        \caption{Linear mode connectivity of stitched networks trained from different random initializations.}
    \end{subfigure}
    
    \begin{subfigure}[t]{1.0\textwidth}
      \centering
      \vspace{40pt}
        \includegraphics[width=1.0\linewidth]{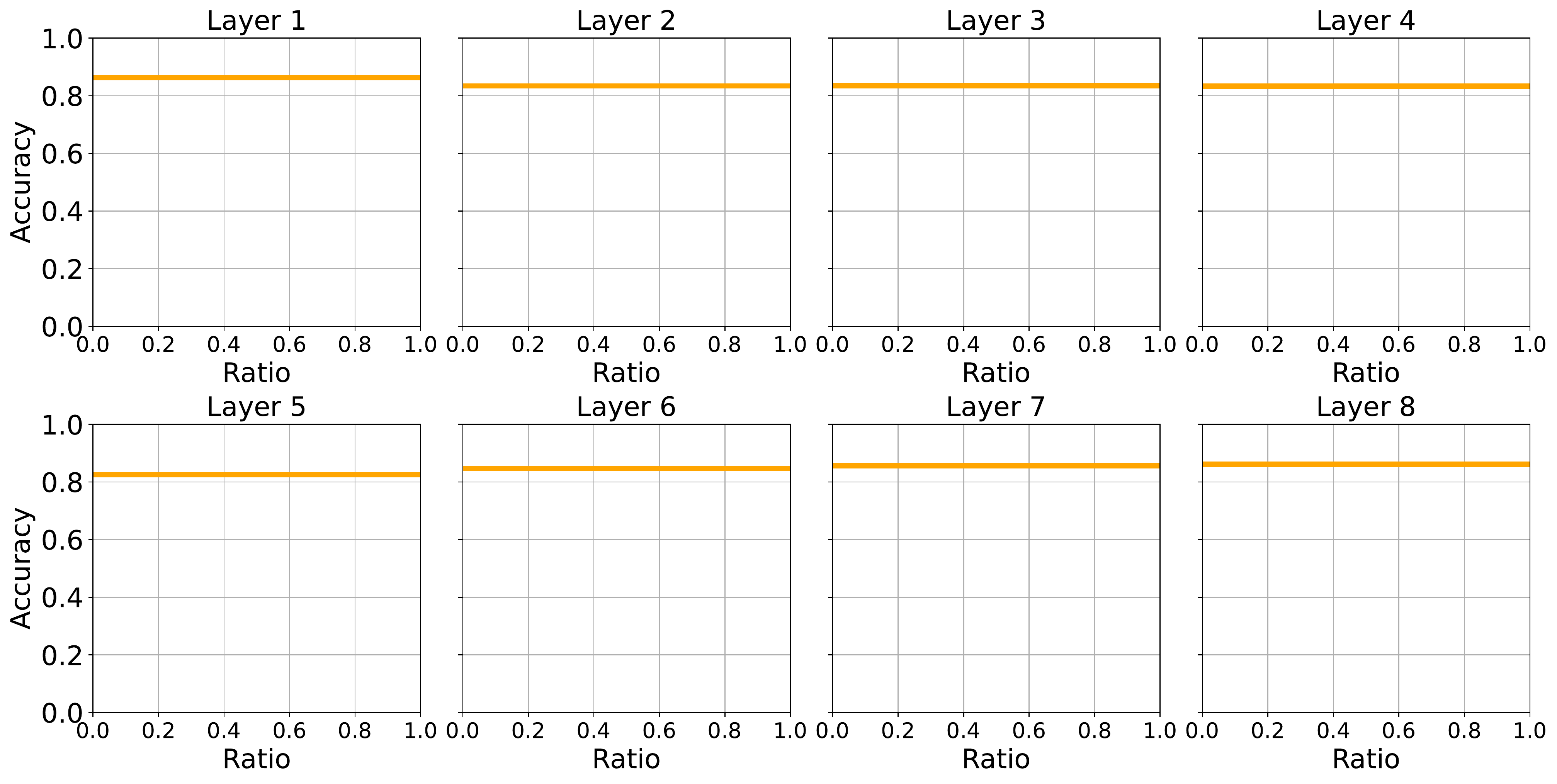}
        \caption{Linear mode connectivity of stitched networks initialized with the least squares solution. Here the result is still not deterministic, because of the dataset iteration order randomness, which differs in each solution.}
    \end{subfigure}
    \vspace{20pt}
    \caption{Relative accuracy of the stitched network with respect to $\lambda$, where the transformation matrix is $M_{\lambda}=\lambda M_1 +(1-\lambda) M_{2}$ for pairs of transformation matrices $M_1$ and $M_2$. Different plots correspond to matchings on different layers.}
    \label{fig:mode_connect}
\end{figure}

\subsection{Experiment 7: Details for sparsity — direct vs. task loss matching}
\label{appendix:sparsity}
\begin{figure}[t]
    \centering
    \begin{subfigure}[t]{.48\textwidth}
      \centering
        \vspace{0pt}
        \includegraphics[width=1.0\linewidth]{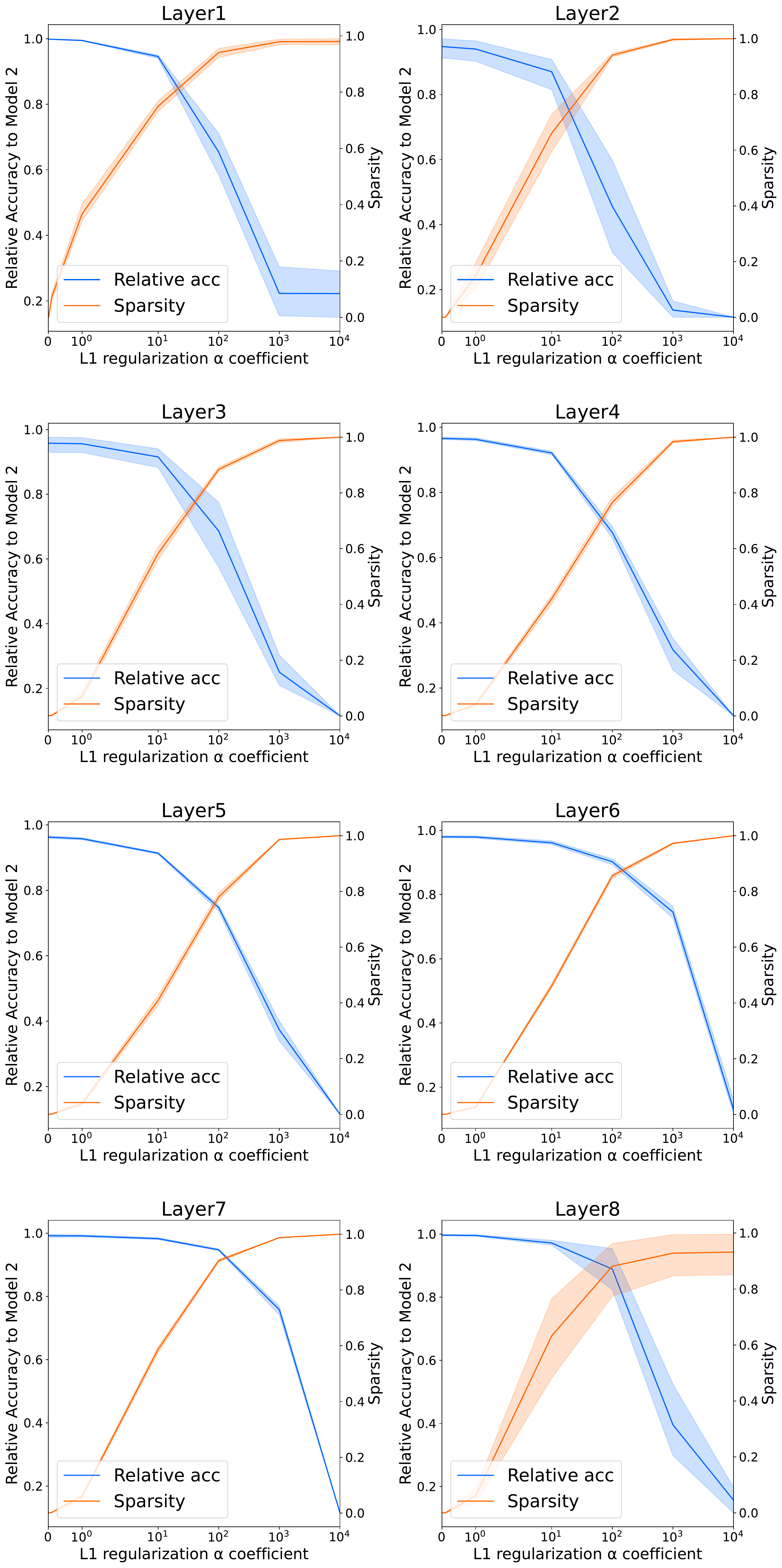}
        \caption{Task loss matching}
    \end{subfigure}
    \quad
    \begin{subfigure}[t]{.48\textwidth}
      \centering
        \vspace{0pt}
        \includegraphics[width=1.0\linewidth]{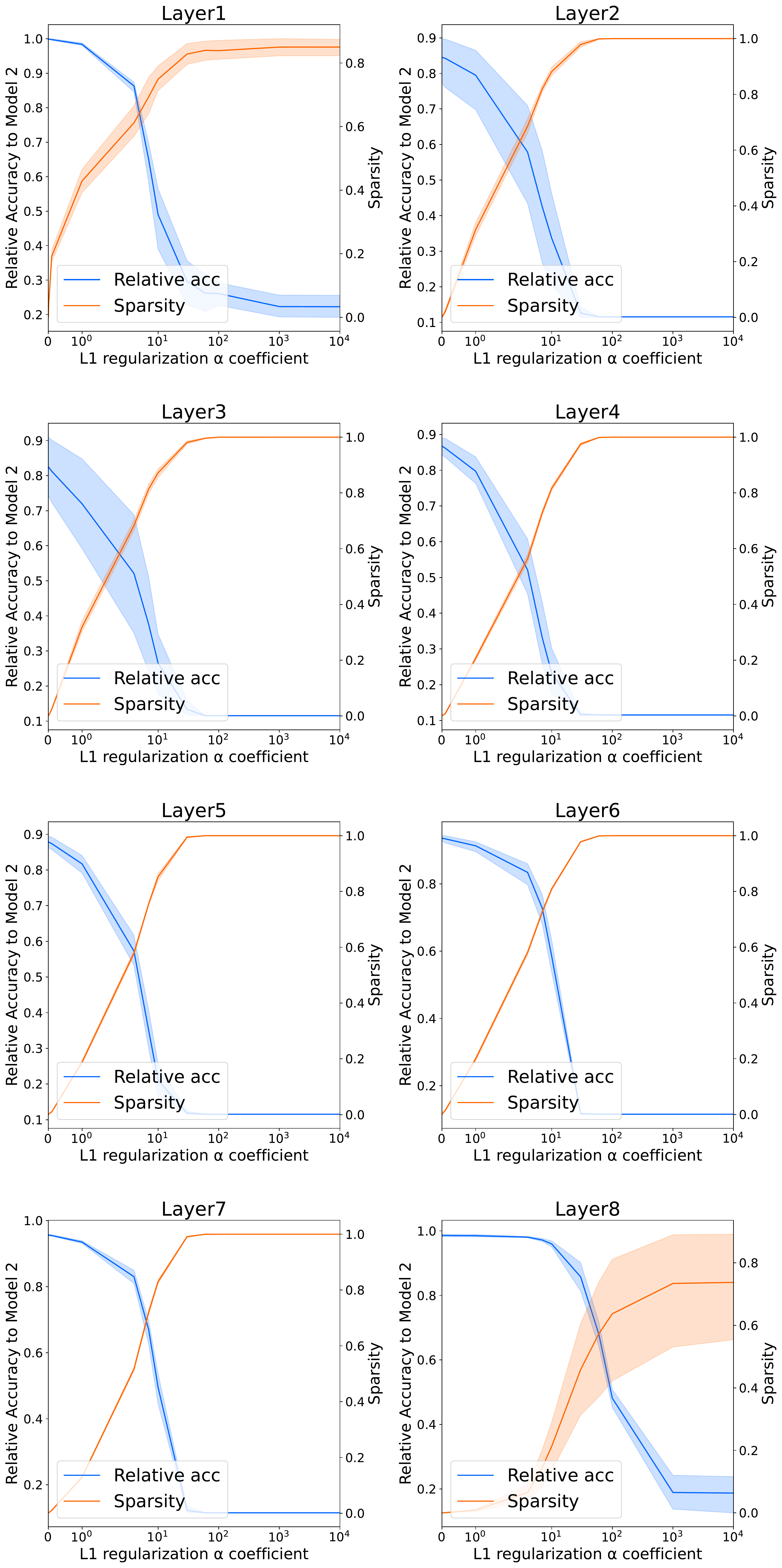}
        \caption{Direct matching}
    \end{subfigure}
    \caption{Plotting relative accuracy to Model 2 and sparsity for different L1-regularization $\alpha$ coefficients. Different plots correspond to matchings on different layers of the Tiny-10 architecture. Results are averages of 5 runs, bands denote standard deviations.}
    \label{fig:l1_acc_sparse}
\end{figure}

\begin{figure}[t]
    \centering
    \includegraphics[width=1.0\linewidth]{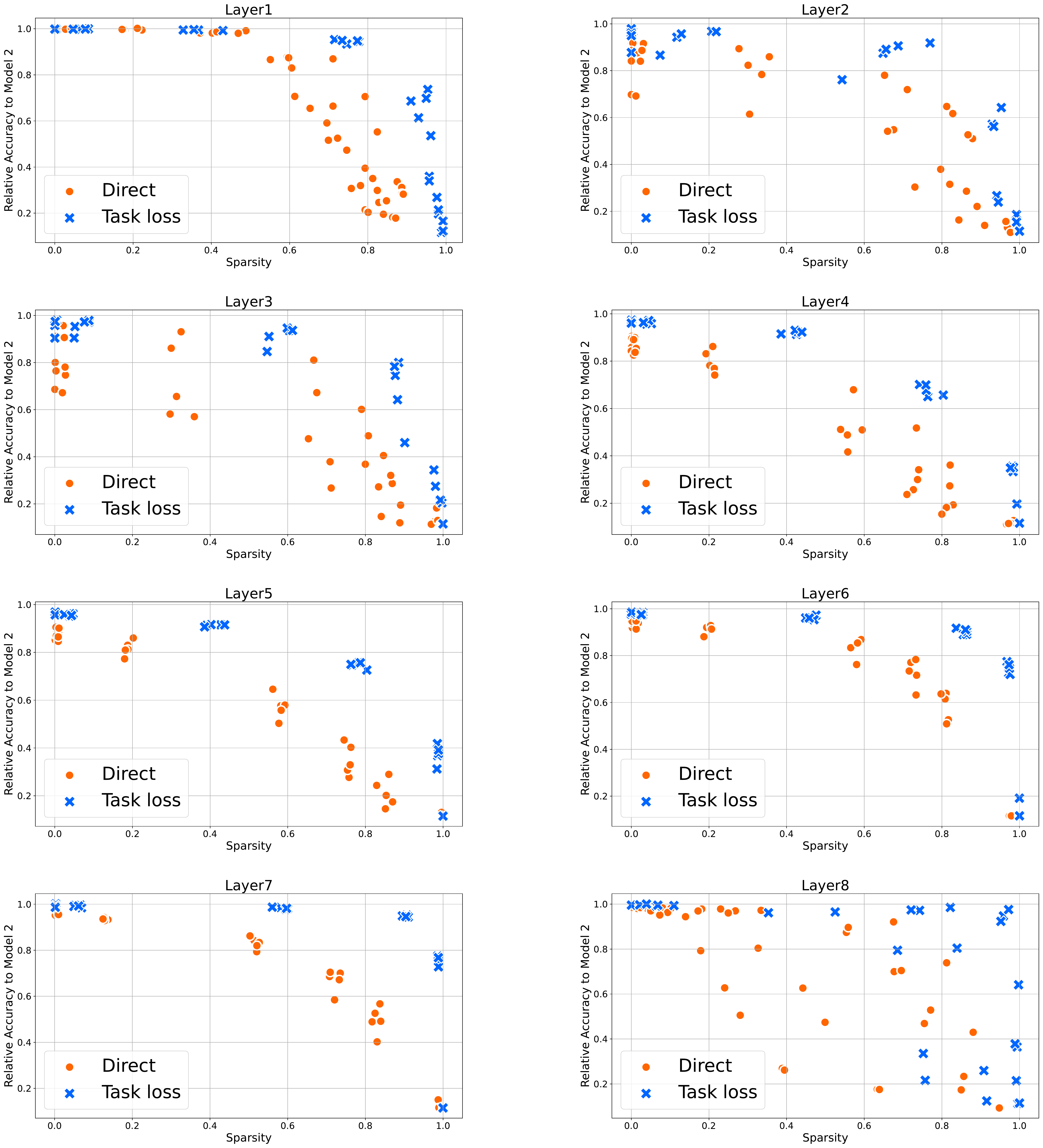}
    \caption{Relative accuracy to Model 2 with respect to sparsity. Different plots correspond to matchings on different layers.}
    \label{fig:sparse_acc}
\end{figure}

To compare the sparsity tolerance of task loss matching and direct matching, we trained 5 stitching layers with both methods until convergence, which means 30 and 200 epochs, respectively. 
We used the Tiny-10 architecture, and the CIFAR-10 dataset. For task loss matching we utilized the same methodology and settings as described in Appendix~\ref{appendix:exp1_details}. We trained the direct matching using the Adam optimizer with parameters $\beta_1=0.9$ and $\beta_2=0.999$, learning rate was set to $10^{-2}$. We ran the trainings with different L1-regularization coefficients, namely: $0$, $10^{-4}$, $10^{-3}$, $10^{-2}$, $10^{-1}$, $10^0$, $10^1$, $10^2$, $10^3$, $10^4$ to achieve increasing sparsity. 
In order to achieve an actual sparse matrix, we set every entry of the matrix to $0$ below a certain threshold, $10^{-4}$.
We evaluated the performance of the sparse stitching matrices on the task. 
Figure \ref{fig:l1_acc_sparse} shows the relative accuracy and sparsity (ratio of zero elements) with respect to L1-regularization term, and Figure \ref{fig:sparse_acc} shows the relative accuracy with respect to sparsity for all matched layers.

\subsection{Experiment 8: Details for low rank representations}
\label{appendix:low_rank}

For task loss matching, the low rank transformation is ensured by a bottleneck in the stitching layer: for a prescribed rank $k \in \mathbb{N}$, an $n \times n$ transformation matrix is parametrized by the product of two $k \times n$ and $n \times k$ sized matrices.

For least squares matching, we use the reduced rank regression outlined in the main text.

In all other respects, we utilize the same methodology and settings to set or train the stitching layer as described in Appendix~\ref{appendix:exp1_details}. 

Figure~\ref{fig:low_rank_all} depicts further results for all the layers of the Tiny-10 architecture for this experiment.

\begin{figure}[t!]
\centering
\begin{subfigure}[t]{.41\textwidth}
  \centering
    \vspace{0pt}
    \includegraphics[width=1.0\linewidth]{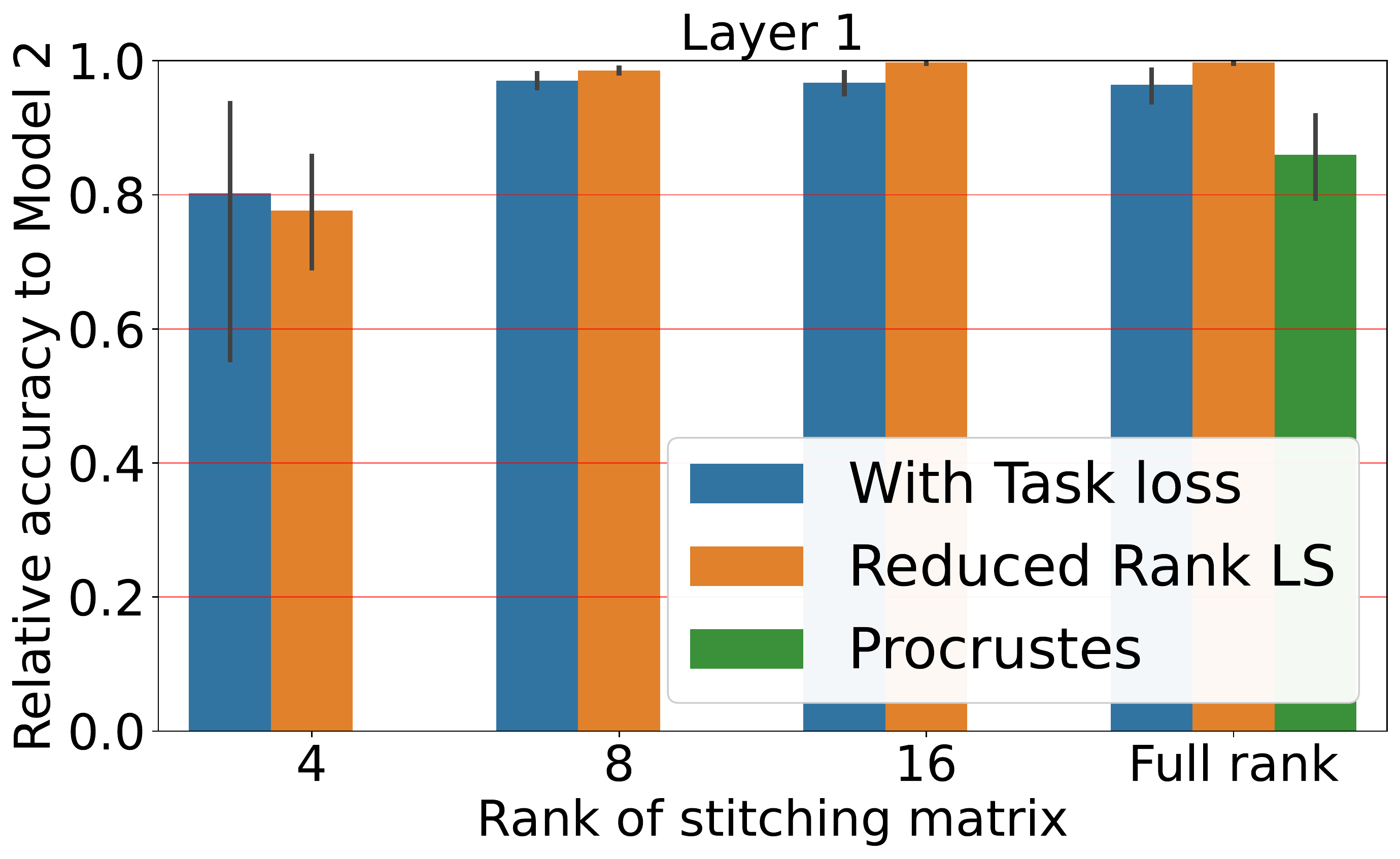}
\end{subfigure}
\qquad
\begin{subfigure}[t]{.41\textwidth}
  \centering
  \vspace{0pt}
    \includegraphics[width=1.0\linewidth]{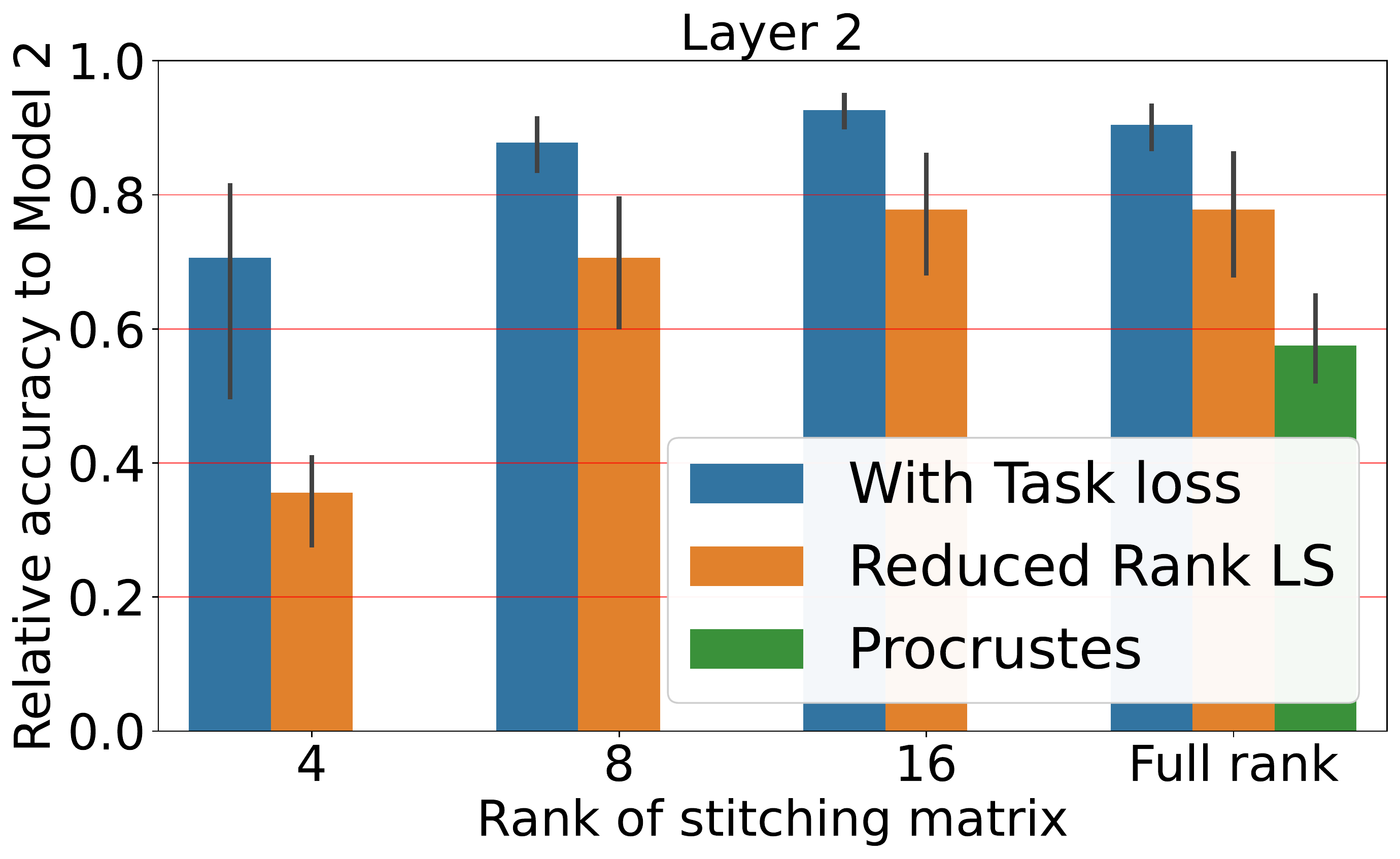}
\end{subfigure}
\qquad
\begin{subfigure}[t]{.41\textwidth}
  \centering
  \vspace{10pt}
  \includegraphics[width=1.0\linewidth]{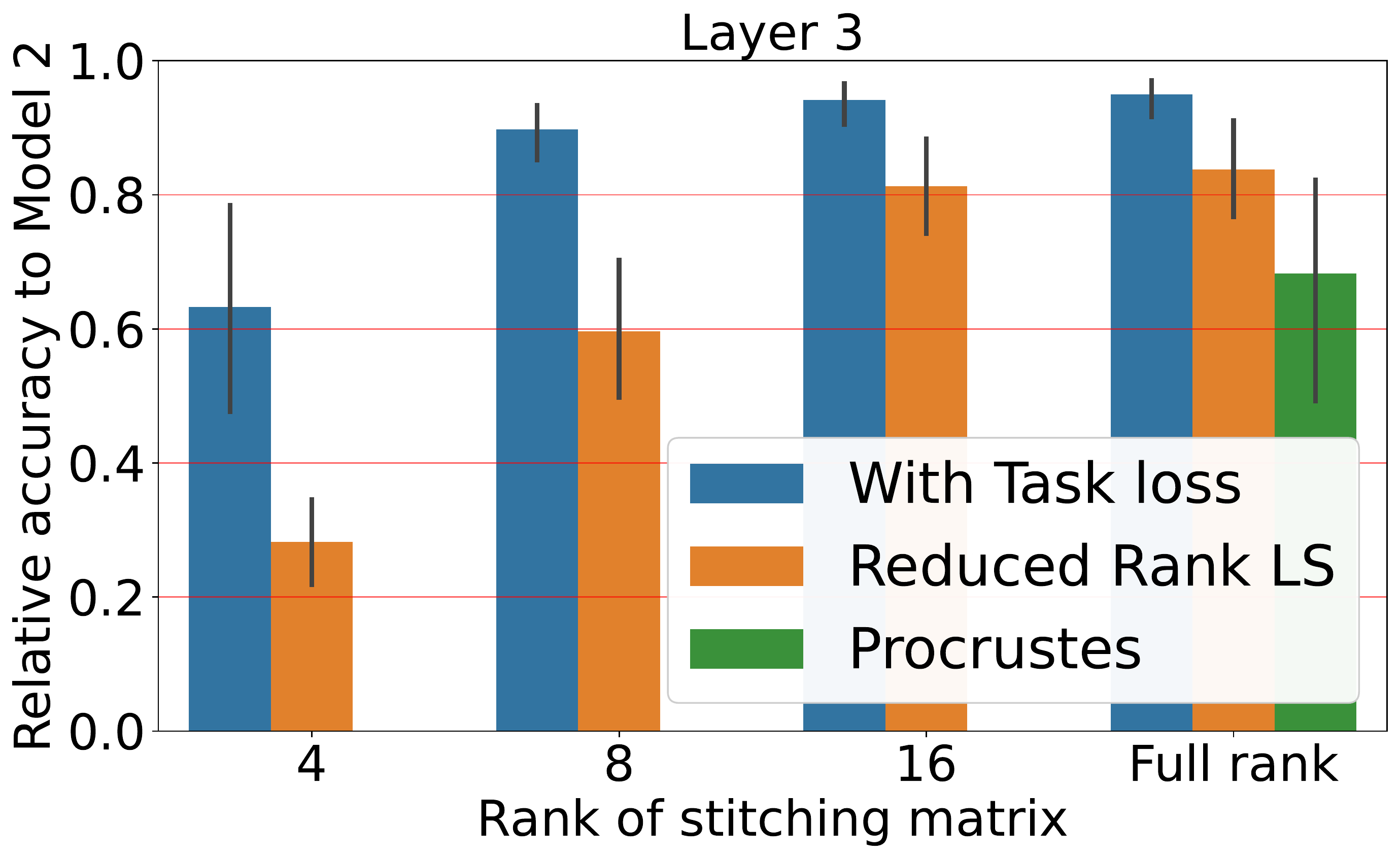}
\end{subfigure}
\qquad
\begin{subfigure}[t]{.41\textwidth}
  \centering
  \vspace{10pt}
    \includegraphics[width=1.0\linewidth]{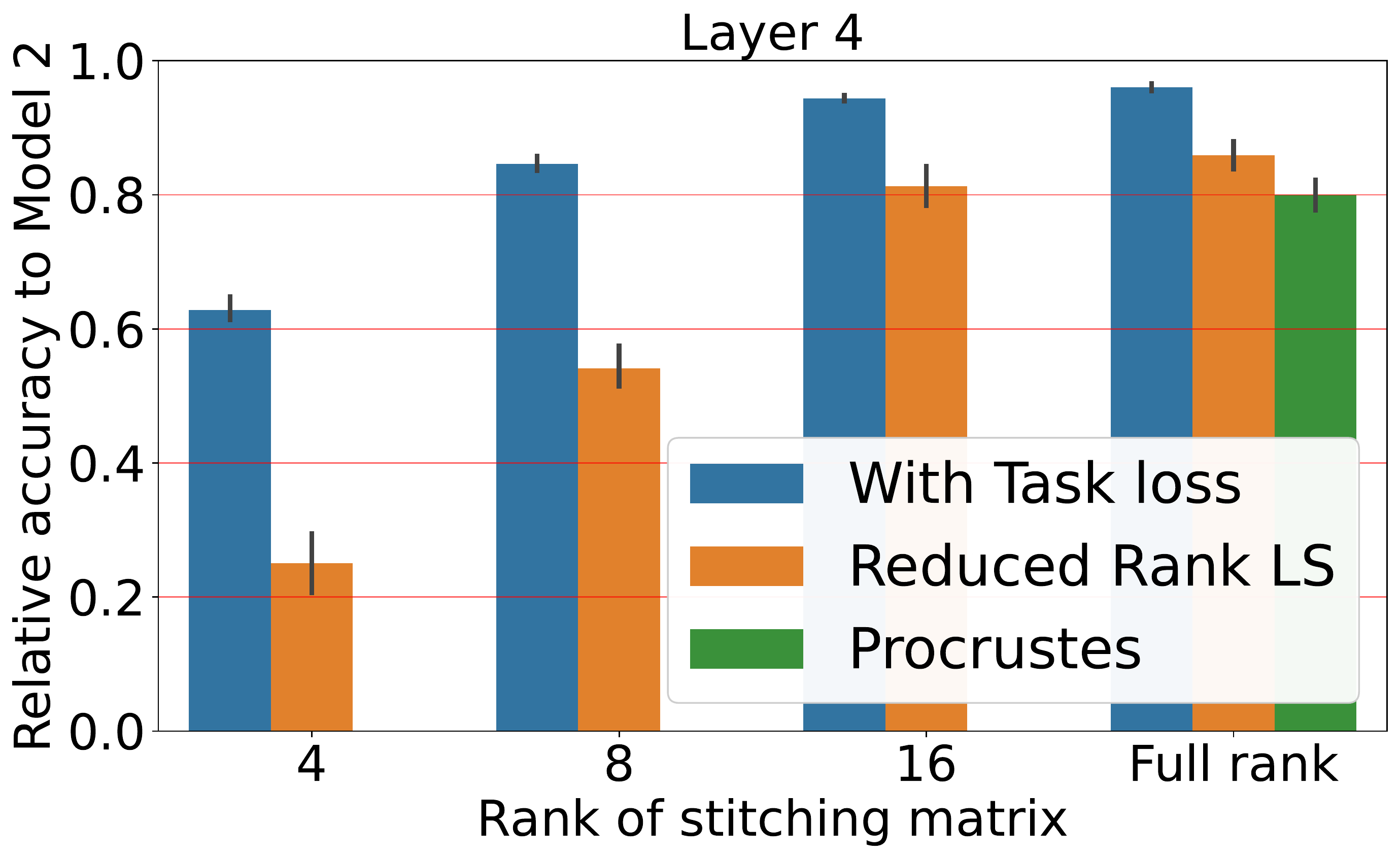}
\end{subfigure}
\begin{subfigure}[t]{.41\textwidth}
  \centering
  \vspace{10pt}
  \includegraphics[width=1.0\linewidth]{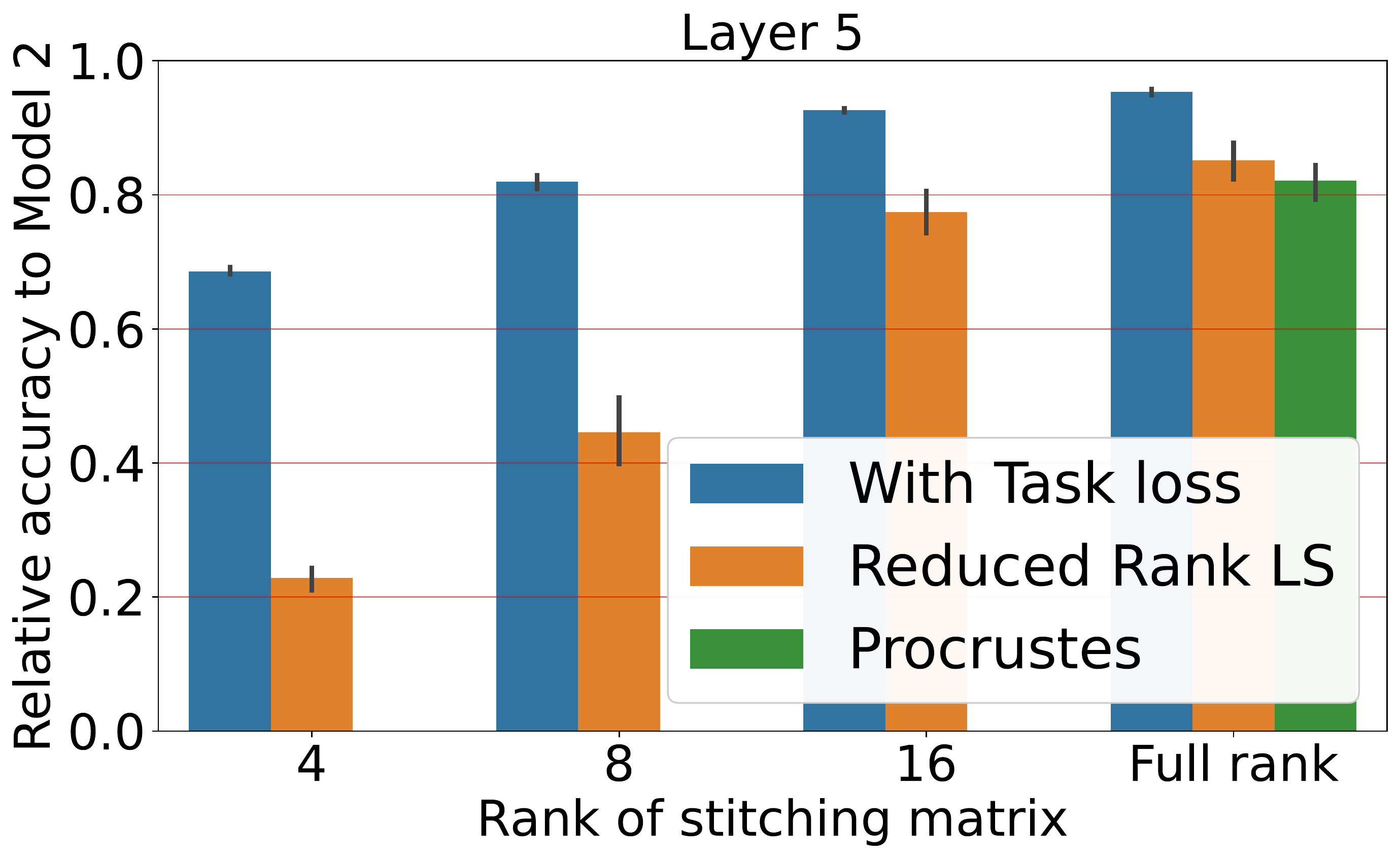}
\end{subfigure}
\qquad
\begin{subfigure}[t]{.41\textwidth}
  \centering
  \vspace{10pt}
    \includegraphics[width=1.0\linewidth]{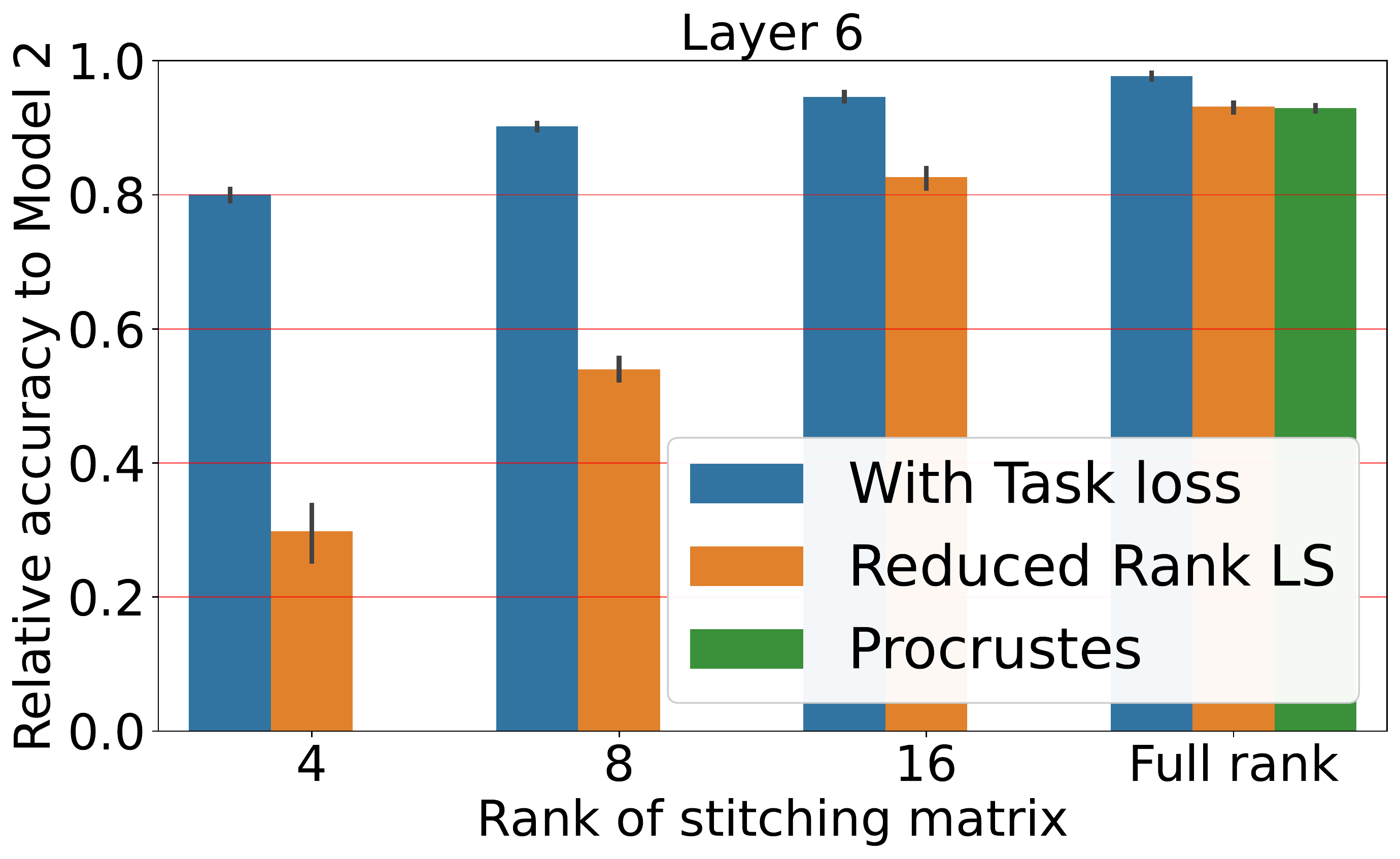}
\end{subfigure}
\qquad
\begin{subfigure}[t]{.41\textwidth}
  \centering
  \vspace{10pt}
  \includegraphics[width=1.0\linewidth]{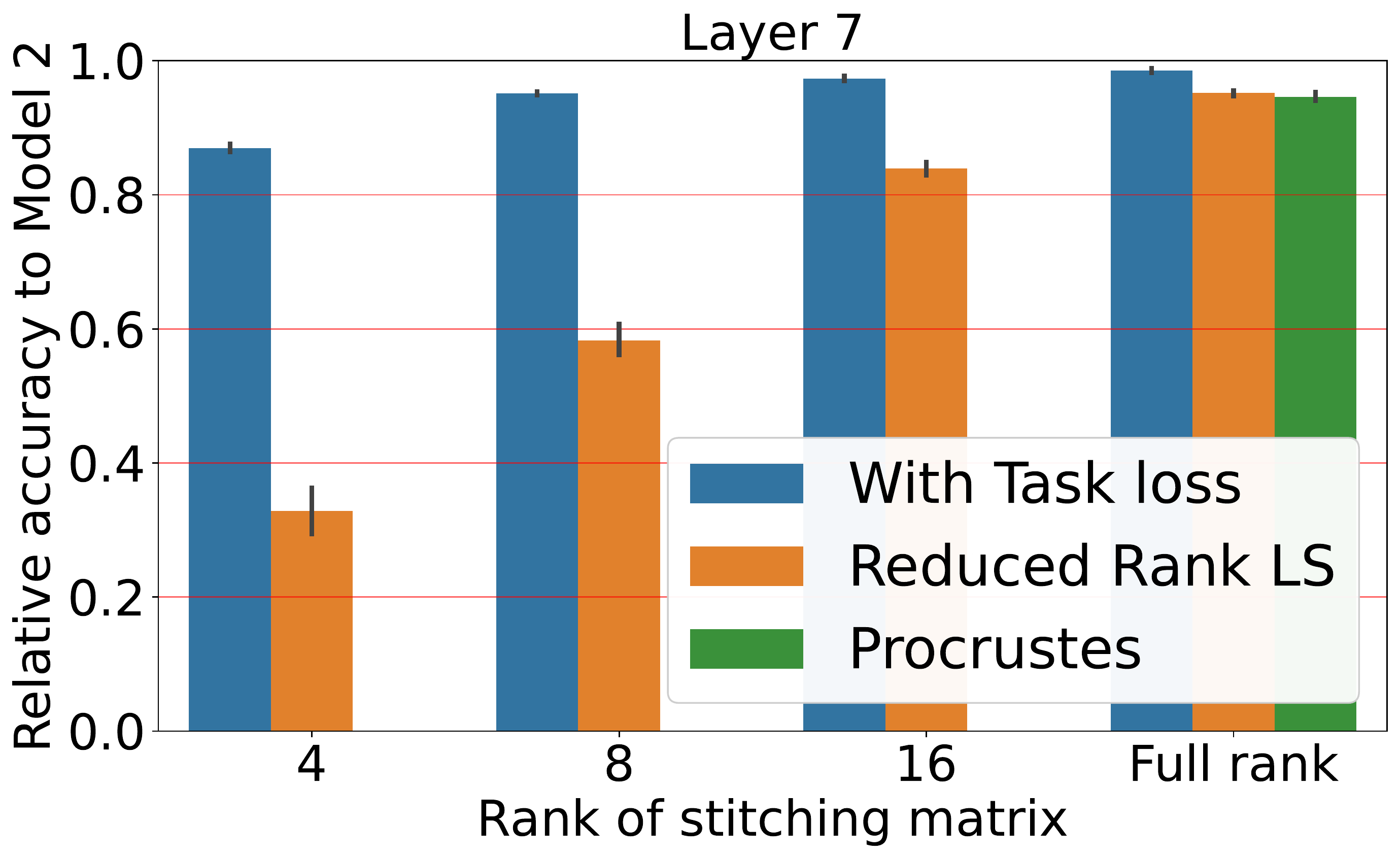}
\end{subfigure}
\qquad
\begin{subfigure}[t]{.41\textwidth}
  \centering
  \vspace{10pt}
    \includegraphics[width=1.0\linewidth]{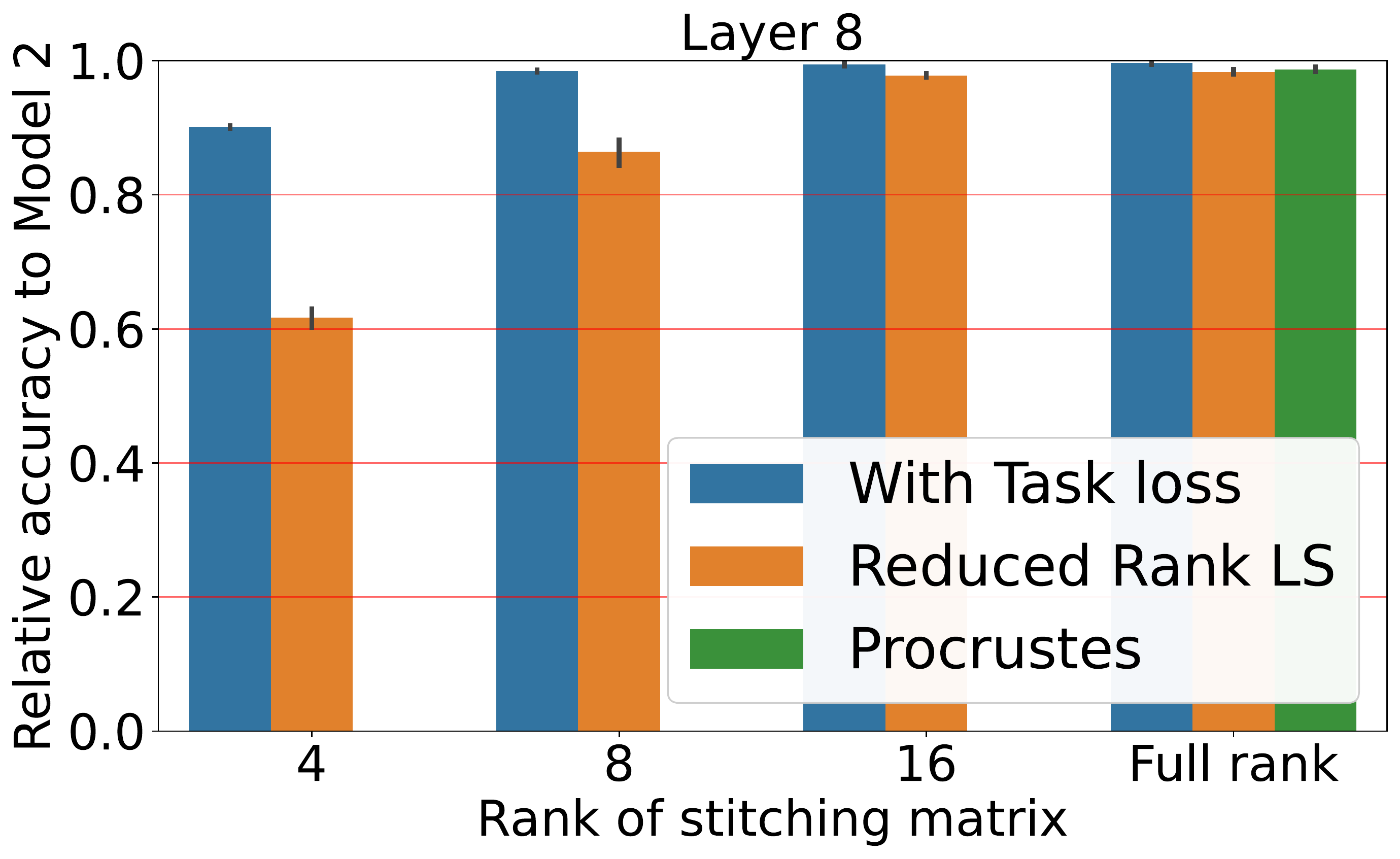}
\end{subfigure}
\caption{Performance of the low-rank analogues of least squares and task loss matchings in terms of relative accuracy. Averages of 5 runs, error bars denote standard deviations.}
\label{fig:low_rank_all}
\end{figure}

\subsection{Compute resources}

We trained and evaluated approximately 20000 stitching layers overall. We used an internal cluster with GeForce 1080 Ti and GeForce 2080 Ti GPUs and dual Intel Xeon E5-2650 v4 CPUs in the machines. Each experimental run for a stitching layer with training and evaluation together used one (or a partial) GPU and generally finished under ten minutes for the Tiny-10 architecture, and under fifteen minutes for a 1-wide ResNet-20 on these machines.

\section{Further direct matching methods}

\subsection{Weighted mean squares matching of activations}
\label{appendix:prioritize_activations}

In these experiments, we train the stitching layer to minimize the Weighted Mean Squared (WMS) objective. Our goal was to find a more or less simple method which sorts the activations, and assigns higher weights to the activations where a more precise matching is beneficial from the perspective of task performance. 
Roughly speaking, this method tries to identify more important features and put them into focus during the direct matching.

Given flattened activations $A, B\in  \mathbb{R}^{n\times s}$ and a matrix of weights $W\in \mathbb{R}^{n\times s}$ we used Stochastic Gradient Descent to find the $M$ which minimizes the WMS distance between $A M$ and $B$:

\begin{equation}\label{eq:WMS}
\min_{M\in \R^{s\times s}}\Vert (A M - B) \circ W \Vert_F,
\end{equation}

where $\circ$ is the Hadamard product $[X\circ Y]_{ij}=X_{ij}\cdot Y_{ij}$. 

For the experiments listed below, we used the Tiny-10 architecture, the Adam optimizer with parameters $\beta_1=0.9$ and $\beta_2=0.999$, learning rate was set to $10^{-2}$, batch size was $64$ and we trained for 200 epochs. 

In the following, we will present different choices for the unnormalized weight matrices $W_u$. In each case, we will normalize the weight matrix with entries in $[0,1]$ as follows:
\begin{equation}
    W = \frac{W_u - w_{\min}\mathbf{1_N}\mathbf{1_s^T}}{w_{\max}},
\label{eq:normalize}
\end{equation}
where $w_{min},w_{max}$ denote the smallest and largest entries in $W$.

\paragraph{Gradient based weighting} As the Class Saliency method points out \cite{simonyan2014deep}, the gradients of a network's outputs with respect to its inputs may contain valuable information about the importance of each part of the input. Our method is analogous, however we inspect the hidden activations instead of the inputs. In this experiment, we solved a weighted mean squares matching \eqref{eq:WMS} where the weights were determined from the gradients in the network. 

In the following, fix the network weights $\phi$ and the layer $L$, and denote by $T_k=T_{\phi,L}^{k}$ the task map from $\mathcal{A}_{\phi,L}$ to the $k$th class, and $\hat{T}_k=\hat{T}_{\phi,L}^{k}$ the task map from $\mathcal{A}_{\phi,L}$ to the $k$th class without the last non-linearity. Denote by $\partial_{ij}h=\partial_jh(x_i)$ the partial derivative of a real-valued function $h:\mathcal{A}_{\phi,L}\to \R$ on the $i$th datapoint according to the $j$th coordinate in the activation space $\mathcal{A}_{\phi,L}$.

We experimented with four variants of gradient based weighting; in each case, we took as unnormalized weights $W^u$ a matrix of the form
$$ W^u_{ij}=\partial_{ij} f (g_k),$$
where $f\in \{\sum_k, \operatorname{argmax}_k\}$ and $g_k\in \{T_k,\hat{T}_k\}$. In each case we solved the weighted mean squares matching problem \eqref{eq:WMS} by taking as weight matrix $W$ the normalization of such a weight matrix $W^u$. In particular, we considered the following variants:
\begin{itemize}
  \item Gradients of summed output of Model 2 with respect to its matched activations: 
  $$ W^u_{ij}=\partial_{ij}\sum_{k}T_k$$
  \item Gradients of the Model 2's output of the predicted class with respect to its matched activations: 
  $$ W^u_{ij}=\partial_{ij}\operatornamewithlimits{argmax}_k T_k$$
  \item Gradients of summed output logits of Model 2 with respect to its matched activations: 
  $$W^u_{ij} =\partial_{ij}\sum_k\hat{T}_k$$
  \item Gradients of Model 2's output logits of the predicted class with respect to its matched activations: 
  $$ W^u_{ij}=\partial_{ij}\operatornamewithlimits{argmax}_k\hat{T}_k$$
\end{itemize}

However, we found that overall these methods did not result in a consistent performance gain compared to the direct matching with unweighted least squares objective.

\paragraph{Activation based weighting} 

Another approach is to take as unnormalized weights $W_u=B$, and use the corresponding normalized weight matrix $W$ defined by \eqref{eq:normalize}. A simple way to emphasize higher values in this weight matrix is to take higher powers of each entry of $W$, i.e.\ use the weighting 
\begin{equation}\label{eq:higherpowers}
    W^{\circ n}= W\circ \ldots \circ W,
\end{equation}
where $\circ$ denotes the (element-wise) Hadamard product.

\begin{figure}[t!]
\centering
\begin{subfigure}[t]{.41\textwidth}
  \centering
    \vspace{0pt}
    \includegraphics[width=1.0\linewidth]{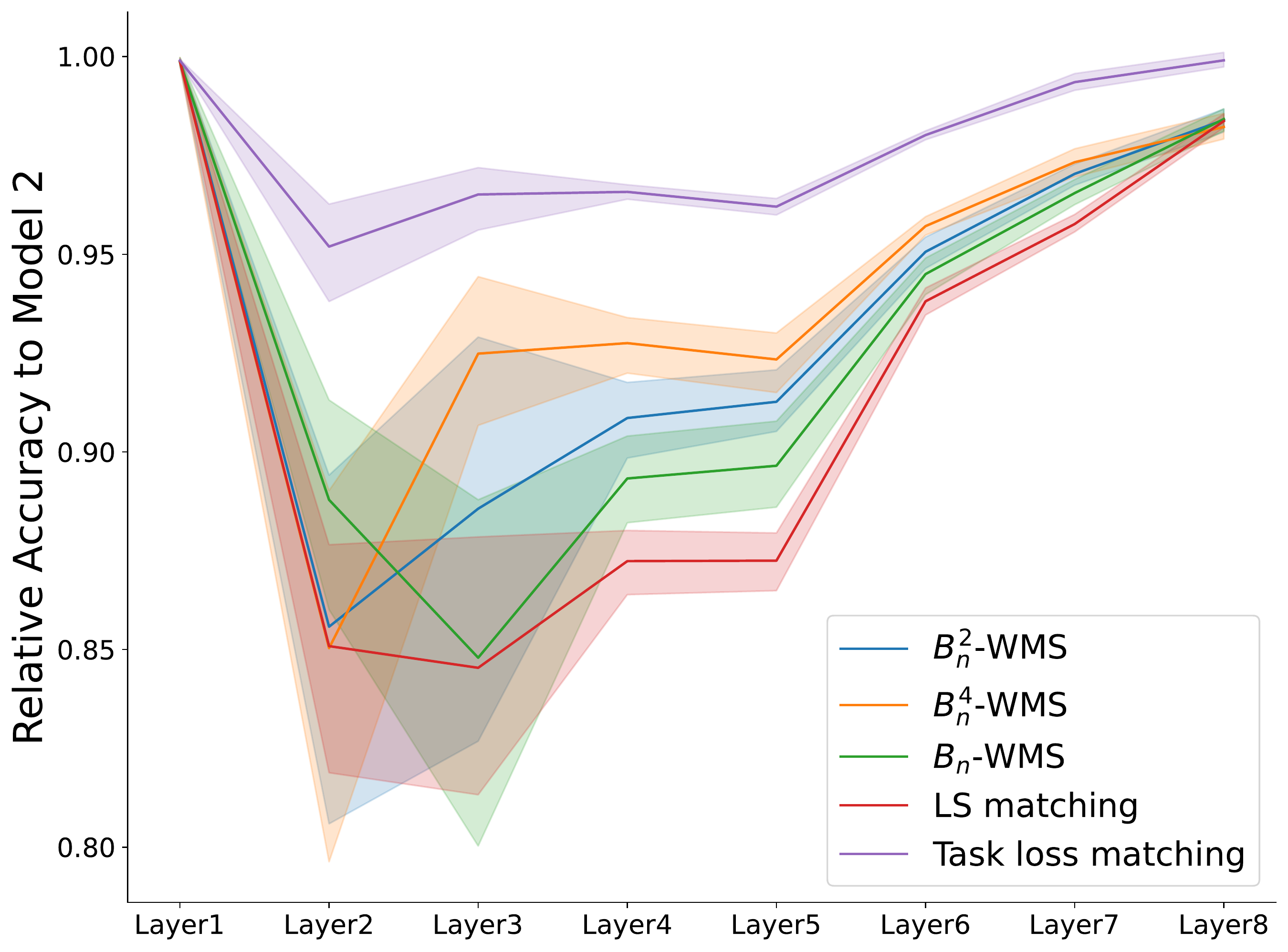}
    \includegraphics[width=1.0\linewidth]{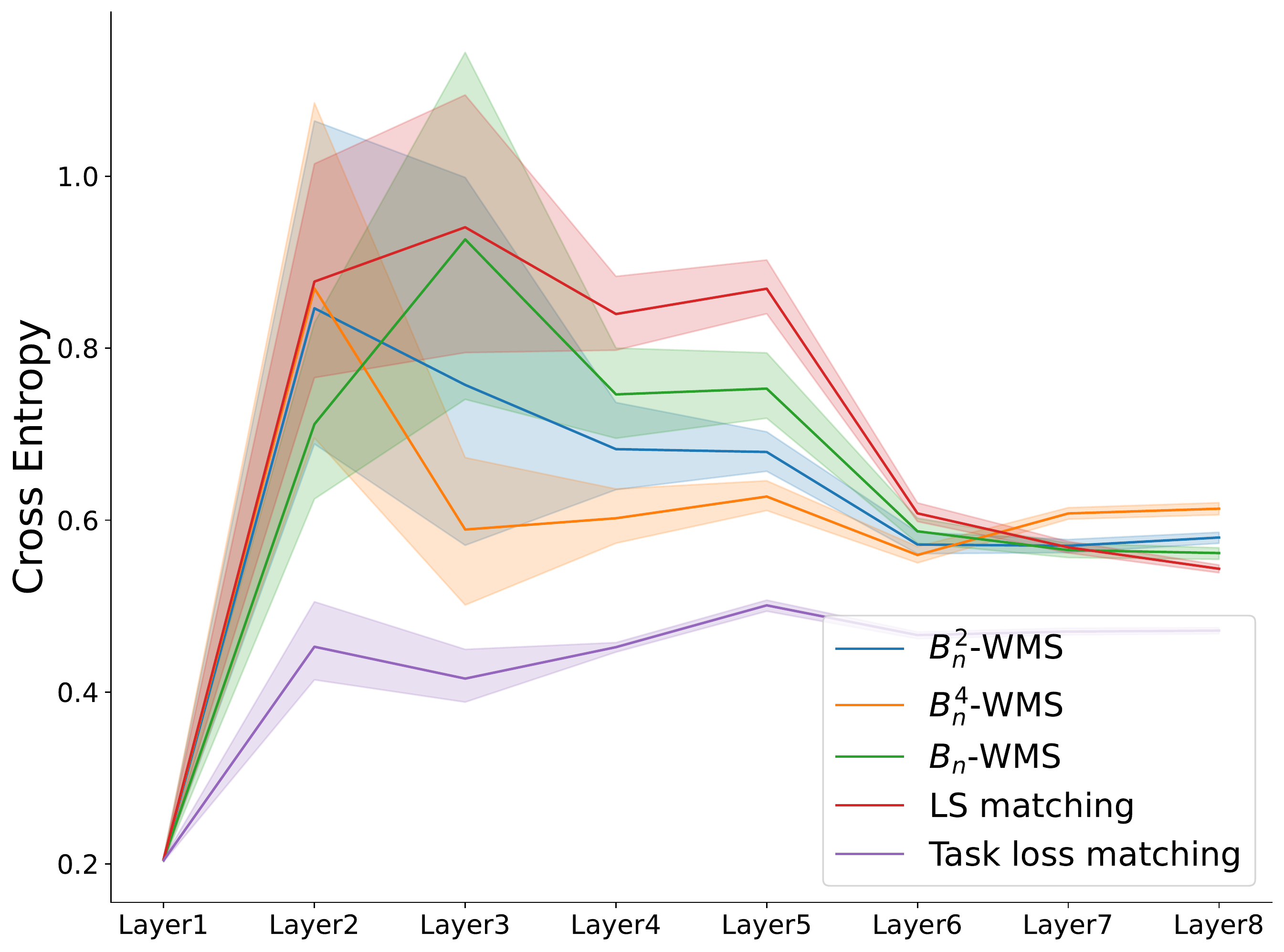}
    \caption{Activation weighted mean squared direct matching \eqref{eq:higherpowers}, activations are raised on different exponents after normalization.}
    \label{fig:act_weighted}
\end{subfigure}
\qquad
\begin{subfigure}[t]{.41\textwidth}
  \centering
  \vspace{0pt}
  \includegraphics[width=1.0\linewidth]{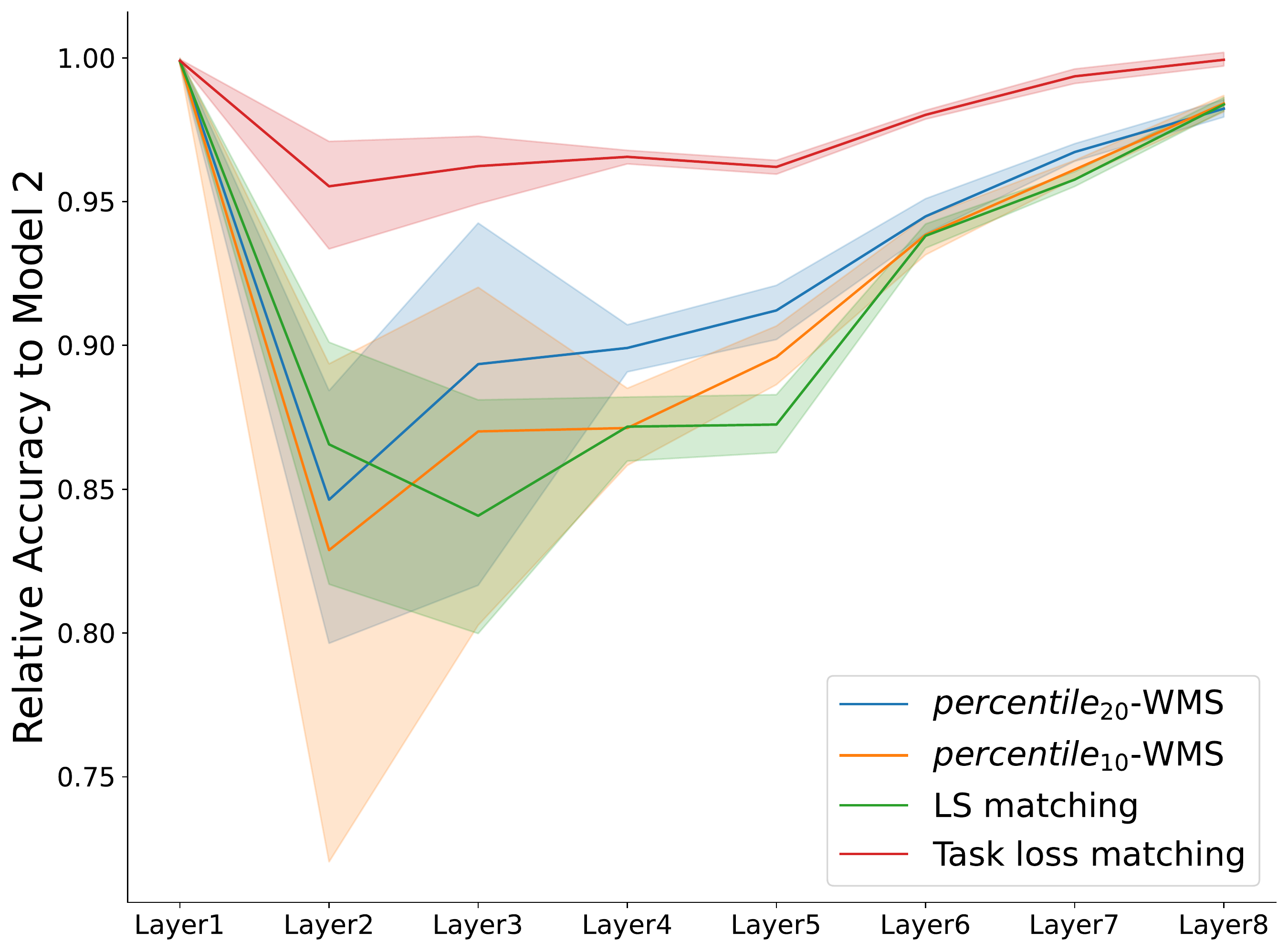}
    \includegraphics[width=1.0\linewidth]{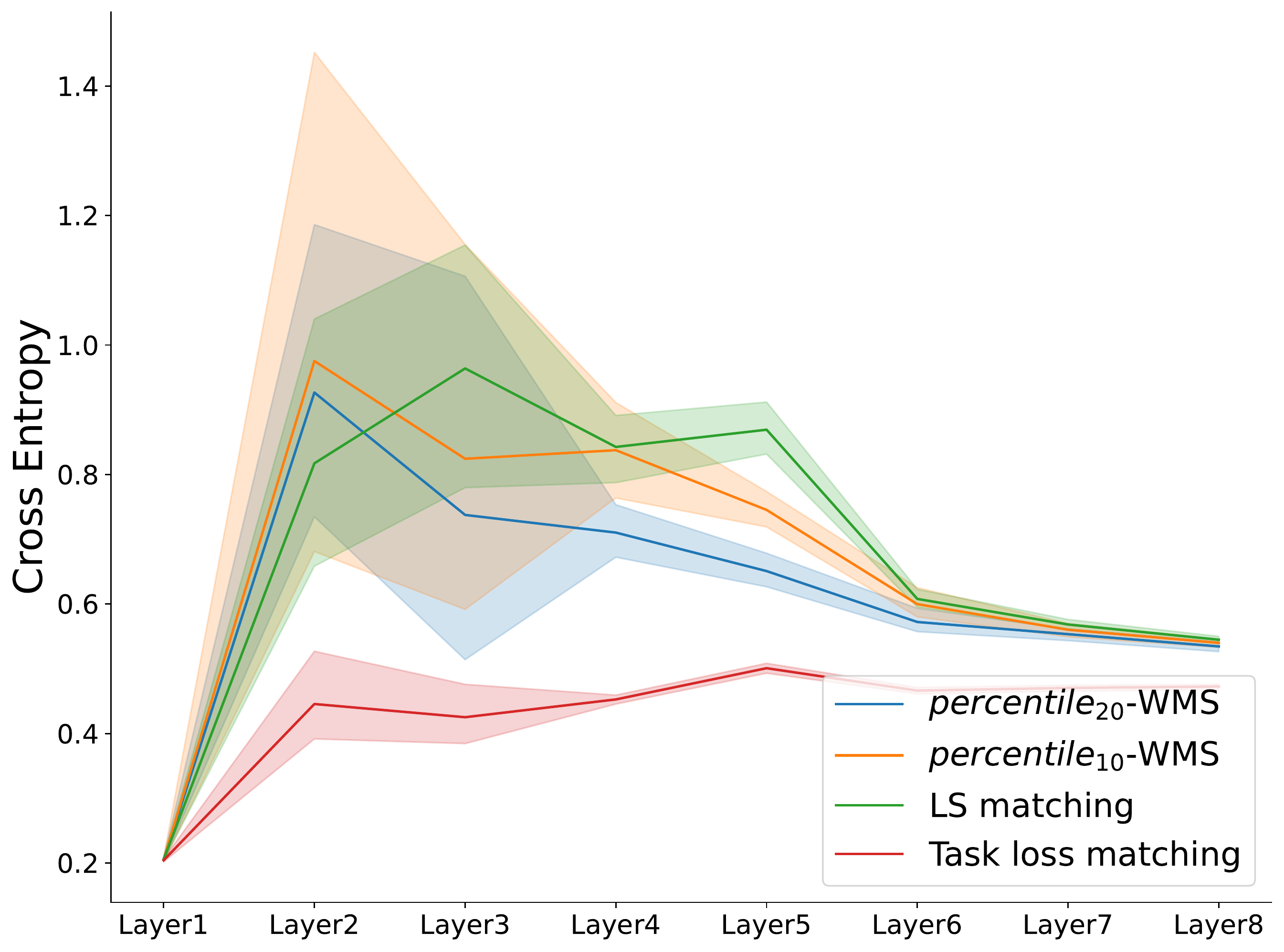}
    \caption{0-1 weighted mean squared direct matching \eqref{eq:thresholdweighting}, with the threshold $T_i$ set according to a percentile of the activations $b_{ij}$, and activations below the threshold are not matched.}
    \label{fig:threshold_weighted_hard}
\end{subfigure}
\begin{subfigure}[t]{.41\textwidth}
  \centering
  \vspace{30pt}
  \includegraphics[width=1.0\linewidth]{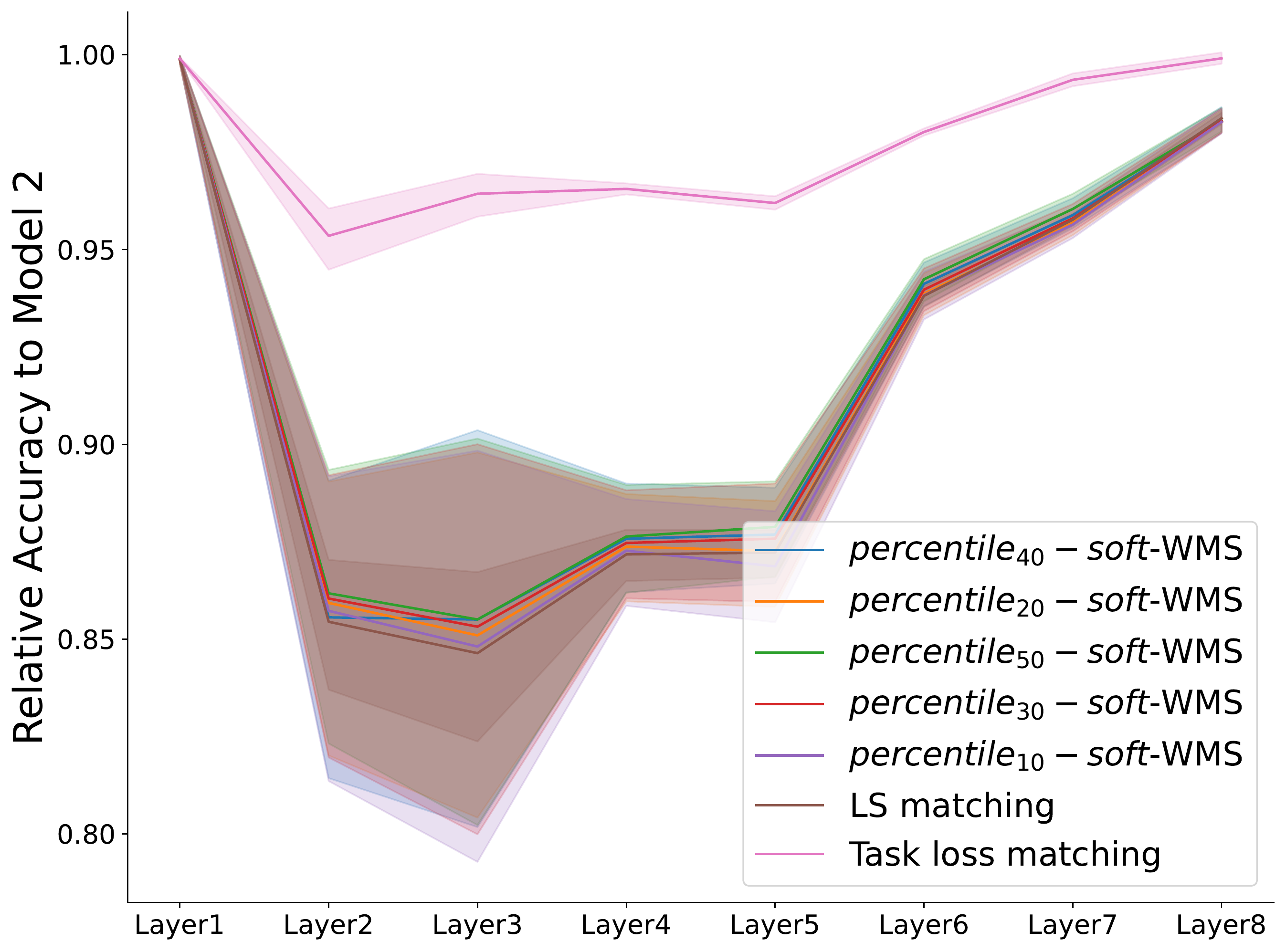}
    \includegraphics[width=1.0\linewidth]{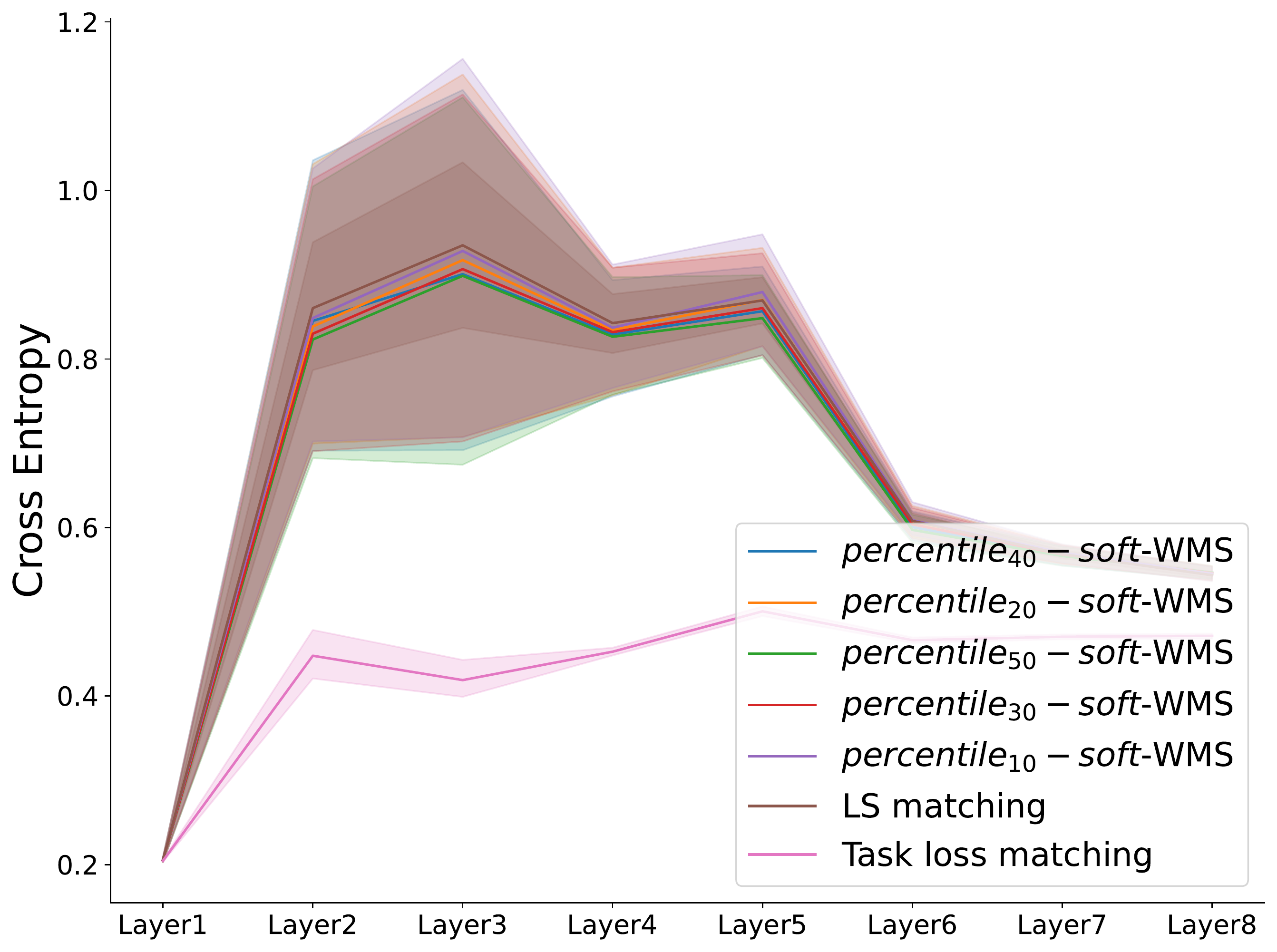}
    \caption{Activation weighted mean squared direct matching \eqref{eq:soft_thresholdweighting}, with the threshold $T_i$ set according to a percentile of the activations $b_{ij}$, and activations below threshold matched to stay below threshold.}
    \label{fig:threshold_weighted_soft}
\end{subfigure}
\qquad
\begin{subfigure}[t]{.41\textwidth}
  \centering
  \vspace{30pt}
  \includegraphics[width=1.0\linewidth]{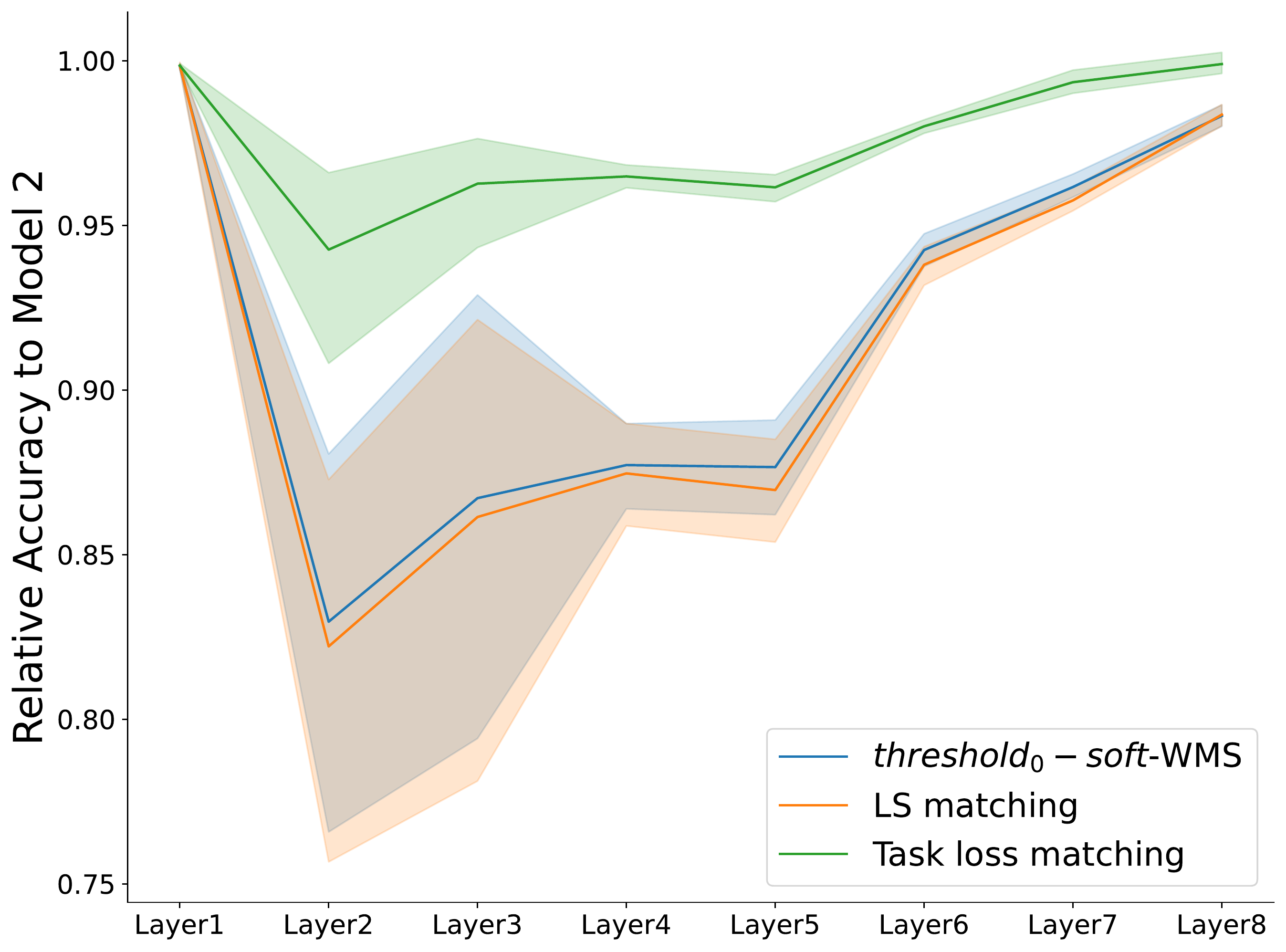}
    \includegraphics[width=1.0\linewidth]{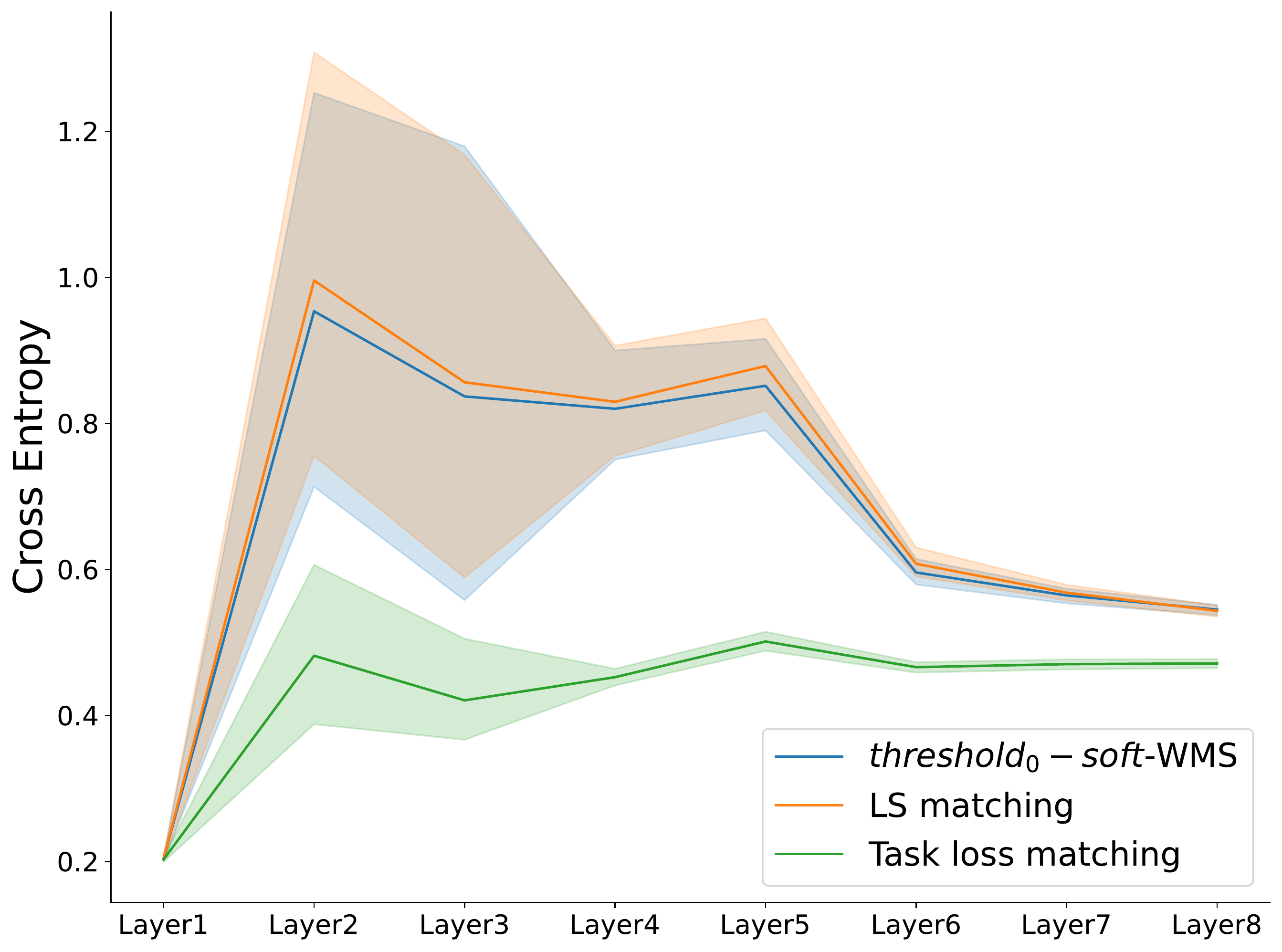}
    \caption{0-1 weighted mean squared direct matching \eqref{eq:soft_thresholdweighting}, with the threshold $T$ set to $0$, and activations below threshold matched to stay below threshold.}
    \label{fig:relu_weighted}
\end{subfigure}
\caption{Relative accuracy to Model 2 and cross-entropy of Tiny10's stitched layers with different weighted mean squared direct matching methods.}
\label{fig:prior_matching}
\end{figure}

We highlight some of these results for different values of $n$ in Figure \ref{fig:act_weighted}. Our experiments show, that matching Tiny-10 architecture's layers this way results in significant performance gain compared to the Least Squares based direct matching. This indicates that matching the higher regime of activations accurately is more important than the lower regime.

Another way to force higher activation focused matching is simply to define a threshold $T$, and only match the activations above this threshold. In particular, this corresponds to a 0 -- 1 weighting matrix 
\begin{equation}\label{eq:thresholdweighting}
    W=[1_{>T}(b_{ij})]
\end{equation}
where $1_{>T}(x)$ denotes the indicator function, which serves as a threshold function. 
See Figure \ref{fig:threshold_weighted_hard} for experiments with different thresholds $T_i$, defined by different percentiles $i\in\{10, 20\}$ of the activation values $b_{ij}$. We also tested higher threshold values defined by higher percentiles, which resulted in weaker performance. With this approach the activations below the threshold are not matched at all. 

We also experimented with a slightly modified setup, where  we define the weights as:
\begin{equation}\label{eq:soft_thresholdweighting}
    W_{ij} = 1-1_{<T}(b_{ij})\cdot 1_{<T}([AM]_{ij}) = \begin{cases}
    0,\qquad &[AM]_{ij} < T \text{ and } B_{ij} < T,\\
    1,\qquad & \text {else}.
    \end{cases}
\end{equation}

This choice of weights achieves that if a target activation in $B$ is above a certain threshold, the corresponding activation distance $(b_{ij}-[AM]_{ij})^2$ is always matched with weight 1. However, if the target activation in $B$ is below the chosen threshold, then the direct matching only penalizes the corresponding distance if the matched activation $[AM]_{ij}$ exceeds the threshold. Informally, we don't care about the residuals as long as both activations are below the threshold. See Figure \ref{fig:threshold_weighted_soft} for experiments with different thresholds $T_i$ defined by different percentiles $i\in\{10, 20, 30, 40, 50\}$. The higher thresholds resulted in weaker performance.

Another version of this last experiment (with weights \eqref{eq:soft_thresholdweighting}), is to set the threshold $T=0$, which is a reasonable choice because Tiny-10 uses ReLU activations. See Figure \ref{fig:relu_weighted} for results.

\section{Further experiments}

\subsection{A sanity check for stitching}
\label{appendix:layer_heatmap}

\begin{figure}[t!]
    \centering
    \begin{subfigure}[t]{.48\textwidth}
        \centering
        \vspace{0pt}
        \includegraphics[width=1.0\linewidth]{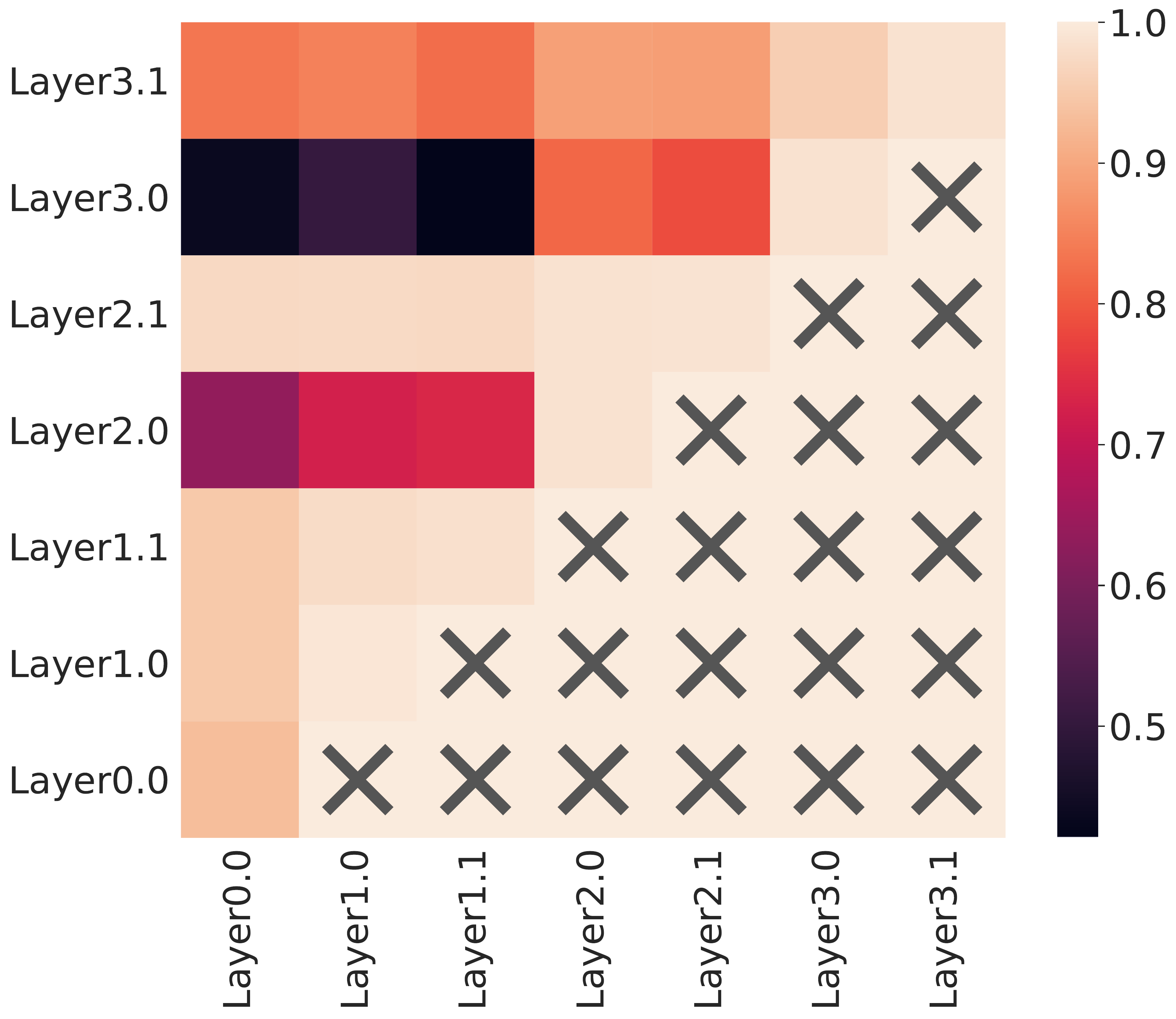}
        \caption{Heatmap of relative accuracies to Model2.}
    \end{subfigure}
    \quad
    \begin{subfigure}[t]{.48\textwidth}
        \centering
        \vspace{0pt}
        \includegraphics[width=1.0\linewidth]{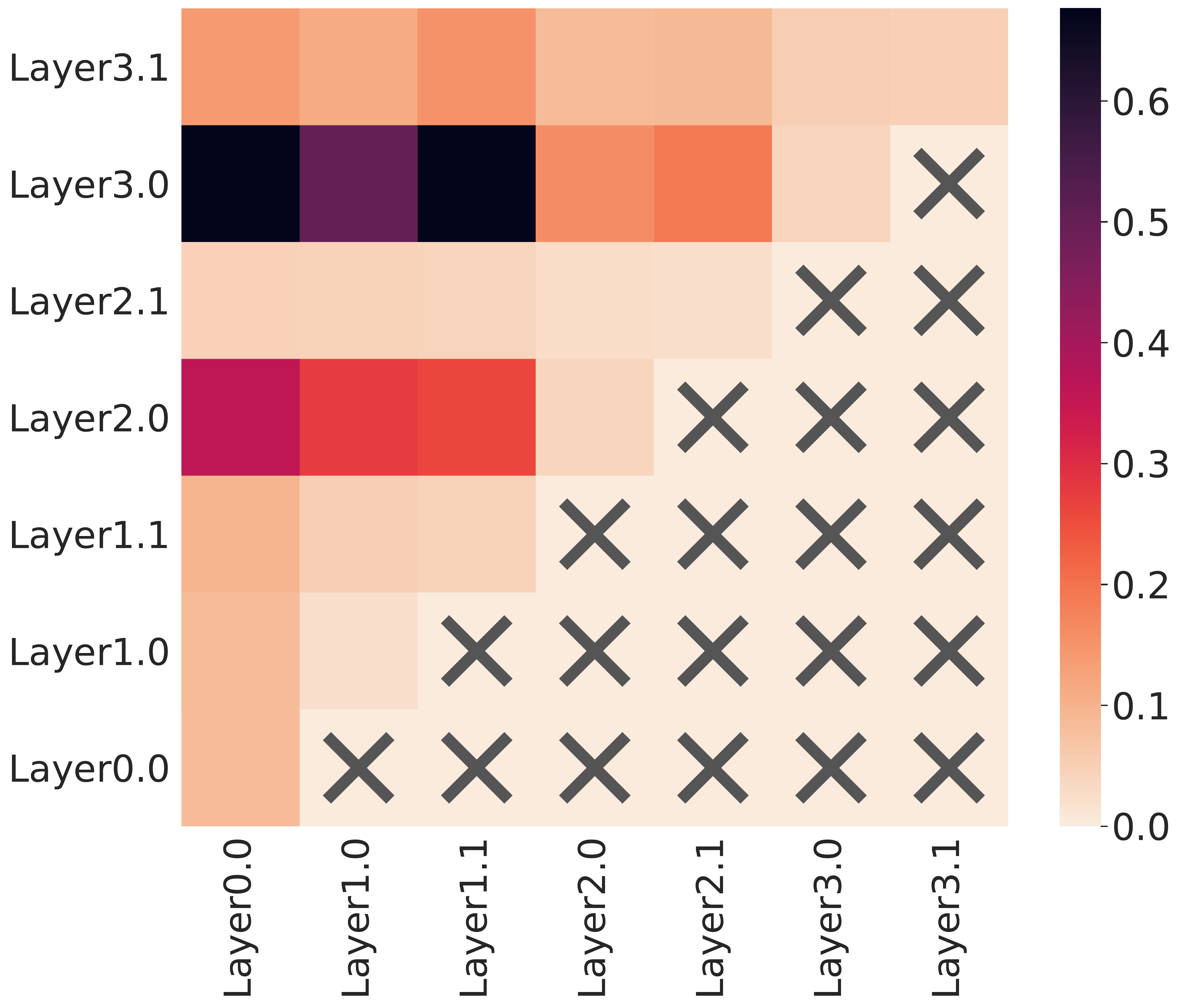}
        \caption{Heatmap of cross entropies between the stitched model's and Model2's outputs.}
    \end{subfigure}
\caption{Stitching between two ResNet-20 networks, trained on the same data but with different sample order. The stitching layer was trained from random initialization. $i$th row and $j$th column refers to the two networks' $i$th and $j$th layers.} 
\label{fig:resnet_heatmap}
\end{figure}

We measured the accuracy of stitching with task loss between $N^1_i$ and $N^2_j$, where $N^1, N^2$ are networks with the same architecture, trained on the same data but with different sample order, and $N_i, N_j$ refers to the $i$th and $j$th layers of the network.

Stitched activations are required to have the same dimensionality, so we downsampled the higher dimensions with maxpooling before stitching. As upsampling without additional information is questionable, we always stitched from higher dimension (earlier layer) to lower dimension (later layer).
Figure~\ref{fig:resnet_heatmap} shows the results measured on ResNet-20 architecture with width 1.
We utilized the same methodology and settings to train the stitching layer as described in Appendix~\ref{appendix:exp1_details}.

\subsection{Cross-task stitching, Feature visualization}
\label{appendix:lucid}

The following experiments use the Lucent port\footnote{\url{https://github.com/greentfrapp/lucent}} of the Lucid interpretability framework \citep{olah2017feature} to visualize how Model 2 channels are constructed from the linear combinations of Model 1 channels during direct matching and stitching.
\footnote{Note that a limitation of backpropagation-based feature visualization is that even though the visualization presents a pattern that activates the channel, it can be activated by very different patterns as well.}

\begin{figure}
\includegraphics[width=0.99\linewidth]{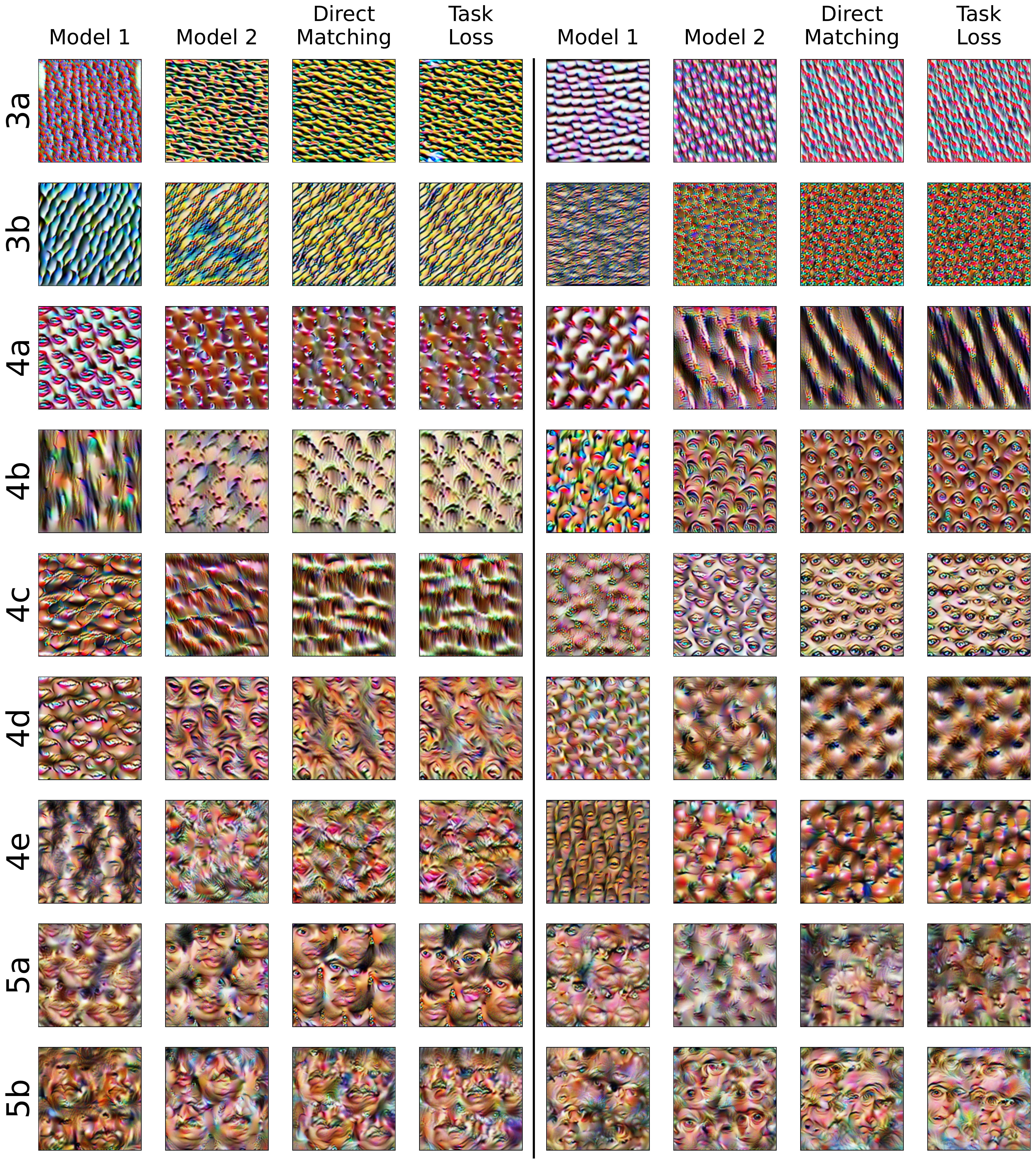}
\caption{Stitching between two different CelebA models. All images are channel visualizations by Lucent. Rows correspond to layers where stitching and visualization happens. Leftmost four columns show first channel, rightmost four columns show second channel of the given layer. In each 4-column block, first column presents Model 1 channel, for comparison. Second column presents Model 2 channel, which in a sense is the target for the last two columns. Third column presents the channel linearly combined from Model 1 channels with the least squares error to Model 2 channel. (Direct matching.) Fourth column presents the substitute of the Model 2 channel created by stitching. Each row present the first two neurons of the corresponding layer.}
\label{fig:lucid_r2r}
\end{figure}

\begin{figure}
\includegraphics[width=0.99\linewidth]{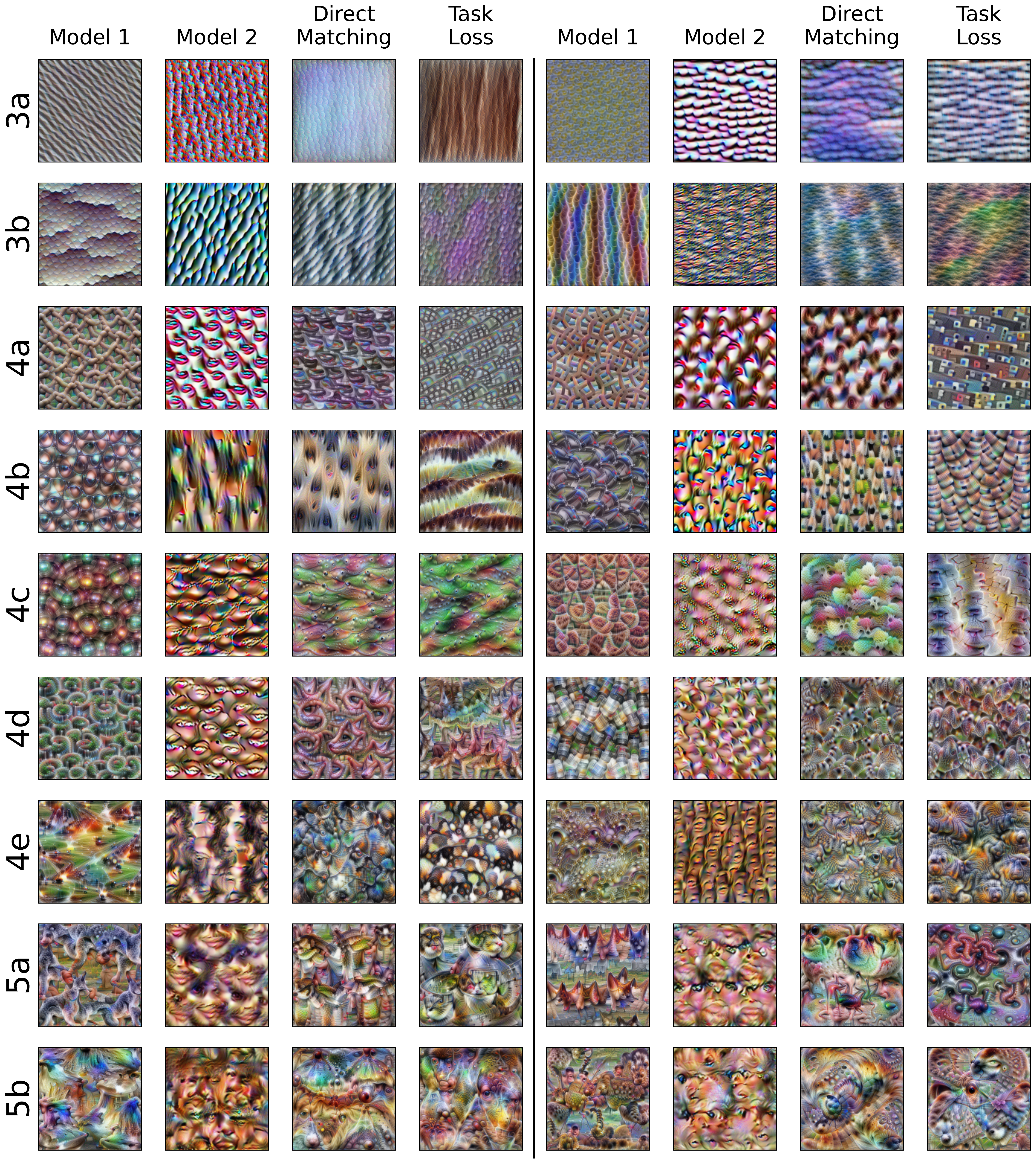}
\caption{Stitching with Model 1 = ImageNet, Model 2 = CelebA. All images are channel visualizations by Lucent. Rows correspond to layers where stitching and visualization happens. Leftmost fours columns show first channel, rightmost four columns show second channel of the given layer. In each 4-column block, first column presents Model 1 channel, for comparison. Second column presents Model 2 channel, which in a sense is the target for the last two columns. Third column presents the channel linearly combined from Model 1 channels with the least squares error to Model 2 channel. (Direct matching.) Fourth column presents the substitute of the Model 2 channel created by stitching. Each row present the first two neurons of the corresponding layer.}
\label{fig:lucid_i2r}
\end{figure}

In the first setup, two Inception V1 networks trained on CelebA are stitched at various layers. The feature visualizations are shown in Figure~\ref{fig:lucid_r2r}. In the second setup, Model 1 is trained on ImageNet, and Model 2 is trained on CelebA. The feature visualizations are shown in Figure~\ref{fig:lucid_i2r}.

As seen in Tables~\ref{table:rel_acc_r2r}~and~\ref{table:rel_acc_i2r}, the relative accuracies are consistently high, close to 100\% in the case of the CelebA-CelebA stitchings. Even more notable is the case of the ImageNet-to-CelebA stitchings, where even the highest layers of the randomly initialized CelebA network can reuse the ImageNet features, with only a 3\% relative drop in accuracy.

The accuracy of the original two CelebA models is 91.860\% and 91.862\%, respectively. Note that the supervised CelebA task consists of 40 binary classification subtasks, and reported accuracy is an average (across the validation set) of averages (across subtasks). The tasks are unbalanced, hence the accuracy of the majority baseline is already quite high, 80.5\% (or 87.63\% relative to Model 2 accuracy), which is important to keep in mind when interpreting the values in Table~\ref{table:rel_acc_r2r_ir2}. Nevertheless, it is apparent that stitchings perform nearly perfectly when done between models trained on the same dataset, and perform well even across datasets, especially when done at the lower layers.

\begin{table}[h]
    \centering
    \begin{subtable}[t]{0.45\textwidth}
        \centering
        \vspace{0pt}
        \begin{tabular}{lrr}
        \toprule
        {} & Least Squares &  Frankenstein \\
        layer &        &         \\
        \midrule
        3a    & 99.45\% &  99.88\% \\
        3b    & 98.73\% &  99.86\% \\
        4a    & 99.41\% &  99.85\% \\
        4b    & 99.21\% &  99.84\% \\
        4c    & 99.05\% &  99.87\% \\
        4d    & 98.80\% &  99.91\% \\
        4e    & 98.88\% & 100.01\% \\
        5a    & 99.98\% & 100.07\% \\
        5b    & 99.92\% & 100.08\% \\
        \bottomrule
        \end{tabular}
        \caption{Stitchings between two CelebA networks with different random initializations.}
        \label{table:rel_acc_r2r}
    \end{subtable}\quad
    \begin{subtable}[t]{0.45\textwidth}
        \centering
        \vspace{0pt}
        \begin{tabular}{lrr}
        \toprule
        {} & Least Squares &  Frankenstein \\
        layer &        &        \\
        \midrule
        3a    & 97.72\% & 99.77\% \\
        3b    & 95.90\% & 99.62\% \\
        4a    & 97.73\% & 99.41\% \\
        4b    & 96.74\% & 99.22\% \\
        4c    & 95.27\% & 98.88\% \\
        4d    & 93.45\% & 98.42\% \\
        4e    & 94.37\% & 98.21\% \\
        5a    & 95.75\% & 97.49\% \\
        5b    & 95.57\% & 97.02\% \\
        \bottomrule
        \end{tabular}
    \caption{Imagenet to CelebA stitchings.}
    \label{table:rel_acc_i2r}
    \end{subtable}
\caption{Inception V1 stitchings, relative accuracies.}
\label{table:rel_acc_r2r_ir2}
\end{table}

\end{document}